\pdfoutput=1
\documentclass[11pt]{article}

\usepackage[margin=1in]{geometry}
\usepackage[T1]{fontenc}
\usepackage[utf8]{inputenc}
\usepackage{amsmath,amsthm}
\usepackage{stix2}
\usepackage{microtype}
\usepackage{graphicx}
\usepackage{booktabs,tabularx,array,adjustbox}
\usepackage[font=small,labelformat=empty]{caption}
\usepackage{xcolor}
\usepackage[round]{natbib}
\usepackage[colorlinks=true,citecolor=blue!50!black,linkcolor=blue!50!black,urlcolor=blue!50!black]{hyperref}
\hypersetup{pdftitle={Certified Sparse Attention via Perturbed Retrieval},pdfauthor={Runguo Li}}

\newenvironment{stmt}[1]{\par\medskip\noindent\textbf{#1}\ \itshape\ignorespaces}{\par\medskip}

\title{Certified Sparse Attention via Perturbed Retrieval}
\author{Runguo Li\\
  University of Illinois Urbana-Champaign\\
  \texttt{runguol2@illinois.edu}}
\date{}

\begin{document}
\maketitle

\begin{abstract}
Sparse attention speeds up long-context decoding by reading few keys of the KV cache, but existing methods do not control the error of a query, and on real KV caches their errors have heavy tails. We show that perturbing the score of every key with an independent Exp(1) variable turns threshold retrieval into sampling keys with probability proportional to their attention weights. Lowering the threshold along a geometric grid gives nested samples, a Horvitz--Thompson estimate of the output at every level, and confidence bounds that hold simultaneously over all levels, so each head stops at the first level whose certificate meets the target error $\varepsilon$. With fresh perturbations at every decoding step, each certificate fails with probability at most $\delta$ during generation. At a suitable threshold, the estimator reads $O(\varepsilon^{-2}\log(1/\delta))$ values regardless of the context length, and no adaptive method reads fewer, up to constants. A 4-bit scan with deterministic bounds, shared across a grouped-query-attention group, generates the candidates exactly but reads every key. On three LLMs and ten settings at 16K--128K tokens, no certificate is violated, and at equal keys read our 90th-percentile error is 4.6--7.9$\times$ lower than that of the best of 15 baseline families. At equal bytes, the best baselines have 0.6--1.6 times our 90th-percentile error and at least 4.9 times our largest. End to end, our method stays within one point of dense attention on tasks and within 0.1\% in PG-19 perplexity, and the strongest baselines at equal bytes score as well. On CPU, decoding steps at 32K--128K tokens run 1.1--1.4$\times$ faster than dense attention on the same 16-bit cache.

\end{abstract}

\section{Introduction}\label{sec:1}

To generate one token, every attention head reads the entire KV cache of its context. For Llama-3.1-8B, which uses grouped-query attention (GQA) with 8 KV heads in each of its 32 layers, a 128K-token context means reading about 16 GiB of 16-bit keys and values per decoding step, so decoding is limited by memory bandwidth. Attention weights concentrate on a small fraction of tokens, and sparse attention reads only the keys that a heuristic predicts to matter: recent tokens and attention sinks \citep{xiao2024sinks}, heavy hitters \citep{zhang2023h2o}, page-level bounds \citep{tang2024quest}, learned or hashed signatures \citep{desai2025hash}, or approximate nearest neighbors of the query \citep{liu2025retrieval,chen2026retroinfer}.

\begin{figure}[tbp]
\centering
\includegraphics[width=\linewidth]{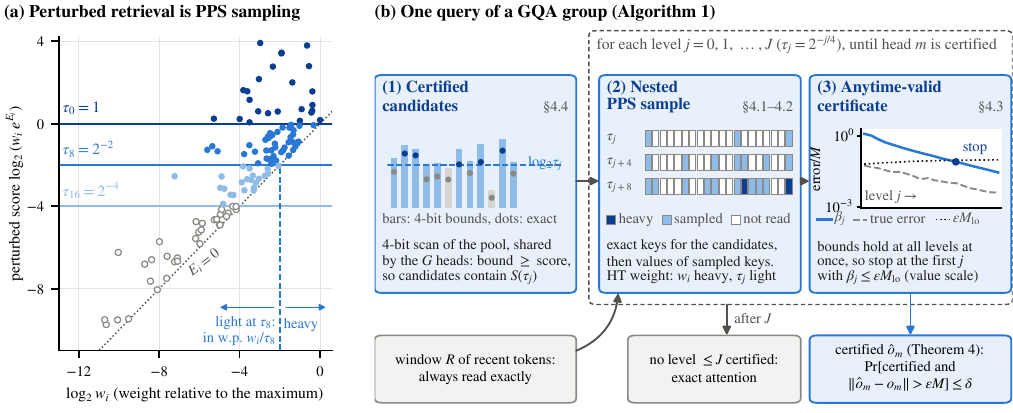}
\caption*{\textbf{Figure 1.} \emph{(a)} Each key $i$ carries a perturbation $E_i\sim\mathrm{Exp}(1)$, drawn afresh at every decoding step, and is in the sample at threshold $\tau$ exactly when its perturbed score lies above the line of $\tau$, which happens with probability $\min(1,w_i/\tau)$; lowering $\tau$ only adds keys (shades give the level at which a key enters). \emph{(b)} One query of a GQA group (Algorithm 1, Appendix B). The shared 4-bit scan bounds every score, so its candidates contain the sample; 8-bit refinement and exact keys decide which candidates enter $S(\tau_j)$, and the values of sampled keys give the Horvitz--Thompson estimate. Each head stops at the first level whose error bound $\beta_j$ (Lemma 1) meets the target, and $\Pr[\text{certified and }\|\hat o_m-o_m\|>\varepsilon M]\le\delta$ (Theorem 4). Synthetic data; logarithms in (a) are base 2.}
\end{figure}

These heuristics do not control the error of an individual query, and on real KV caches this error has a heavy tail. On ten settings from 16K to 128K tokens, the best of 15 baseline families, reading as many keys as our method, has a 90th-percentile error of 0.051--0.090 relative to the value scale but maximum errors of 0.4--1.6 (Section 6.2). Sampling-based methods add randomness, but not a finite-sample certificate for each query. vAttention \citep{desai2026vattention} sizes a uniform sample under a central-limit approximation, MagicPIG \citep{chen2025magicpig} samples keys by locality-sensitive-hashing collision probabilities, and SANTA \citep{lee2026stochastic} samples values from the exact softmax.

We ask whether sparse attention can read few keys and still certify its error at every step. Suppose that each key $i$ carries a perturbation $E_i\sim\mathrm{Exp}(1)$ drawn independently of the query, and let $w_i=e^{s_i-c^\ast}$ be its unnormalized attention weight. Then the keys whose perturbed score $s_i-c^\ast+E_i$ exceeds $\log\tau$ form a Poisson sample that contains key $i$ with probability $\min(1,w_i/\tau)$. This is threshold sampling with probabilities proportional to size (PPS) \citep{duffield2005threshold,duffield2007priority}, under which the Horvitz--Thompson (HT) estimates \citep{horvitz1952} of the softmax numerator and denominator are unbiased. Lowering $\tau$ only adds keys, so we descend a geometric grid of thresholds and stop as soon as a confidence bound, valid simultaneously over all levels, meets the target (Figure 1).

We certify the error of the resulting estimate per head and decoding step, with finite-sample bounds that stay valid at a data-dependent stopping level (Sections 3--4). At a suitable threshold, $2\varepsilon^{-2}\log(4/\delta)(1+o(1))$ values reach error $\varepsilon M$ with probability $1-\delta$ independently of the context length, which is optimal up to constants; our stopping rule, which does not know this threshold, reads more (Section 5). A shared 4-bit scan recovers the sample exactly but reads every key (Section 4.4). On three LLMs and ten settings, we compare with 15 baseline families at matched keys and bytes, end to end, and in speed on CPU (Section 6). The certificates read 1.8--3.5$\times$ the keys of an uncertified central-limit rule, and the linear scan sets a floor on the bytes read. At equal bytes, the strongest baselines reach our typical error and our end-to-end quality, while their largest errors stay at least 4.9$\times$ ours (Sections 6.2 and 7).

\section{Related Work}\label{sec:2}

\textbf{Sparse attention for decoding.} Besides the heuristics above, training-free methods keep a compressed prompt cache \citep{li2024snapkv} or select keys for the current query from page summaries \citep{xiao2024infllm}, low-rank or partial projections \citep{ribar2024sparq,sun2025shadowkv}, product quantization \citep{zhang2025pqcache}, clustering \citep{liu2025clusterkv} or adaptive top-$p$ budgets \citep{lin2025twilight}, and others train sparsity into the model \citep{yuan2025nsa}. These methods drop the mass of the unselected keys and give no per-query error bound (Appendix F.1).

\textbf{Sampling and estimation.} Besides vAttention, MagicPIG and SANTA, SOCKET \citep{joshi2026socket} samples with soft collision kernels and SANTA++ \citep{lee2026santa} samples clusters through representatives. LSH attention, random features and kernel-density estimators \citep{kitaev2020reformer,choromanski2021performer,zandieh2023kde} bound the error of the whole attention matrix, and top-$k$ truncation has deterministic per-query bounds once all scores are known \citep{tzachristas2025topk}. We sample keys in proportion to their weights without scoring the unsampled keys exactly, and we bound the output error with finite-sample bounds that hold at a stopping level chosen from the data.

\textbf{Theory.} Compression and streaming lower bounds limit how far a KV cache can be summarized \citep{haris2025,chen2026streaming}; ours counts the values read by an algorithm that knows all keys. Exponential and Gumbel perturbations implement weighted sampling \citep{efraimidis2006,kool2019}. Closest to our use of them, \citet{mussmann2017} sample from log-linear models by maximum inner product search over Gumbel-perturbed vectors, and \citet{mussmann2026sum} estimate sums such as softmax normalizers from $O(\log n)$ retrieved vectors. Our threshold gives nested samples with constant HT weights and a certificate that can be checked at every level, and we certify a ratio of vector sums.

\section{Problem Setup}\label{sec:3}

\textbf{Attention at one decoding step.} Fix a layer, a KV head and a decoding step. The cache holds keys $k_i\in\mathbb{R}^d$ and values $v_i\in\mathbb{R}^d$ for $i=1,\dots,n$. A query $q$ has scores $s_i=q^\top k_i/\sqrt d$ and maximum $c^\ast=\max_i s_i$. Write $w_i=e^{s_i-c^\ast}\in(0,1]$, $D=\sum_i w_i$, $N=\sum_i w_iv_i$, the attention output $o=N/D$ and the weights $a_i=w_i/D$. With GQA, $G$ query heads share the keys and values of one KV head (Llama-3 and Qwen3 have $G=4$, Qwen2.5-7B has $G=7$).

\textbf{Error measures.} We measure the error of an estimate $\hat o$ on three scales: the denominator ($|\hat D-D|\le\varepsilon D$), the output norm ($\|\hat o-o\|\le\varepsilon\|o\|$) and the value scale ($\|\hat o-o\|\le\varepsilon M$, where $M^2=\sum_i a_i\|v_i\|^2$). With the cancellation factor $\kappa=M/\|o\|\ge1$, the value scale is $\kappa$ times the output norm. On Llama-3.1-8B at 16K, $\kappa$ has a median of 3.6 and a 90th percentile of 8.7. At $\varepsilon=0.1$, a value-scale certificate therefore allows an error of 36\% of $\|o\|$ for the median head and 87\% at the 90th percentile; it bounds how accurately a head mixes its values. We use it as the main measure because the output-norm certificate costs 2.3--4.6 times the keys (Section 6.3); whether it suffices for model quality is an empirical question (Section 6.4).

\textbf{Certificates.} A certified sparse attention algorithm takes a target error $\varepsilon$ and a failure probability $\delta$, and it either outputs an estimate that carries a certificate or falls back to exact attention. Its guarantee concerns one call, a head at one decoding step: whenever the query, keys and values of the call are independent of the algorithm's internal randomness,
\[\Pr\big[\hat o\text{ is certified and its error exceeds }\varepsilon\big]\le\delta .\]
Our default redraws the randomness at every decoding step, with seeds that differ between layers and KV heads (Section 4.1). A call's query, keys and values then depend only on earlier steps and layers, whose randomness is independent of its own, so the requirement holds at every call of closed-loop generation, and a union bound over the $H$ query heads and $T$ steps bounds the probability of any certified error above $\varepsilon$ by $HT\delta$. Fixing the randomness across steps is not covered (Appendix F.4).

\section{Method}\label{sec:4}

\subsection{Perturbed keys and threshold sampling}\label{sec:4.1}

We split the cache of each KV head into a window $R$ and a pool $P$. The window holds the most recent 64 tokens and the incomplete last block, and we always read it exactly. The pool holds complete blocks of 128 tokens. Each pool key $i$ carries a perturbation $E_i=-\ln u_i$, where $u_i\in(0,1]$ is a 53-bit uniform number obtained by hashing the token position with a seed. By default the seed depends on the layer, the KV head and the decoding step, and a call recomputes the perturbations of all pool keys before it scans them (Section 6.5 gives the cost). We treat the $E_i$ as independent $\mathrm{Exp}(1)$ variables, independent of the queries, keys and values. This is an assumption about the hash, and the 53-bit uniforms cannot represent inclusion probabilities below $2^{-53}$. For a threshold $\tau>0$, we define the sampled set, the heavy set and the light set
\[\begin{aligned}S(\tau)&=\{i\in P:\ s_i-c^\ast+E_i\ge\log\tau\},\\ B(\tau)&=\{i\in P:\ w_i\ge\tau\},\qquad A(\tau)=P\setminus B(\tau).\end{aligned}\]
Since $\Pr[E_i\ge\log(\tau/w_i)]=\min(1,w_i/\tau)$, every heavy key is sampled, and a light key is sampled independently with probability $w_i/\tau$. This is threshold (Poisson PPS) sampling \citep{duffield2005threshold}; the $m$ largest perturbed scores would give priority sampling \citep{duffield2007priority}. Within a call the perturbations are fixed, so lowering $\tau$ only adds keys to the current sample.

\subsection{Horvitz--Thompson estimation}\label{sec:4.2}

For a fixed $\tau$, we compute the sums over the window and the heavy set exactly, which gives $D_R,N_R$ and $D_B=\sum_{B}w_i$, $N_B=\sum_{B}w_iv_i$. Each sampled light key has inclusion probability $w_i/\tau$, so its Horvitz--Thompson weight is $w_i/(w_i/\tau)=\tau$ \citep{horvitz1952}. With $k=|S\cap A|$,
\[\begin{aligned}\hat D&=D_R+D_B+\tau k,\qquad \hat o=\hat N/\hat D,\\ \hat N&=N_R+N_B+\tau\textstyle\sum_{i\in S\cap A}v_i .\end{aligned}\]
The light keys enter the estimate through their values alone, so a single retrieval suffices and no reweighting pass follows.

\subsection{Anytime-valid certificates}\label{sec:4.3}

We evaluate the estimator on the grid $\tau_j=2^{-j/r}$, $j=0,\dots,J$ (we use $r=4$ and $J=160$), and give each level the failure budget $\delta_j=\delta/((J+1)e)$. Here $e\in\{1,3,4\}$ is the number of probabilistic events the target needs: one for the denominator, three for the output norm, four for the value scale. At each level we compute the following.

\begin{enumerate}
\item
\textbf{Denominator interval.} The light-sample count $k$ is Poisson-binomial with mean $\mu=D_A/\tau$, and its moment generating function is dominated by that of a Poisson variable with the same mean. Hence the Chernoff set $\{\mu:\ k\ln(k/\mu)+\mu-k\le\ln(2/\delta_j)\}=[\mu_{\rm lo},\mu_{\rm hi}]$ contains $\mu$ with probability at least $1-\delta_j$. This gives $|\hat D-D|\le dD:=\tau\max(k-\mu_{\rm lo},\,\mu_{\rm hi}-k)$.
\item
\textbf{Centered numerator.} Let $c=(N_R+N_B)/(D_R+D_B)$, which is deterministic given $\tau$. The vector $Z_c=\hat N-N-c(\hat D-D)=\sum_{i\in A}(\tau X_i-w_i)(v_i-c)$ is a sum of independent zero-mean terms of norm at most $\tau R_c$, with $R_c=V_{\max}+\|c\|$. A Bernstein inequality in Hilbert space \citep{pinelis1994} gives $\|Z_c\|\le\zeta:=\sqrt{2\tau\hat G^+\ln(2/\delta_j)}+\tfrac23\tau R_c\ln(2/\delta_j)$. Here $\hat G^+$ is a Bernstein upper confidence bound on $G_A(c)=\sum_{A}w_i\|v_i-c\|^2$, computed from its unbiased estimate $\tau\sum_{S\cap A}\|v_i-c\|^2$.
\item
\textbf{Output error.} By Lemma 1 below, on these events $\|\hat o-o\|\le\beta:=(\zeta+\|\hat o-c\|\,dD)/(\hat D-dD)$.
\item
\textbf{Stopping rules.} A level is certified if $dD\le\varepsilon(\hat D-dD)$ (denominator), $\beta\le\varepsilon\|\hat o\|/(1+\varepsilon)$ (output norm) or $\beta\le\varepsilon M_{\rm lo}$ (value scale), where $M_{\rm lo}^2=(Q_R+Q_B+Q_A^{\rm lo})/(\hat D+dD)$, $Q=\sum w_i\|v_i\|^2$, and $Q_A^{\rm lo}$ is a Bernstein lower confidence bound on $Q_A$. The first certified level determines the output, and if no level up to $J$ is certified, the head falls back to exact attention.
\end{enumerate}

We center at $c$ because, without centering, the deviation of $\hat N$ scales with the norms of the values, whereas with centering it scales with their spread around the known part of the output. Together with the Chernoff interval, centering reduces the keys read by 1.8--1.9 times compared with an uncentered Bernstein certificate (Appendix D.7).

\subsection{Certified candidate generation}\label{sec:4.4}

To return exactly $S(\tau_j)$ without reading all keys, we bound every score from above using compressed keys and read exact keys only for the remaining candidates (details in Appendix B.2).

\begin{enumerate}
\item
\textbf{4-bit scan.} We quantize keys per 128-token block and per channel to 4 bits, as in KIVI \citep{liu2024kivi}, and split the query weights into two int8 levels so that AVX-512 VNNI computes exact integer dot products. Adding the dequantization, residual and rounding errors gives upper bounds $u_i\ge s_i$.
\item
\textbf{Candidate levels.} The bound gives the earliest level at which key $i$ could enter the sample, $j^c_i=\lceil -r(\max_m(u_{im}-c^\ast_m)+E_i)/\ln2\rceil$, where $m$ ranges over the query heads of the GQA group. We bucket keys by this level with a counting sort and expand buckets only as deep as the descent reaches.
\item
\textbf{Refinement and exact reads.} We rescore the keys that reach the current level with 8-bit codes. For keys that may still enter the sample, we read the exact 16-bit keys to obtain exact scores and inclusion levels, and for sampled keys we also read the values.
\item
\textbf{Exact maximum.} Only keys whose upper bound exceeds the best lower bound can attain $c^\ast$. We score these keys exactly, and they make up 0.1--1\% of the pool.
\end{enumerate}

The query heads of a GQA group share the scan, the exact-key reads and the perturbations. We take the union of the heads' candidates, and each head certifies and stops on its own.

\subsection{Cost}\label{sec:4.5}

When we append a token, we store its 16-bit key and value. When a block fills, we also write the block's 4-bit and 8-bit codes and the largest value norm in the block, from which a query obtains $V_{\max}$ (Algorithm 1 in Appendix B states a query in full). The index stores about 720 bytes per token for $d=128$, against 512 bytes for 16-bit keys and values (a 40\% overhead, or 16\% without the 8-bit codes). Each query moves a fixed 84 bytes per pool token: it writes the step's perturbations (4 bytes) and reads them with the 4-bit codes and per-block scales (80 bytes). This is 16.4\% of the bytes of dense attention and the floor of the bytes we count; it grows linearly with $n$, whereas the keys and values read beyond it scale with the sample size, which Theorem 2 bounds independently of $n$.

\section{Theory}\label{sec:5}

Throughout, $\tau$, the window and the sets $B(\tau)$, $A(\tau)$ are deterministic, and the only randomness comes from the perturbations. Proofs are in Appendix A.

\begin{stmt}{Theorem 1 (Unbiasedness and variance).}
For a fixed $\tau>0$: (a) $\mathbb E\hat D=D$ and $\mathbb E\hat N=N$; (b) $\mathrm{Var}\,\hat D=\sum_A w_i(\tau-w_i)\le\tau D_A$; (c) for any fixed $c$, $\mathbb E\|Z_c\|^2=\sum_A w_i(\tau-w_i)\|v_i-c\|^2\le\tau\,G_A(c)$; (d) $\mathbb E|S|=|B|+D_A/\tau\le D_P/\tau$.
\end{stmt}

\begin{stmt}{Lemma 1 (Error identity).}
For any $c\in\mathbb R^d$, $\hat o-o=\big[Z_c-(\hat o-c)(\hat D-D)\big]/D$.
\end{stmt}

\begin{stmt}{Theorem 2 (Sample size independent of $n$).}
Let $m>0$ and $\tau=D/m$, so that $\mathbb E|S\setminus R|\le m$. Let $L=\ln(4/\delta)$, $\eta=\sqrt{2L/m}+L/(3m)<1$, $\sigma_v^2=\sum_i a_i\|v_i-o\|^2=M^2-\|o\|^2$ and $R_o=V_{\max}+\|o\|$. With probability at least $1-\delta$, $|\hat D-D|/D\le\eta$ and
\[\|\hat o-o\|\le\frac{\sigma_v\sqrt{2L/m}+2R_oL/(3m)}{1-\eta}.\]
Hence $m=2L\varepsilon^{-2}(1+o(1))$ values suffice for error $\varepsilon M$, and $m\approx2L(\kappa^2-1)\varepsilon^{-2}$ for error $\varepsilon\|o\|$, whatever $n$ and the entropy of the attention distribution.
\end{stmt}

Theorem 2 analyzes the estimator at the threshold $\tau=D/m$, which requires $D$. Proposition 1 below bounds what the adaptive rule reads, and the candidate scan touches every key regardless (Section 4.5).

\begin{stmt}{Theorem 3 (Lower bound).}
Let $\varepsilon\in(0,1/4]$, $\delta\in(0,1/12)$, $m\ge1$ and $n\ge m^2/\delta$, and consider any algorithm that knows all keys and weights and reads at most $m$ values, chosen adaptively and possibly at random. (i) For every $\kappa$ with $\kappa^2\ge1+22\varepsilon^2$, there are two instances with cancellation factor $\kappa$ such that achieving $\Pr[\|\hat o-o\|<\varepsilon\|o\|]\ge1-\delta$ on both requires $m\ge(\kappa^2-1)\ln(1/(6\delta))/(8.8\varepsilon^2)$. (ii) For $\varepsilon\le0.17$, achieving $\Pr[\|\hat o-o\|<\varepsilon M]\ge1-\delta$ on two instances requires $m\ge\ln(1/(6\delta))/(17.6\varepsilon^2)$.
\end{stmt}

Up to constants, the PPS estimator at a suitable threshold thus reads the optimal number of values in both $\varepsilon^{-2}\log(1/\delta)$ and $\kappa^2-1$, the unavoidable price of measuring the error against $\|o\|$. For our adaptive algorithm, Proposition 1 gives a weaker upper bound. We have not optimized the constants: as $\delta\to0$ the leading terms of the upper and lower bounds differ by about 18$\times$ (output norm) and 35$\times$ (value scale). The proof uses identical keys whose values differ between two instances, so that adaptive reading reduces to sampling without replacement, and applies the Bretagnolle--Huber inequality \citep{bretagnolle1979}.

\begin{stmt}{Lemma 2 (Candidate generation misses nothing).}
In exact arithmetic, every key in $S(\tau_j)$ is scored exactly no later than level $j$, and the computed $c^\ast$ is the true maximum.
\end{stmt}

The implementation covers floating-point rounding with explicit margins.

\begin{stmt}{Theorem 4 (Anytime-valid certificates).}
Consider one attention call whose query, keys and values are independent of the perturbations. Then for each query head and each of the three error measures, with the probability taken over the perturbations,
\[\Pr\big[\hat o\text{ is certified and its error exceeds }\varepsilon\big]\le\delta .\]
\end{stmt}

The proof intersects the confidence events of all levels into one event of probability at least $1-\delta$, on which every stopping rule implies its target, so the data-dependent stopping level is harmless. The $G$ query heads of a group share perturbations and are certified simultaneously with probability at least $1-G\delta$. Covering all $H$ query heads and $T$ steps of a generation requires $\delta/(HT)$ per call, which enters only through $\log(1/\delta)$ and at $T=512$ costs 1.50--1.69 times the keys of $\delta=0.05$ (Section 6.3).

\begin{stmt}{Proposition 1 (Adaptive stopping reads $O(\nu^2\varepsilon^{-2}L)$ values).}
Let $V$ be the largest value norm and $\nu=V/M\ge1$, and let $L=\ln(2/\delta_j)$ and $\varepsilon\le1$. There is an event of probability at least $1-\delta$ on which, for the value-scale measure, every level with $\tau_j\le\tau^\dagger=\varepsilon^2D/(400\nu^2L)$ satisfies the stopping rule, so that the algorithm stops with at most $\max(2D,\ 2^{1+1/r}\cdot400\,\nu^2L/\varepsilon^2)+L$ values read. For the output-norm measure, the same holds with $\tau^\dagger=\varepsilon^2D/(1024\,\kappa^2\nu^2L)$.
\end{stmt}

Apart from the term $2D$, which comes from starting the grid at $\tau=1$, the bound does not depend on $n$; compared with Theorem 2 it loses a constant and the factor $\nu^2$ (median 2.6--4.1 on real KV caches). In practice, the rule reads 2.7--4.9 times the keys of the oracle that knows the true error (Table D.7b).

\section{Experiments}\label{sec:6}

\subsection{Setup}\label{sec:6.1}

\textbf{Models and data.} We use Llama-3.1-8B-Instruct \citep{grattafiori2024llama3} (32 layers, $G=4$), Qwen3-8B \citep{yang2025qwen3} (36 layers, $G=4$) and Qwen2.5-7B-Instruct \citep{yang2024qwen25} (28 layers, $G=7$). After a bf16 prefill, we capture keys, values and the queries of decoding steps 0, 8, 16, 24 and 31 on LongBench at 16K tokens \citep{bai2024longbench} and RULER at 32K \citep{hsieh2024ruler} for all three models, and on RULER and LongBench-v2 \citep{bai2025longbench2} at 64K and 128K for Llama: ten settings with 20K--69K queries each (Appendix C.1).

\textbf{Metrics and methods.} Our main metric is $\|\hat o-o\|/M$ (median, 90th percentile and maximum over all layers, heads and evaluated steps). \emph{Keys read} is the number of keys whose values a method reads, divided by the context length, and \emph{bytes} count 4-bit, 8-bit and 16-bit traffic relative to dense attention. Our method uses the value-scale measure with $\varepsilon\in\{0.05,0.1,0.2,0.3\}$, $\delta=0.05$, $r=4$ and a window of 64 tokens. We compare with 15 baseline families, variants of 11 methods: twelve sparse-attention-hub maskers (sink + local, oracle top-$k$ and top-$p$, uniform sampling, Quest, PQCache, SOCKET, MagicPIG, HashAttention, and vAttention on oracle top-$k$, PQCache or HashAttention), ports of the authors' code for MagicPIG and Quest, and our CPU port of RetroInfer, whose released implementation requires CUDA. We also evaluate SANTA and its variants. Every family is tuned over a grid of budgets (Appendix C.2). All experiments run on Intel Xeon CPUs (8555C and 8358) without GPUs.

\subsection{Error at a matched number of keys read}\label{sec:6.2}

Table 1 compares our method at $\varepsilon=0.1$ with the best configuration of each baseline family among those that read at most 1.1 times as many keys. Our 90th-percentile error is 0.010--0.012 and our maximum at most 0.029, whereas the best baseline's 90th percentile is 4.6--7.9 times higher (0.051--0.090), with maxima of 0.39--1.62 (Figures 2 and 3). Even oracle top-$k$, which selects keys by their true scores, has a 90th-percentile error of 0.068--0.109, because it discards the mass of the unselected keys. With four times our keys, the best baseline at 64K--128K is still 2.9--3.3 times worse (Table D.2b); relative to the output norm the gap is 3.2--6.2 times (Table 1).

Keys read ignore the metadata that each method scans for every key, such as our 4-bit codes. We therefore also charge our kernel its measured bytes and each baseline a lower bound on its bytes (Appendix C.2). Within our bytes, the best baseline, usually the MagicPIG port reading 16--37\% of the keys, reaches 0.6--1.6 times our 90th-percentile error, so it matches or beats our typical error on five of the ten settings (Table D.2c, Figure D.2). Its largest error remains 13--83 times ours, and over all baseline configurations within our bytes the smallest maximum error is still 4.9--29 times ours. Dense attention over a 4-bit quantized cache reads similar bytes with 2.6--5.3 times our 90th-percentile error, and over an 8-bit cache it has a 3--5 times lower one but reads 1.5--2.8 times our bytes. A PPS sample from the 4-bit scores matches our 90th percentile only with our per-query budget (Appendix D.2), which the certificate chooses.

\begin{table}[tbp]
\centering\small
\caption*{\textbf{Table 1.} Error/$M$ at matched keys read ($\varepsilon=0.1$; baselines read $\le$ 1.1$\times$ our keys). ``Families'' counts the baseline families that have a configuration within the budget. The last column gives the 90th-percentile error relative to $\|o\|$ for our method and for the best baseline under that error.}
\begin{adjustbox}{max width=\linewidth}\begin{tabular}{llllll}
\toprule
Model $\cdot$ data & Ours: keys / p90 / max & Families & Best baseline: keys, p90 / max & Ratio (p90) & Error/$\|o\|$ p90: ours / best (ratio) \\
\midrule
Llama $\cdot$ LongBench 16K & 2.70\% / 0.011 / 0.021 & 11/15 & MagicPIG (ref.) 2.75\%, 0.059 / 0.83 & 5.3$\times$ & 0.087 / 0.311 (3.6$\times$) \\
Llama $\cdot$ RULER 32K & 1.35\% / 0.011 / 0.025 & 10/15 & RetroInfer 1.24\%, 0.058 / 0.49 & 5.0$\times$ & 0.075 / 0.287 (3.8$\times$) \\
Llama $\cdot$ RULER 64K & 0.70\% / 0.012 / 0.028 & 14/15 & RetroInfer 0.58\%, 0.059 / 0.68 & 5.1$\times$ & 0.076 / 0.294 (3.8$\times$) \\
Llama $\cdot$ LongBench-v2 64K & 0.61\% / 0.011 / 0.021 & 12/15 & RetroInfer 0.57\%, 0.051 / 0.39 & 4.6$\times$ & 0.085 / 0.277 (3.2$\times$) \\
Llama $\cdot$ RULER 128K & 0.46\% / 0.012 / 0.029 & 12/15 & RetroInfer 0.47\%, 0.077 / 0.65 & 6.5$\times$ & 0.073 / 0.352 (4.8$\times$) \\
Llama $\cdot$ LongBench-v2 128K & 0.42\% / 0.012 / 0.022 & 12/15 & RetroInfer 0.32\%, 0.069 / 0.65 & 5.9$\times$ & 0.081 / 0.340 (4.2$\times$) \\
Qwen3 $\cdot$ LongBench 16K & 2.15\% / 0.010 / 0.021 & 10/13 & RetroInfer 1.67\%, 0.072 / 1.46 & 7.2$\times$ & 0.049 / 0.270 (5.5$\times$) \\
Qwen3 $\cdot$ RULER 32K & 1.50\% / 0.011 / 0.025 & 10/13 & RetroInfer 1.37\%, 0.073 / 1.62 & 6.9$\times$ & 0.046 / 0.252 (5.5$\times$) \\
Qwen2.5 $\cdot$ LongBench 16K & 2.67\% / 0.011 / 0.022 & 11/13 & RetroInfer 2.26\%, 0.072 / 0.95 & 6.6$\times$ & 0.057 / 0.308 (5.4$\times$) \\
Qwen2.5 $\cdot$ RULER 32K & 1.73\% / 0.011 / 0.027 & 10/13 & RetroInfer 1.22\%, 0.090 / 0.98 & 7.9$\times$ & 0.054 / 0.333 (6.2$\times$) \\
\bottomrule
\end{tabular}\end{adjustbox}
\end{table}

\begin{figure}[tbp]
\centering
\includegraphics[width=\linewidth]{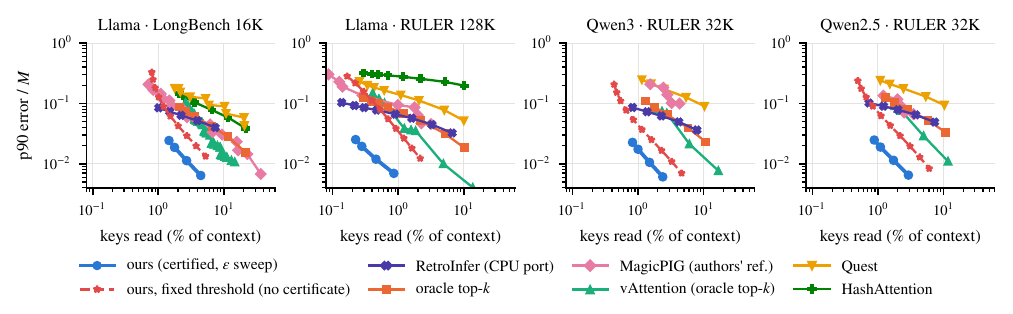}
\caption*{\textbf{Figure 2.} 90th-percentile error/$M$ versus keys read for our method ($\varepsilon\in\{0.05,0.1,0.2,0.3\}$), its fixed-threshold variant without certificates, and the strongest baseline families (Pareto fronts of their tuning grids) on four settings (Llama LongBench 16K and RULER 128K, Qwen3 and Qwen2.5 at 32K). All ten settings are in Appendix D.}
\end{figure}

\begin{figure}[tbp]
\centering
\includegraphics[width=\linewidth]{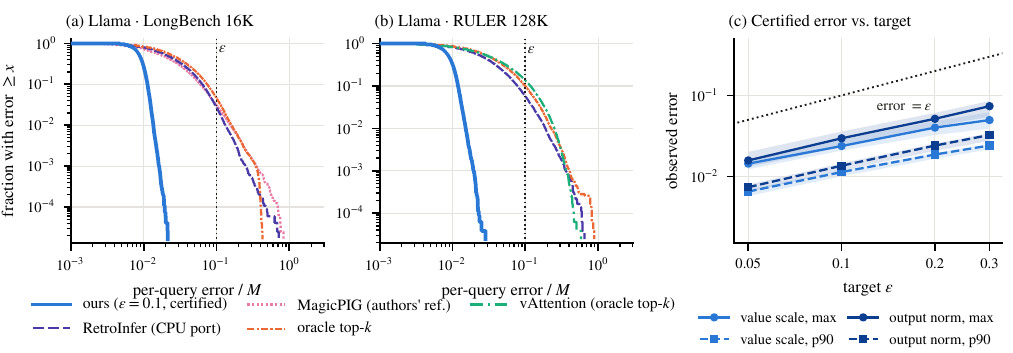}
\caption*{\textbf{Figure 3.} Per-query errors at matched keys read, and the calibration of the certificates. \emph{(a, b)} Fraction of queries whose error$/M$ is at least $x$, for our method at $\varepsilon=0.1$ and the three best baseline families of Table 1 at their configurations in Table D.2a. The dotted line marks $\varepsilon$. \emph{(c)} Observed error against the target $\varepsilon$ for the value-scale and output-norm measures, as the maximum and the 90th percentile over the queries of a setting (median over the ten settings, range shaded). The dotted line is the certified bound.}
\end{figure}

\subsection{Validity, measures and context length}\label{sec:6.3}

Across 120 configurations (10 settings $\times$ 3 measures $\times$ 4 values of $\varepsilon$), no (layer, head, step) error exceeds $\varepsilon$, no single layer has a violation, and no head falls back to exact attention. The certificates are conservative: observed errors are about $\varepsilon/10$ at the 90th percentile, and the largest is $0.31\varepsilon$ for the value scale and $0.41\varepsilon$ for the output norm (Figure 3c). With perturbations drawn afresh at every call, our default, keys read and the 90th-percentile error at $\varepsilon=0.1$ are unchanged and the largest error is $0.37\varepsilon$ (Table D.3b). A failure probability for whole generations, $\delta=0.05/(HT)$ with $T=512$ ($9.5\times10^{-8}$ per call on Llama), reads 1.50--1.69 times the keys of $\delta=0.05$, again without violations. The output-norm measure reads 2.3--4.6 times the keys of the value scale, a cost that Theorem 3 shows to be partly inherent, and the denominator measure reads fewer but leaves the output error uncontrolled (Table D.3).

On Llama RULER at $\varepsilon=0.1$, our method reads 443 keys at 32K, 458 at 64K and 601 at 128K (Table D.4, Figure D.6a). A fourfold longer context thus costs 36\% more keys, because a larger share of the mass lies in the sampled tail, whereas bytes read fall only from 21.8\% to 18.8\% of dense attention, because the 4-bit scan reads every key.

\subsection{End-to-end tasks and perplexity}\label{sec:6.4}

We evaluate on six LongBench and four RULER tasks with 12--50 samples per task. All methods share one dense prefill and decode greedily in fp32 as one batch. Our method draws fresh perturbations at every step and audits its certificates with exact attention at every fourth step, without a violation (Appendix C.6). Mean scores stay within one point of dense attention, and 79--98\% of the outputs are identical to dense attention's (Table 2). Two of the eight task-stratified intervals exclude zero, in opposite directions; on Llama LongBench, two further seed families give intervals that contain zero (Table D.9e).

\begin{table}[tbp]
\centering\small
\caption*{\textbf{Table 2.} End-to-end score change versus dense attention ($\varepsilon=0.1$, fresh perturbations at every step), with task-stratified 95\% CIs. The last column is the ablation with perturbations fixed per KV head (Table D.9a).}
\begin{adjustbox}{max width=\linewidth}\begin{tabular}{lllll}
\toprule
Model $\cdot$ data (samples) & Keys read & $\Delta$ score [95\% CI] & Same output as dense & Fixed perturbations: $\Delta$ score \\
\midrule
Llama $\cdot$ LongBench (300) & 5.26\% & $-$0.29 [$-$0.63, $-$0.03] & 0.80 & $-$0.07 [$-$0.30, +0.12] \\
Llama $\cdot$ RULER 32K (200) & 1.50\% & $-$0.65 [$-$2.05, +0.35] & 0.93 & +0.25 [$-$0.30, +0.95] \\
Llama $\cdot$ RULER 64K (100) & 0.78\% & 0.00 [0.00, 0.00] & 0.98 & $-$0.60 [$-$1.80, 0.00] \\
Llama $\cdot$ RULER 128K (48) & 0.49\% & 0.00 [0.00, 0.00] & 0.90 & 0.00 [0.00, 0.00] \\
Qwen3 $\cdot$ LongBench (150) & 5.10\% & +0.17 [+0.06, +0.28] & 0.80 & +0.10 [$-$0.05, +0.25] \\
Qwen3 $\cdot$ RULER 32K (100) & 1.53\% & +1.00 [$-$0.20, +3.10] & 0.85 & +1.00 [0.00, +3.00] \\
Qwen2.5 $\cdot$ LongBench (150) & 5.38\% & +0.07 [$-$0.09, +0.25] & 0.79 & +0.28 [$-$0.22, +1.02] \\
Qwen2.5 $\cdot$ RULER 32K (100) & 1.73\% & $-$0.10 [$-$0.30, 0.00] & 0.83 & +0.20 [$-$0.50, +1.20] \\
\bottomrule
\end{tabular}\end{adjustbox}
\end{table}

Oracle top-$k$ also matches dense attention on these tasks (Table D.9b), so we add RULER common-words extraction (cwe) at 32K, which aggregates over the whole context. Reading as many keys as our method at every layer, step and query head, oracle top-$k$ drops from 27.8, 86.0 and 74.8 points to 12.0, 44.4 and 34.4 on Llama, Qwen3 and Qwen2.5, whereas our method scores 25.2, 86.0 and 74.4 (Table D.9c). The strongest baselines are another matter: MagicPIG, RetroInfer and, on Llama, vAttention, at configurations that read 1.0--1.3 times our bytes, score within 3 points of dense attention, and no paired difference from our method is significant by the sign test (Table D.9d).

On PG-19 \citep{rae2020pg19}, we prefill each book densely and score its last 512 tokens with teacher forcing (Appendix C.7). Our method changes perplexity by $-$0.02\% to +0.07\% in all five settings, and its KL divergence from dense attention's next-token distribution is 0.0002--0.0006 (Table D.12). Oracle top-$k$ reading the same keys raises perplexity by 4.8--11.5\% (Figure D.6b), because it drops the mass of the unread keys. At 0.92--1.04 times our bytes, the baselines change perplexity by $-$0.07\% to +0.17\%, none significantly different from our method, but their KL divergence is 1.6--4.7 times ours, except for MagicPIG on Qwen2.5 (0.45 times; Table D.12b).

We also sample at temperature 0.7, four sequences per prompt with the same Gumbel noise for every method (Appendix C.6). On cwe at 32K (11 samples each on Llama and Qwen2.5), no score difference is significant (Table D.9f). On Llama, 77\% of our sequences equal dense attention's, against 50--66\% for the baselines (significantly more only than RetroInfer's), and before they differ our KL divergence from dense attention is 2.8--7.9 times lower; on Qwen2.5, MagicPIG is closer (68\% against 57\%; 0.44 times our KL). Over 1,024 tokens generated on seven PG-19 books with Llama, the baselines' mean KL divergence is 1.6--4.3 times ours, and none grows (Table D.12c).

\subsection{Speed on CPU}\label{sec:6.5}

We report speedups relative to our own dense kernel over the same 16-bit keys and values, which is 1.7--2.1 times faster than llama.cpp's CPU flash-attention operator on the same cache (Table D.11c). Measured alone, attention for all heads of a step is 1.8--3.5 times faster than this kernel on Llama at 64K--128K and 1.0--2.1 times faster at 16K--32K (1 and 24 threads; Figure D.7a, Table D.11a); fresh perturbations at every step add a median 4\% to this time (at most 14\%). With seven query heads per group (Qwen2.5), attention is not faster (0.9--1.1$\times$). In whole fp32 decoding steps of Llama, attention is 1.4--2.0 times faster, a smaller gain because each layer's call can use only its 8 KV heads in parallel, and the unchanged linear layers take 200--220 ms, so steps are 1.1$\times$ faster at 32K and 1.4$\times$ at 128K (Figure D.7b, c; Table D.11b). Indexing the prompt adds 0.6--2.5 s to the first step, recovered after 11--18 steps (Figure D.7).

\subsection{Ablations}\label{sec:6.6}

Table D.0 summarizes the ablations at $\varepsilon=0.1$ (Appendix D.5--D.8). Adaptive stopping matters most: a single fixed threshold over-reads on peaked heads and under-reads on flat ones, so at the closest number of keys read its 90th-percentile error is 2.4--9.6 times ours (Figure D.4). The guarantee costs keys: our rule reads 1.8--3.5 times the keys of an uncertified central-limit rule and 4.1--4.9 times those of the unrealizable oracle stopping level (Figure D.5). The central-limit rule fails on up to 30\% of the queries of single heads (Table D.7c), and run at a larger $\varepsilon$ to read about as many keys, our rule has a lower 90th-percentile and maximum error in all ten comparisons (Table D.7d). Sharing the scan across a GQA group cuts bytes by 2.7--4.0 times.

\section{Limitations and Discussion}\label{sec:7}

\textbf{Conservative certificates.} The slack of the certificates (Section 6.3) comes from the union bound over levels and events, the range term set by the largest value norm, and the upper bound on the spread of the light values. Time-uniform confidence sequences and betting estimators \citep{howard2021,waudbysmith2023} address the first and the third (Appendix F.2).

\textbf{Linear candidate generation.} The 4-bit scan moves a fixed 16.4\% of the bytes of dense attention. This floor caps the speedup, and at equal bytes it lets the best baselines read many more keys and match our 90th-percentile error on half of the settings, though not our maximum error. A sublinear index that misses at most an $\eta$ fraction of the attention mass would bias the denominator by at most $\eta D$ (Proposition 2); a certified one remains open.

\textbf{End-to-end value.} At equal bytes, the strongest baselines score as well as our method, greedy or sampled, so their heavier operator-level error tail did not cost accuracy on our tasks. Our method adds a per-call guarantee and, on Llama, closer agreement with dense attention when sampling; on Qwen2.5, MagicPIG agrees more closely. Within a call, the certificate sets each head's budget, which a fixed budget on the same scan does not match (Section 6.2), and sends heads it cannot certify to exact attention; its effect on task outcomes is open (Appendix F.5).

\textbf{Large groups, strict measures and flat heads.} With seven query heads per KV head (Qwen2.5), we see no speedup. The output-norm measure is slower than dense attention at 32K and below. Because the grid starts at the largest weight, 0.013\% of head-steps stop at its first level and read 3.9--8.1\% of the keys at 128K. Starting the grid from the 4-bit bound on $D$ lowers this only to 3.4--5.7\% (Appendix F.3).

\textbf{Reproducibility and fixed perturbations.} End-to-end outputs are bitwise reproducible for a given CPU model and thread count. Perturbations fixed per KV head, which Theorem 4 does not cover in generation, make whole steps no faster, score like fresh ones, and have 1.4--1.8 times their KL divergence on PG-19 (Appendix F.4).

\textbf{Scope.} We evaluate only on CPUs and compare baselines by the bytes they read rather than by time. With 48--300 samples per setting, the end-to-end intervals exclude only large losses (Appendix F.5).

\section{Conclusion}\label{sec:8}

Perturbing the score of every key with an exponential variable turns threshold retrieval into weight-proportional sampling, so a sparse attention kernel can estimate its output without bias and certify the error of each call. On real KV caches the certificates held in every configuration we ran and removed the heavy error tail of heuristic sparse attention. End to end, at equal bytes, the strongest baselines score as well as our method. The linear scan that generates candidates sets a floor on the bytes read, which a sublinear index with a recall guarantee would remove.

\section*{Impact Statement}

This work makes long-context inference cheaper, which lowers the cost and energy of serving language models for all of their uses, beneficial or harmful alike. It adds no new capability and uses only publicly released models and datasets, with no human subjects or personal data. Our guarantee is specific. For each head and decoding step, the error of the attention output is at most $\varepsilon$ on its scale with probability at least $1-\delta$, for calls whose query, keys and values are independent of the perturbations. It is not a guarantee on the model's outputs, and our theory does not cover closed-loop generation with fixed perturbations (Section 7). Deployments should not present it as a guarantee of unchanged model behavior.

\bibliographystyle{plainnat}
\bibliography{refs}

\clearpage
\appendix
\section{Proofs}\label{app:A}

Throughout, fix a head, a query and a threshold $\tau>0$. The window $R$, the heavy set $B=B(\tau)$, the light set $A=A(\tau)$ and $c^\ast$ are deterministic. For $i\in A$, the indicators $X_i=\mathbb 1\{i\in S(\tau)\}$ are independent Bernoulli variables with means $p_i=w_i/\tau$. We write $D_A=\sum_A w_i$, $G_A(c)=\sum_A w_i\|v_i-c\|^2$, $Q=\sum_i w_i\|v_i\|^2=DM^2$, and $V_{\max}=\max_{i\in P}\|v_i\|$.

\subsection{Concentration tools}\label{app:A.1}

\begin{stmt}{Lemma A.1 (Poisson domination and the KL interval).}
Let $k=\sum_A X_i$ and $\mu=\mathbb E k=D_A/\tau$. For every real $\lambda$, $\mathbb E e^{\lambda k}\le\exp(\mu(e^\lambda-1))$. Consequently, with $\mathrm{kl}(x\|\mu)=x\ln(x/\mu)-x+\mu$, $\Pr[k\ge x]\le e^{-\mathrm{kl}(x\|\mu)}$ for $x\ge\mu$ and $\Pr[k\le x]\le e^{-\mathrm{kl}(x\|\mu)}$ for $x\le\mu$. Hence the set $\{\mu':\mathrm{kl}(k\|\mu')\le\ln(2/\delta')\}$ contains $\mu$ with probability at least $1-\delta'$.

\end{stmt}
\begin{proof}
$\mathbb E e^{\lambda k}=\prod_A(1+p_i(e^\lambda-1))\le\prod_A e^{p_i(e^\lambda-1)}$. Chernoff's method with the optimal $\lambda$ gives the two tail bounds, which are the Poisson tails. The event $\mathrm{kl}(k\|\mu)>\ln(2/\delta')$ is contained in $\{k\ge x_+\}\cup\{k\le x_-\}$, where $x_\pm$ are the two solutions of $\mathrm{kl}(x\|\mu)=\ln(2/\delta')$, and each has probability at most $\delta'/2$.
\end{proof}

\begin{stmt}{Lemma A.2 (Bernstein bounds used by the certificates).}
(i) Let $Y_i\in[0,b]$ be independent with $\sum\mathrm{Var}\,Y_i\le b\,\mathbb E\sum Y_i$. Write $\theta=\mathbb E\sum Y_i$, $\hat\theta=\sum Y_i$ and $a=\sqrt{2bL}$. Define the upper confidence bound $\theta^+=\big(\tfrac12(a+\sqrt{a^2+4(\hat\theta+bL/3)})\big)^2$ and the lower confidence bound $\theta^{\rm lo}=y^2$, where $y\ge0$ solves $y^2+ay=\max(0,\hat\theta-bL/3)$. Each of the following holds with probability at least $1-e^{-L}$: (a) $\theta\le\theta^+$; (b) $\theta\ge\theta^{\rm lo}$; (c) $\hat\theta\le\theta+\sqrt{2b\theta L}+bL/3$; (d) $\theta-\theta^{\rm lo}\le2a\sqrt\theta+\tfrac23bL$. Statements (a) and (d) hold on the same lower-tail event, and (b) and (c) on the same upper-tail event.

(ii) Let $Z_i$ be independent zero-mean random vectors in a Hilbert space with $\|Z_i\|\le B$ and $\sum\mathbb E\|Z_i\|^2\le\sigma^2$. By the Bernstein-type inequality of Pinelis (1994), $\Pr[\|\sum Z_i\|\ge r]\le2\exp\!\big(-r^2/(2\sigma^2+\tfrac23Br)\big)$, hence $\Pr\big[\|\sum Z_i\|\ge\sqrt{2\sigma^2L}+\tfrac23BL\big]\le2e^{-L}$.

\end{stmt}
\begin{proof}
(i) Since $Y_i\ge0$ and $\mathbb EY_i\le b$, Bernstein's inequality applies to both tails. The lower tail gives $\hat\theta\ge\theta-\sqrt{2b\theta L}-bL/3$, and the upper tail gives (c).

For (a), the lower-tail inequality is a quadratic inequality in $\sqrt\theta$, whose solution is $\sqrt\theta\le\sqrt{\theta^+}$.

For (b), the upper-tail inequality gives $\theta+a\sqrt\theta\ge\hat\theta-bL/3=y^2+ay$ when the right-hand side is positive. Since $x\mapsto x^2+ax$ is increasing, $\sqrt\theta\ge y$.

For (d), let $s=\sqrt\theta$ and suppose $y<s$. The lower tail gives $y^2+ay\ge s^2-as-\tfrac23bL$, so $s^2-y^2\le a(y+s)+\tfrac23bL\le2as+\tfrac23bL$. If $\hat\theta\le bL/3$, then $y=0$ and $s^2\le as+\tfrac23bL$ directly.

(ii) Solving $r^2/(2\sigma^2+\tfrac23Br)=L$ for $r$ gives $r=\tfrac13BL+\sqrt{\tfrac19B^2L^2+2\sigma^2L}\le\sqrt{2\sigma^2L}+\tfrac23BL$.
\end{proof}

The implementation computes $\hat G^+=\theta^+$ with $Y_i=\tau X_i\|v_i-c\|^2$ and $b=\tau R_c^2$, and $Q_A^{\rm lo}=\theta^{\rm lo}$ with $Y_i=\tau X_i\|v_i\|^2$ and $b=\tau V_{\max}^2$, both at $L=\ln(1/\delta_j)$. Both satisfy the variance condition, since $\mathrm{Var}(\tau X_iy_i)\le\tau w_iy_i^2\le b\,w_iy_i$. Theorem 4 uses (a) and (b), and Proposition 1 uses (c) for $\hat G$ and (d) for $\hat Q_A$.

\subsection{Proof of Theorem 1}\label{app:A.2}

For $i\in A$, $\mathbb E[\tau X_i]=w_i$ and $\mathbb E[(\tau X_i-w_i)^2]=\tau^2p_i-w_i^2=w_i(\tau-w_i)$. Since the $X_i$ are independent, cross terms vanish. Applying this to the vector $v_i-c$ gives (a) and (c), and (b) is the scalar case. For (d), every $i\in B$ has $w_i\ge\tau$, so $|B|\le D_B/\tau$, and $\mathbb Ek=D_A/\tau$. \hfill$\blacksquare$

\subsection{Proof of Lemma 1}\label{app:A.3}

$Z_c-(\hat o-c)(\hat D-D)=\hat N-N-c(\hat D-D)-(\hat o-c)(\hat D-D)=\hat N-N-\hat o(\hat D-D)=\hat oD-N$, using $\hat N=\hat o\hat D$. Dividing by $D$ gives the identity. \hfill$\blacksquare$

\subsection{Proof of Theorem 2}\label{app:A.4}

Let $\tau=D/m$. The error of the denominator, $\hat D-D=\sum_A(\tau X_i-w_i)$, has terms bounded by $\tau$ and variance at most $\tau D$, so Bernstein's inequality gives $|\hat D-D|\le\sqrt{2\tau DL}+\tau L/3=\eta D$ with probability at least $1-2e^{-L}$. For the numerator, take $c=o$ (for the analysis only). The terms of $Z_o$ have norm at most $\tau R_o$, and $\mathbb E\|Z_o\|^2\le\tau\sum_Aw_i\|v_i-o\|^2\le\tau D\sigma_v^2$, so Lemma A.2(ii) gives $\|Z_o\|\le\sqrt{2\tau D\sigma_v^2L}+\tfrac23\tau R_oL$ with probability at least $1-2e^{-L}$. Lemma 1 with $c=o$ gives $\|\hat o-o\|(1-|\hat D-D|/D)\le\|Z_o\|/D$, and substituting $\tau/D=1/m$ gives the bound on the intersection of the two events, which has probability at least $1-4e^{-L}=1-\delta$.

For the value scale, $\sigma_v\le M$, so $m=2L\varepsilon^{-2}(1+o(1))$ suffices, and the term $R_o/M$ enters only at lower order. For the output norm, $\sigma_v^2=\|o\|^2(\kappa^2-1)$. \hfill$\blacksquare$

\subsection{Proof of Theorem 3}\label{app:A.5}

\emph{Instances.} All $n$ keys are identical, so $a_i=1/n$. Values are $v_i=e_1+\sigma\xi_ie_2$ with $\xi_i\in\{\pm1\}$. Instance $I_+$ has exactly $n(1+\eta)/2$ indices with $\xi_i=+1$ and $I_-$ has $n(1-\eta)/2$, at uniformly random positions. Then $o_\pm=e_1\pm\sigma\eta e_2$, $M^2=1+\sigma^2$ and $\|o_+-o_-\|=2\sigma\eta$.

\emph{Adaptivity does not help.} The keys and weights carry no information about $\xi$. Given the history of reads, each unread position is a uniform draw from the remaining multiset of values, so whatever the reading policy, the observed sequence is a uniform sample without replacement. Its total variation distance from a sample with replacement is at most the probability of a repeat in $m$ draws with replacement, which is at most $m^2/(2n)\le\delta/2$ for $n\ge m^2/\delta$.

\emph{Two-point argument.} Under sampling with replacement, $\mathrm{KL}(P_+^m\|P_-^m)=m\eta\ln\frac{1+\eta}{1-\eta}\le2.2m\eta^2$ for $\eta\le1/2$. Let $\mathcal A=\{\|\hat o-o_+\|\ge\sigma\eta\}$. By the triangle inequality, $\mathcal A^c\subseteq\{\|\hat o-o_-\|>\sigma\eta\}$. By the Bretagnolle--Huber inequality, $P_+(\mathcal A)+P_-(\mathcal A^c)\ge\tfrac12e^{-\mathrm{KL}}$. If the algorithm fails with probability at most $\delta$ on both instances, then $2\delta\ge\tfrac12e^{-2.2m\eta^2}-m^2/n\ge\tfrac12e^{-2.2m\eta^2}-\delta$, that is, $m\ge\ln(1/(6\delta))/(2.2\eta^2)$.

\emph{Output norm (i).} Let the target radius be $\varepsilon'=\varepsilon\|o\|$, and choose $\eta=2\varepsilon'/\sigma$ so that $\sigma\eta=2\varepsilon'>\varepsilon'$. Then $\|o\|^2=1+4\varepsilon'^2$, hence $\varepsilon'\le1.16\varepsilon$ for $\varepsilon\le1/4$. With $\sigma^2=\kappa^2(1+4\varepsilon'^2)-1$, both instances have cancellation factor exactly $\kappa$, and the condition $\eta\le1/2$, which needs $\sigma\ge4\varepsilon'$, follows from $\kappa^2\ge1+22\varepsilon^2$. Since $\sigma^2/\|o\|^2\ge\kappa^2-1$, the bound becomes $m\ge\sigma^2\ln(1/(6\delta))/(8.8\varepsilon^2\|o\|^2)\ge(\kappa^2-1)\ln(1/(6\delta))/(8.8\varepsilon^2)$. When $n(1\pm\eta)/2$ is not an integer, rounding changes the constants slightly.

\emph{Value scale (ii).} Take $\sigma=1$, so $\varepsilon'=\varepsilon M=\sqrt2\varepsilon$ and $\eta=2\varepsilon'\le1/2$ for $\varepsilon\le0.17$. Then $m\ge\ln(1/(6\delta))/(2.2\cdot8\varepsilon^2)$. \hfill$\blacksquare$

\subsection{Proof of Lemma 2}\label{app:A.6}

The 4-bit and 8-bit upper bounds satisfy $u_{im}\ge s_{im}$, by the dequantization bound plus the bound on the integer residual (Appendix B). The level map $x\mapsto\lceil-r(x+E_i)/\ln2\rceil$ is nonincreasing in $x$, so the candidate level computed from an upper bound is at most the true inclusion level $\lceil-r(s_{im}-c^\ast_m+E_i)/\ln2\rceil$. A key is therefore examined, refined and, if it is still a candidate, scored exactly no later than the level at which it enters $S$. For the maximum, the key attaining it satisfies $u\ge s=c^\ast\ge\max_i(\text{lower bound}_i)$, so it is among the keys scored exactly. \hfill$\blacksquare$

\subsection{Proof of Theorem 4}\label{app:A.7}

\emph{Good event.} By Lemma 2, the quantities computed at level $j$ are exactly the estimator of Theorem 1 at threshold $\tau_j$. At each level we take the events $\mu_j\in[\mu_{\rm lo},\mu_{\rm hi}]$ (Lemma A.1), $G_A(c_j)\le\hat G_j^+$ (Lemma A.2(i)(a)) and $\|Z_{c_j}\|\le\zeta_j$ (Lemma A.2(ii), which applies because $c_j$ is determined by $\tau_j$), and for the value scale also $Q_A\ge Q_A^{\rm lo}$ (Lemma A.2(i)(b)). The target uses $e$ of these events, each failing with probability at most $\delta_j=\delta/((J+1)e)$. By a union bound over levels and events, the intersection $\mathcal G$ has probability at least $1-\delta$.

\emph{On $\mathcal G$, every stopping rule implies its target.} We have $|\hat D-D|=\tau|k-\mu|\le dD$. Lemma 1 with $c=c_j$ gives $\|\hat o-o\|\le(\|Z_{c_j}\|+\|\hat o-c_j\|\,|\hat D-D|)/D\le\beta$, using $D\ge\hat D-dD$. For the denominator, $dD\le\varepsilon(\hat D-dD)\le\varepsilon D$. For the output norm, $\|o\|\ge\|\hat o\|-\beta\ge\|\hat o\|/(1+\varepsilon)$, so $\beta\le\varepsilon\|\hat o\|/(1+\varepsilon)\le\varepsilon\|o\|$. For the value scale, $M^2=(Q_R+Q_B+Q_A)/D\ge M_{\rm lo}^2$, so $\beta\le\varepsilon M_{\rm lo}\le\varepsilon M$.

\emph{The stopping level.} It depends on the data, but $\mathcal G$ covers all levels simultaneously, and the fallback is exact attention, which has zero error. \hfill$\blacksquare$

\subsection{Proof of Proposition 1}\label{app:A.8}

Let $V=\max(\max_{i\in P}\|v_i\|,\max_{i\in R}\|v_i\|)$, $\nu=V/M$, $L=\ln(2/\delta_j)$ and, for a level with threshold $\tau$, $t=\tau L/D$. The known set $K=R\cup B(\tau)$ has $D_K\ge1$, because the key attaining the maximum has weight $1\ge\tau$, and $c=N_K/D_K$ is a convex combination of known values, so $\|c\|\le V$, $R_c\le2V$, and $\|v_i-c\|\le2V$ for $i\in A$.

The implementation uses $L_G=L_Q=\ln(1/\delta_j)\le L$. Since $\hat G^+$ increases and $Q_A^{\rm lo}$ decreases in its logarithm, replacing both by $L$ only weakens the bounds we prove.

\emph{Event.} At every level we take four events: (F1) $|k-\mu|\le\sqrt{2\mu L}+L/3$, which fails with probability at most $2e^{-L}$; (F2) $\hat G\le G+\sqrt{2bGL}+bL/3$ with $G=G_A(c)$ and $b=\tau R_c^2$ (Lemma A.2(i)(c)), failing with probability at most $e^{-L}$; (F3) $\|Z_c\|\le\sqrt{2\tau GL}+\tfrac23\tau R_cL$, failing with probability at most $2e^{-L}$; and, for the value scale, (F4) the lower-tail event of Lemma A.2(i) for $\hat Q_A$ with $b_q=\tau V_{\max}^2$, on which (d) holds, failing with probability at most $e^{-L}$.

Each level fails with probability at most $6e^{-L}=3\delta_j$, or $2.5\delta_j$ for the output norm. Over $J+1$ levels this gives $3\delta/4$ (value scale, $e=4$) or $5\delta/6$ (output norm, $e=3$), so the event $\mathcal G_2$ has probability at least $1-\delta$. On $\mathcal G_2$, consider any level with $t\le1/400$.

\emph{Denominator.} (F1) gives $k\le2\mu+L$. For $\mu'$ in the KL interval, $\mathrm{kl}(k\|\mu')\ge(k-\mu')^2/(2\max(k,\mu'))$, by the integral form of the remainder with second derivative $1/x$. Hence $k-\mu_{\rm lo}\le\sqrt{2Lk}$, and $x=\mu_{\rm hi}-k$ satisfies $x^2\le2L(k+x)$, i.e. $x\le2L+\sqrt{2Lk}$. So $dD\le\tau(\sqrt{2Lk}+2L)\le2\sqrt{\tau LD_A}+3.5\tau L\le D(2\sqrt t+3.5t)$ and $|\hat D-D|\le D(\sqrt{2t}+t/3)$. With $\sqrt t\le1/20$, we get $dD\le0.109D$ and $|\hat D-D|\le0.072D$, so $\hat D-dD\ge0.8D$ and $\hat D+dD\le1.2D$.

\emph{The bounds $\hat G^+$ and $\zeta$.} We have $G\le4V^2D$ and $b\le4\tau V^2$, and (F2) gives $\hat G\le2G+bL$. With $(x+y)^2\le2x^2+2y^2$, $\hat G^+\le2bL+2\hat G+\tfrac23bL\le4G+5bL\le16V^2D+20\tau V^2L$. Hence $\zeta=\sqrt{2\tau\hat G^+L}+\tfrac23\tau R_cL\le V(5.66\sqrt{\tau LD}+7.66\tau L)$.

\emph{The distance $\|\hat o-c\|$.} $\hat o=(D_Kc+\tau\sum_{S\cap A}v_i)/(D_K+\tau k)$ is a convex combination of $c$ and sampled light values, so $\|\hat o-c\|\le2V$ deterministically, and $\|\hat o-c\|\,dD\le V(4\sqrt{\tau LD}+7\tau L)$.

\emph{The bound $\beta$.} Adding the two parts, $\beta\le V(9.66\sqrt{\tau LD}+14.66\tau L)/(0.8D)=V(12.08\sqrt t+18.33t)$, and since $t\le\sqrt t/20$, $\beta\le13V\sqrt t$.

\emph{Value scale.} By Lemma A.2(i)(d) on (F4), $Q_A-Q_A^{\rm lo}\le2\sqrt{2b_qLQ_A}+\tfrac23b_qL$. Hence $Q_K+Q_A^{\rm lo}\ge Q-2\sqrt{2\tau V^2LQ}-\tfrac23\tau V^2L$, which is at least $DM^2(1-2.83\nu\sqrt t-0.67\nu^2t)$. If $\nu\sqrt t\le1/20$, then $M_{\rm lo}^2\ge0.857M^2/1.2\ge0.71M^2$, i.e. $M_{\rm lo}\ge0.8M$, and the rule $\beta\le\varepsilon M_{\rm lo}$ holds whenever $13V\sqrt t\le0.8\varepsilon M$, i.e. $\sqrt t\le\varepsilon/(16.25\nu)$. All conditions hold when $t\le\varepsilon^2/(400\nu^2)$, i.e. $\tau\le\tau^\dagger=\varepsilon^2D/(400\nu^2L)$.

\emph{Output norm.} By Lemma 1, (F1) and (F3), $\|\hat o-o\|\le(\|Z_c\|+\|\hat o-c\|\,|\hat D-D|)/D\le V(5.66\sqrt t+2.01t)\le6V\sqrt t$. With $\varepsilon/(1+\varepsilon)\ge\varepsilon/2$, the rule holds if $13V\sqrt t\le(\varepsilon/2)(\|o\|-6V\sqrt t)$. It suffices that $16V\sqrt t\le\varepsilon\|o\|/2$, i.e. $\sqrt t\le\varepsilon/(32\kappa\nu)$, which holds for $\tau\le\varepsilon^2D/(1024\kappa^2\nu^2L)$.

\emph{Stopping level and reads.} The descent checks levels from $j=0$, so on $\mathcal G_2$ it stops at $j^\dagger=\min\{j:\tau_j\le\tau^\dagger\}$ or earlier, and $\tau_{j^\dagger}\ge\min(1,2^{-1/r}\tau^\dagger)$. Samples are nested, so $|S(\tau_{\rm stop})\setminus R|\le|B(\tau_{j^\dagger})|+k_{j^\dagger}\le D_B/\tau_{j^\dagger}+2D_A/\tau_{j^\dagger}+L\le2D/\tau_{j^\dagger}+L$. Intersecting with the event of Theorem 4 gives the combined statement with probability at least $1-2\delta$. \hfill$\blacksquare$

\textbf{Remarks.} The factor $\nu^2$ arises twice: in bounding the spread of the light values around $c$ by $4V^2$ and $\|\hat o-c\|$ by $2V$, and in the Bernstein range $\tau V_{\max}^2$ of the lower bound on $M$. If $G_A(c)\le\gamma DM^2$ and the light keys carry at most half of $Q$, the same argument gives a leading term $\tau^\dagger\propto\varepsilon^2D/(\gamma L)$.

On real KV caches of five settings (two samples each, every fourth layer, decoding steps 1, 16 and 31), $\nu$ has a median of 2.6--4.1 and a 90th percentile of 5.7--12.4. On 3,616 heads (Llama RULER 32K and Qwen2.5 LongBench 16K, $\varepsilon\in\{0.1,0.3\}$), every stopping threshold observed in the kernel lies above $2^{-1/r}\tau^\dagger$, with a median ratio of 460--2,170, so the constants are far from tight.

\subsection{Proposition 2: approximate candidate generation}\label{app:A.9}

Proposition 2 is the interface for replacing our linear scan with a sublinear index. Our scan has $\eta=0$ by Lemma 2.

\begin{stmt}{Proposition 2 (Approximate candidate generation).}
Suppose candidate generation misses a set $U_j\subseteq S(\tau_j)$. Then $\hat D$ is replaced by $\hat D-\sum_{U_j}\max(w_i,\tau_j)$, a downward bias of $\mathbb E\sum_{U_j}\max(w_i,\tau_j)$. If, in expectation, the retriever misses at most an $\eta$ fraction of the pool's Horvitz--Thompson mass, the bias is at most $\eta D$.

\end{stmt}
\begin{proof}[Proof of Proposition 2]
Missing $U_j$ removes $\sum_{U_j}\max(w_i,\tau_j)$ from $\hat D$: heavy keys contribute $w_i$ and sampled light keys contribute $\tau_j$. Taking expectations gives the bias formula. The pool part of the estimate, $\hat D_P=D_B+\tau_jk$, satisfies $\mathbb E\hat D_P=D_P$ by Theorem 1. If the expected missed mass is at most $\eta\,\mathbb E\hat D_P=\eta D_P$, the bias is at most $\eta D_P\le\eta D$.
\end{proof}

A certificate in that setting additionally needs the recall condition to hold with high probability and the confidence intervals to account for the missed count. We leave this to future work.

\section{Implementation Details}\label{app:B}

The kernel is about 1,300 lines of C++ with AVX-512 intrinsics and a pybind11 binding. One \texttt{KVHead} object holds the cache and index of one KV head, and a call \texttt{query\_many} answers the queries of all heads of a decoding step in parallel. A line-by-line NumPy reference implementation with exact scores serves as the test oracle.

Algorithm 1 states one query; Figure 1(b) shows the same flow.

\begin{center}\fbox{\begin{minipage}{0.96\linewidth}\small
\textbf{Algorithm 1} (one query of a GQA group with heads $m=1..G$)
\begin{enumerate}
\item
4-bit VNNI scan of the pool $\rightarrow$ upper and lower score bounds per key and head.
\item
Exact maxima $c^\ast_m$ from keys whose upper bound exceeds the best lower bound; window terms $D_R,N_R,Q_R$ exactly.
\item
Candidate level $j^c_i$ for every pool key; bucket by level.
\item
\textbf{For} $j=0,1,\dots,J$:
\begin{itemize}
\item
refine the candidates of level $j$ with 8-bit codes, and score the survivors exactly;
\item
\textbf{for} each head that has not stopped: add the newly sampled keys to $B$ or $A$ (moving keys from $A$ to $B$ once $w_i\ge\tau_j$), update $\hat D,\hat N,dD,\beta$ (and $M_{\rm lo}$); if the stopping rule holds, output $\hat o_m$ and stop this head.
\item
\textbf{if} all heads have stopped, \textbf{break}.
\end{itemize}
\item
Heads that never stopped: exact attention.
\end{enumerate}
\end{minipage}}\end{center}

\subsection{Data layout and storage}\label{app:B.1}

Each KV head keeps the following arrays (for $d=128$ and blocks of 128 tokens):

\begin{table}[tbp]
\centering\small
\begin{tabularx}{\linewidth}{>{\raggedright\arraybackslash}X>{\raggedright\arraybackslash}X>{\raggedright\arraybackslash}X>{\raggedright\arraybackslash}X}
\toprule
Array & Layout & Bytes per token & Read per query \\
\midrule
Keys and values & fp16, token-major (256 B per row) & 512 & exact keys of candidates; values of sampled keys; window \\
4-bit codes & VNNI-interleaved per block (below) & 64 & every pool token \\
8-bit codes & token-major, one byte per channel & 128 & refined candidates \\
Perturbation $E_i$ & fp32, rewritten at every step & 4 & every pool token (and written before the scan) \\
Block scales ($\mathrm{lo}$, $\mathrm{step}$, $\mathrm{step}^{(8)}$ per channel) & fp32, per block & 12 & every pool token \\
Total &  & 720 (+40\%) & floor 80 B read + 4 B written per pool token (16.4\%) \\
\bottomrule
\end{tabularx}
\end{table}

In the 4-bit layout a block occupies $64d$ bytes. For channel group $g$ (four channels) and segment $m$, one 64-byte line holds, in byte $4t+k$, channel $4g+k$ of token $16m+t$ in the low nibble and of token $64+16m+t$ in the high nibble. One 64-byte load, an AND and a shift then yield two \texttt{vpdpbusd} operands of 16 tokens $\times$ 4 channels each.

Recomputing the perturbations inside the scan, instead of writing them before it, would remove the array and lower the floor to 76 bytes per token (14.8\%). Our byte counts compared with baselines and quantized caches (Tables D.2c, D.2e, D.9d and D.12b) include the 4 bytes written per pool token; the ablation tables of our method (D.0, D.3--D.8) report bytes read, to which the write adds 0.78 points. Dropping the 8-bit level (Appendix D.6 shows that it barely helps) lowers storage to 592 bytes per token (+16\%).

\subsection{Deterministic score bounds}\label{app:B.2}

Let the query $q$ include the factor $1/\sqrt d$. In a block, channel $c$ of key $i$ is stored as a code $x_{ic}\in\{0,\dots,15\}$ with $|k_{ic}-(\mathrm{lo}_c+\mathrm{step}_c\,x_{ic})|\le\mathrm{step}_c/2$, where $\mathrm{lo}_c$ is the channel minimum in the block and $\mathrm{step}_c=(\max_c-\mathrm{lo}_c)/15$.

\textbf{Two-level query weights.} Per block, the weights $q_c\,\mathrm{step}_c$ are written as
\[q_c\,\mathrm{step}_c=d_1\alpha_c+d_2\beta_c+r_c,\qquad \alpha_c,\beta_c\in\{-127,\dots,127\},\quad d_2=d_1/254,\quad |r_c|\le d_2/2 .\]
The integer dot products $\sum_c\alpha_cx_{ic}$ and $\sum_c\beta_cx_{ic}$ are computed exactly with \texttt{vpdpbusd} (unsigned codes times signed bytes). With $\hat s_i=\sum_cq_c\mathrm{lo}_c+d_1\sum_c\alpha_cx_{ic}+d_2\sum_c\beta_cx_{ic}$ and $x_{ic}\in[0,15]$,
\[u_i=\hat s_i+15\sum_c\max(r_c,0)+\tfrac12\sum_c|q_c|\,\mathrm{step}_c+\sigma,\qquad \ell_i=\hat s_i+15\sum_c\min(r_c,0)-\tfrac12\sum_c|q_c|\,\mathrm{step}_c-\sigma\]
satisfy $\ell_i\le s_i\le u_i$. The margin $\sigma=10^{-5}\sum_c|q_c|(|\mathrm{lo}_c|+15\,\mathrm{step}_c)+10^{-6}/\sqrt d$ covers rounding in dequantization, in the conversion of the integer sums and in the scale factors.

\textbf{8-bit refinement.} The 8-bit codes use $\mathrm{step}^{(8)}_c=(\max_c-\mathrm{lo}_c)/255$ and are dequantized to fp32. The bound is the dequantized score plus $\tfrac12\sum_c|q_c|\mathrm{step}^{(8)}_c+\sigma$.

\textbf{Candidate levels.} Key $i$ enters $S(\tau_j)$ for head $m$ exactly when $j\ge-r(s_{im}-c^\ast_m+E_i)/\ln2$. The kernel computes $j^c_i=\lceil-r(\max_m(u_{im}-c^\ast_m)+E_i)/\ln2-10^{-6}\rceil$ in double precision, so that $j^c_i$ never exceeds the true inclusion level of any head of the group (Lemma 2). Keys are bucketed by $j^c_i$ with a counting sort that is expanded lazily, one segment of levels at a time, as the descent reaches it.

\subsection{Arithmetic and memory access}\label{app:B.3}

\begin{itemize}
\item
\textbf{Scan.} The quantization of the query weights is branch-free AVX-512. The scan kernel is a compile-time template over groups of four query heads, with 16 int32 accumulators held in registers.
\item
\textbf{Tiled scoring.} Exact scores (from fp16 keys) and 8-bit refinements are computed in tiles of 4 keys $\times$ 4 heads. Each channel block of a key is converted once and multiplied with four queries, and a transpose-and-add tree produces 16 dot products at once.
\item
\textbf{Exponentials.} A vectorized exp uses the Cephes range reduction and polynomial and restores $2^n$ with \texttt{vscalefps} (about 1 ulp).
\item
\textbf{Prefetching.} Random-row accesses to exact keys, 8-bit codes and sampled values are software-prefetched four rows ahead.
\item
\textbf{Dense baseline.} Our dense kernel uses the same tiling and exp, and keeps register-blocked accumulators of 4 heads $\times$ 64 channels for the values. Single-threaded, it is not slower than any PyTorch CPU path we measured (fp32 matmul, and scaled-dot-product attention in fp32, fp16 and bf16), and 2.6--3.7 times faster than fp32 scaled-dot-product attention.
\item
\textbf{Confidence intervals.} The Poisson-KL interval is solved by Newton's method and finished by bisection, rounding outward.
\end{itemize}

\subsection{Threading and index maintenance}\label{app:B.4}

\texttt{query\_many} schedules KV heads dynamically over OpenMP threads. Each thread answers all $G$ query heads of one KV head, and threads share no writable state. With per-step seeds, the task of each KV head first rewrites the perturbations of its pool from the step's seed, with a vectorized splitmix64 hash and logarithm (the draws equal those of the scalar code; Appendix B.5). When a token is appended, the kernel writes its fp16 key and value and its perturbation, and when a block fills, it also writes the block's 4-bit and 8-bit codes. At the first decoding step, \texttt{append\_list} indexes the prompt of all layers' KV heads in parallel from views of the model's KV cache, and reserves 1/8 extra capacity so that later appends do not copy the cache. Appendix D.10 reports the costs.

\subsection{Correctness tests}\label{app:B.5}

\begin{itemize}
\item
\textbf{Hashing and dense attention.} The hashed perturbations match the NumPy implementation bit for bit, and the vectorized per-step recomputation matches the scalar hash on more than $10^8$ draws. The dense kernel's maximum absolute error is $1.3\times10^{-7}$.
\item
\textbf{Real KV caches.} We compare the kernel with the reference with exact scores on Llama-3.1-8B, with decoded tokens appended after prefill (layers 0/15/31 $\times$ KV heads 0/3/7 $\times$ decoding steps 0/63 $\times$ 4 query heads $\times$ 3 measures, 216 queries). The two stop at the same level in every query, and their outputs differ by at most $1.2\times10^{-6}$ relative. Since the reference does not use candidate generation, this also checks that candidate generation misses no key.
\item
\textbf{Score bounds.} On the same data, the smallest margin $u_i-s_i$ of the 4-bit bound is 0.37 > 0, and the median margin is 1.82 (in units of the log-weight).
\item
\textbf{Unbiasedness.} On synthetic peaked data at fixed levels 8, 16 and 24 (400 seeds), the largest $|z|$ of $\hat D$ is 1.67, 0.29 and 0.82. The fraction of coordinates of $\hat N$ with $|z|>3$ is 0.98\%, 0.59\% and 0.20\%. It exceeds the normal value of 0.27\% only at level 8, where few samples make the estimate skewed, and it decreases with depth. No systematic bias is visible.
\item
\textbf{Coverage under stress.} On five synthetic datasets (300 seeds $\times$ 4 heads, $\varepsilon=0.1$, $\delta=0.05$), no certified estimate violates its measure:
\end{itemize}

\begin{table}[tbp]
\centering\small
\begin{adjustbox}{max width=\linewidth}\begin{tabular}{llll}
\toprule
Data & $\kappa$ & Violations: denominator / $\|o\|$ / $M$ & Keys read: denominator / $\|o\|$ / $M$ \\
\midrule
flat & 58.9 & 0 / 0 / 0 & 84.5\% / 100\% / 98.0\% \\
peaked & 8.4 & 0 / 0 / 0 & 11.6\% / 38.6\% / 14.8\% \\
heavy-tailed & 1.6 & 0 / 0 / 0 & 3.6\% / 4.5\% / 4.1\% \\
cancelling values & 65.5 & 0 / 0 / 0 & 37.7\% / 97.8\% / 43.8\% \\
short (600 tokens) & 14.6 & 0 / 0 / 0 & 70.4\% / 99.3\% / 78.3\% \\
\bottomrule
\end{tabular}\end{adjustbox}
\end{table}

On the cancelling data, the value-scale measure reads 45\% of the keys of the output-norm measure, as the $\kappa^2$ dependence of Theorems 2 and 3 predicts. Flat attention cannot be sparsified because the grid starts at $\tau=1$, relative to the largest weight, so that $\mathbb E|S|\ge D\approx n$. This does not occur on real data, where the largest ratio $D_P/n_P$ over 5,120 query heads is 0.0099 (99th percentile 0.0015). Extending the grid above $\tau=1$ would remove this limitation at the cost of about $\log n$ further levels.
\begin{itemize}
\item
\textbf{Index construction.} Indexing through \texttt{append\_list} reproduces the outputs and read counts of the per-head append path bit for bit (180 output arrays).
\end{itemize}

\subsection{Design choices and negative results}\label{app:B.6}

Single-threaded, the kernel answers one KV head with four query heads at 80K tokens in 2.82 ms (value scale, $\varepsilon=0.1$), against 6.61 ms for our dense kernel. Of this, the 4-bit scan takes 1.01 ms, the candidate levels 0.07 ms and the level loop 1.48 ms. We tried and abandoned three alternatives:
\begin{itemize}
\item
\textbf{Switching to a sequential 8-bit scan when many candidates remain.} Bytes read rose to 49--61\%, and the query became slower.
\item
\textbf{A random Hadamard rotation before quantization.} It widened the 4-bit bounds by 1.3--2.4 times.
\item
\textbf{One int8 level for the query weights.} It makes the scan about 30\% faster but loosens the bound. The kernel then refines 12--40\% more keys and reads 0.5--2.1 more percentage points of bytes, and the total time is about the same. Two levels remain the default.
\end{itemize}

\section{Experimental Details}\label{app:C}

\subsection{Captured KV caches and queries}\label{app:C.1}

For each sample we ran a bf16 prefill on a Xeon 8555C, decoded 32 tokens greedily, and recorded the keys, the values and the decoding queries of every layer. Operator-level results use decoding steps 0, 8, 16, 24 and 31. Each measured query is one (sample, layer, query head, step), and the ten settings contain the following numbers of queries:

\begin{table}[tbp]
\centering\small
\begin{adjustbox}{max width=\linewidth}\begin{tabular}{lllllll}
\toprule
Model & LongBench 16K & RULER 32K & RULER 64K & LongBench-v2 64K & RULER 128K & LongBench-v2 128K \\
\midrule
Llama-3.1-8B-Instruct & 61,440 & 40,960 & 40,960 & 20,480 & 40,960 & 20,480 \\
Qwen3-8B & 69,120 & 46,080 & -- & -- & -- & -- \\
Qwen2.5-7B-Instruct & 47,040 & 31,360 & -- & -- & -- & -- \\
\bottomrule
\end{tabular}\end{adjustbox}
\end{table}

In total there are 418,880 queries per method configuration. Versions and checksums of all code, data and weights are listed in Appendix E.2.

\begin{itemize}
\item
\textbf{LongBench} uses the official prompts and truncation.
\item
\textbf{RULER} data were generated with the official scripts (seed 42) for each model's tokenizer.
\item
\textbf{LongBench-v2} uses the official data file and the official zero-shot prompt, with the document truncated in the middle to the target length.
\item
\textbf{Qwen3} was captured in its default thinking mode. The end-to-end runs use the non-thinking mode, because the LongBench tasks allow only 32--128 new tokens. The query distributions of the two parts therefore differ, but this does not affect the validity of the operator-level certificates.
\end{itemize}

\subsection{Baselines and tuning}\label{app:C.2}

We prefix all sparse-attention-hub maskers with the static part that the hub's own benchmark builders use, namely 128 sink tokens and 128 local tokens. Grids follow the hub's search spaces, and the method papers' recommended settings where the hub's grid is empty or degenerate. Top-$k$-style budgets beyond sink and local range over $\{0.5, 1, 2, 5, 10, 20\}\%$ of the keys.

\begin{itemize}
\item
\textbf{HashAttention} and vAttention on top of HashAttention use the authors' released Llama-3.1-8B weights. No weights exist for the Qwen models, which therefore have 13 families.
\item
\textbf{MagicPIG and Quest reference ports} follow the authors' code.
\item
\textbf{RetroInfer} requires CUDA, so we ported its estimator to the CPU from the authors' GPU code to measure its accuracy. The port reproduces:
\begin{itemize}
\item
its index: segmented k-means with about 16 tokens per cluster;
\item
its three zones: an exact steady zone, a retrieval zone of top-ranked clusters read exactly, and an estimation zone whose clusters contribute through their centroids and value sums.

We count as keys read the tokens of the steady and retrieval zones. The estimation zone reads precomputed cluster summaries, which we do not count.
\end{itemize}
\end{itemize}

\textbf{Grids.}
\begin{itemize}
\item
\emph{Full official grid} (356 configurations over 15 families: 330 for the hub maskers and the RetroInfer port, 26 for the MagicPIG and Quest reference ports): Llama LongBench 16K and Llama RULER 32K.
\item
\emph{Reduced grid} (64 configurations): the other eight settings. We chose it on Llama LongBench 16K, log-spaced along each family's Pareto front and around our number of keys read.
\item
\emph{Sparse extension} (44 configurations reading 0.1--0.3\% of the keys) at 64K--128K. The reduced grid's sparsest configurations read 0.4--0.9\% there, more than our method. The extension extrapolates each family's existing parameters to smaller budgets and was not tuned separately at 128K.
\end{itemize}

\textbf{Selection rule.} For Table 1, each family contributes its configuration with the lowest 90th-percentile error among those reading at most 1.1 times our keys. Appendix D.2 relaxes the budget to 4 times.

\textbf{Bytes.} Our bytes are those the kernel reads. For the baselines we only record keys read per query head, so each configuration is charged a lower bound. Every key whose value is read also has its 16-bit key read, and the union over the query heads of a GQA group is ignored. To this we add the metadata the method must scan for every key: page labels for Quest, PQ codes for PQCache, one 32-bit signature per KV head for HashAttention, bucket ids for SOCKET, centroid keys and value sums for RetroInfer, and the inverted lists that a hash-table implementation of MagicPIG probes. Oracle methods and SANTA score every key and read at least 50\%. The denser configurations are PQCache with budgets of 15\% and 20\% (RULER 64K and 128K), RetroInfer with retrieval budgets of 10\% and 14.4\% of the clusters, and the MagicPIG port with $(K,L)=(8,300)$ and $(9,300)$ (the four Llama settings at 64K--128K and the four Qwen settings). On the four Qwen settings we also ran the MagicPIG port with $(K,L)\in\{(8,200),(8,250),(9,450),(9,600)\}$ and RetroInfer with retrieval budgets of 20\% and 25\%. For the end-to-end comparison (Appendix C.6) we added RetroInfer with 20--35\% on Llama RULER 32K and Qwen2.5, and vAttention (HashAttention) with top-$k$ budgets of 10--20\% and a base rate of 5\% on Llama RULER 32K.

\textbf{SANTA.} We implemented SANTA and its two variants from the paper's description \citep{lee2026stochastic}. Each computes the exact softmax over all keys and averages $S$ value rows. SANTA draws them i.i.d., S$^2$ANTA-strat draws one row in each of $S$ equal-mass strata of the CDF, and S$^2$ANTA-sys uses one offset shared by all strata, the rule of the paper's GPU kernels. We use 19 sample sizes from 32 to 16,384 and count the distinct value rows as keys read. The randomness comes from a seed per (sample, layer, step, configuration).

\subsection{Memory fixes in the baselines}\label{app:C.3}

PQCache's k-means, the key signatures of MagicPIG (hub), the key hashing of SOCKET (hub) and the hash codes of the MagicPIG reference port each materialize tensors whose size grows with $n$, at 0.2--1 MB per key. At 128K a single job needed 30--60 GB. We changed these four routines to compute in chunks along the key dimension, without changing any arithmetic. On the same job, each modified routine reproduces every result array of the original bit for bit (36, 12, 20 and 20 arrays), and peak memory falls by 6--12 times.

\subsection{Hardware and software}\label{app:C.4}

\begin{table}[tbp]
\centering\small
\begin{adjustbox}{max width=\linewidth}\begin{tabular}{llll}
\toprule
Machine & CPU & Memory & Used for \\
\midrule
A & 2 $\times$ Intel Xeon Platinum 8555C (AVX-512, VNNI, AMX) & 1 TB & KV capture, operator-level runs, timing \\
B--D & 2 $\times$ Intel Xeon Platinum 8358 (AVX-512, VNNI) & 2 TB each & operator-level and end-to-end runs, timing \\
\bottomrule
\end{tabular}\end{adjustbox}
\end{table}

We use no GPU anywhere. Software: Python 3.12.3, PyTorch 2.14.0 (CPU build), Transformers 5.17.0, NumPy 2.5.3, and g++ 13.3.0 with \texttt{-O3 -march=native -fopenmp}. The machines are shared.

\subsection{Timing protocols}\label{app:C.5}

\begin{itemize}
\item
\textbf{Attention only (Table D.11a).} For one sample, we time all layers and KV heads of one decoding step inside \texttt{query\_many}, with the C++ wall clock. Our dense kernel and all configurations are timed in turn within each of three repetitions, and each uses its best repetition. Configurations with fresh perturbations draw new ones in every repetition, inside the timed call. Threads are pinned to physical cores: 26 cores of one NUMA node of the 8555C, with memory on that node, and the 32 cores of one socket of an 8358, whose two sub-NUMA nodes hold the memory interleaved, as within one socket-wide node.
\item
\textbf{Whole decoding step (Table D.11b).} Llama-3.1-8B in fp32 on one RULER sequence (niah\_multikey\_2) per length, with the same threads for all methods (26 threads on one NUMA node of the 8555C, 32 threads on one socket of an 8358, pinned to physical cores):
\begin{itemize}
\item
After one dense prefill, three methods decode exactly 32 tokens each from the prompt's KV cache, one after another in the same process, for five rounds, with the order rotated every round. They are the model's own attention (PyTorch scaled-dot-product attention over its fp32 KV cache), our dense kernel over the same 16-bit keys and values as our method (\emph{dense16}, the same-precision reference), and our method. Every decode starts from a new instance of the method, so the first step indexes the prompt again.
\item
We record the wall time of every step and of every attention call. ``Steady'' is the median of steps 2--32, and each method is reported in its fastest round. The first step includes indexing the prompt, also for dense16, which reads the same index. There is no closed-loop audit.
\item
PyTorch tensors are allocated on transparent huge pages, which does not change any output.
\end{itemize}
\item
\textbf{Idle machines.} Each run started after its machine had been idle for ten minutes, and during the runs the load came from the measurement's own threads.
\item
\textbf{Comparisons within a session.} Dense step times measured in different sessions differed by up to 30\% because of the load on the shared machines and of thread placement. We therefore compare methods only within a session. In Table D.11b, the methods are compared within minutes of each other.
\end{itemize}

\subsection{End-to-end evaluation}\label{app:C.6}

\begin{itemize}
\item
\textbf{Tasks.} LongBench: narrativeqa, hotpotqa, gov\_report, triviaqa, passage\_retrieval\_en and lcc, with the official prompts, maximum generation lengths, chat-template rules and metrics. We use the first $k$ records of each task, truncated in the middle to 32K tokens. RULER: niah\_multikey\_2, variable tracking (vt), common words extraction (cwe) and qa\_1, with the official chat template and answer prefix and at most 128, 30, 120 and 32 new tokens.
\item
\textbf{Decoding.} fp32 greedy decoding. All methods of a run share one dense prefill and then decode as one batch, so that they see the same prompt cache. Our method computes exact attention at every fourth step and records whether the certificate held.
\item
\textbf{Additional RULER samples.} For Qwen3 and Qwen2.5 at 32K (25 per task) and Llama at 128K (12 per task), samples beyond the first 8 per task were generated with the same official scripts and seed; the first 8 match the earlier files exactly (checksums in Appendix E.2).
\item
\textbf{Aggregation task (Table D.9c).} RULER cwe at 32K (Llama 50 samples, Qwen3 and Qwen2.5 25 each, the same files as above). Dense attention, ours at $\varepsilon=0.1$, and oracle top-$k$ reading at every (layer, step, query head) as many keys as ours, with and without dense layers 0--1, decode in one batch after a shared dense prefill.
\item
\textbf{Baseline comparison (Table D.9b).} Short-output subsets: five LongBench tasks (gov\_report excluded) and three RULER tasks (cwe excluded), 20 samples each. The configurations, all with 128 sink and 128 local tokens:
\begin{itemize}
\item
oracle top-$k$ with a 2\% budget;
\item
Quest with a 2\% budget, pages of 16 tokens and 16-bit labels;
\item
HashAttention with a 2\% budget;
\item
vAttention (HashAttention) with 1\% top-$k$, 1\% base sampling, $\varepsilon=0.1$ and $\delta=0.1$.
\end{itemize}
\item
\textbf{Perturbation seeds.} Our default seeds the perturbations of a call with crc32 of ``layer/KV head/step'', and the fixed variant with crc32 of ``layer/KV head''. Tables 2 and D.9a take the per-step runs of Llama at $\varepsilon=0.1$ from the same batches as the fixed variant, and those of Qwen3 and Qwen2.5, and of Llama at $\varepsilon=0.2$, from separate runs on the same samples; Table D.9b takes its per-step runs from separate runs on the same subsets. A rerun of the per-step variant at $\varepsilon=0.1$ on the 200 Llama RULER 32K samples reproduced every output. Two further seed families (a tag appended to the seed string) rerun both variants on Llama LongBench (Table D.9e). Every run recomputes dense attention, and before joining samples across runs we check that its outputs are identical.
\item
\textbf{Matched-bytes baselines (Tables D.9d and D.12b).} Our method reads more bytes end to end than at the operator level, 25.2\% (Llama), 30.0\% (Qwen3) and 40.4\% (Qwen2.5) of dense attention on PG-19 at 32K, against 21.8--32.4\% on RULER 32K (bytes read; the per-step write adds 0.78 points). Each baseline therefore runs at its configuration with the lowest operator-level 90th-percentile error on RULER 32K among those whose bytes are at most 1.1 times our end-to-end bytes:
\begin{itemize}
\item
Llama: the MagicPIG reference implementation with $K=8$, $L=200$ (23.5\% of the bytes at the operator level), the RetroInfer CPU port with a retrieval budget of 20\% and an estimation budget of 40\% (25.8\%), and vAttention (HashAttention) with 20\% top-$k$, 5\% base sampling, $\varepsilon=0.1$ and $\delta=0.1$ (26.5\%);
\item
Qwen3: MagicPIG with $K=9$, $L=450$ (29.4\%) and RetroInfer with a retrieval budget of 25\% (31.9\%);
\item
Qwen2.5: MagicPIG with $K=9$, $L=600$ (37.7\%) and RetroInfer with a retrieval budget of 35\% (40.9\%).

HashAttention has released weights only for Llama. We also ran sparser configurations that read 0.4--0.95 times our end-to-end bytes (MagicPIG $K=9$, $L=300$ on the Qwen models, RetroInfer with a 14.4\% retrieval budget, and vAttention with 5\% top-$k$ and 2.5\% base sampling); Tables D.9d and D.12b also report them. The configurations matched to our end-to-end bytes ran on the Qwen cwe samples directly and, on PG-19 and Llama cwe, in a second run of dense attention and these baselines on the same books and samples (with our per-step method on cwe). Before joining by book or sample, we compare dense attention across the two runs. On the PG-19 books, 7--18\% of its per-token negative log-likelihoods differ, by at most 0.003, which moves no book's mean by more than $1.1\cdot10^{-5}$ and leaves perplexity unchanged at the reported precision. Each baseline's KL divergence is taken from dense attention in its own run. On Llama cwe, dense attention and our per-step method give identical outputs in both runs. MagicPIG and RetroInfer build their indexes from the prompt at the first decoding step, and generated tokens join their exact zones, as at the operator level.
\end{itemize}
\item
\textbf{Temperature sampling (Table D.9f).} RULER cwe at 32K on Llama and Qwen2.5, 11 of the samples of Table D.9c each, with dense attention, our per-step method and the matched configurations above. Every method decodes four sequences per sample in one batch after a shared dense prefill, by Gumbel-max sampling at temperature 0.7 without truncation. Draw $j$ of every method adds the same Gumbel noise at every step (in float64, seeded with crc32 of ``task/sample/draw/step''), so two methods of a draw emit the same token until their next-token distributions differ enough to change the perturbed argmax; our perturbation seeds get a tag for each draw after the first. Our method audits its certificates at every fourth step, as in greedy decoding. Rows whose sequence has ended skip attention, which leaves the other rows unchanged (checked against a run without skipping on four LongBench samples, all 48 rows identical); vAttention samples from PyTorch's global generator, so its draws also depend on the samples decoded before in the same process. In these runs, the MagicPIG port stores its hash codes in 8 or 16 bits instead of 32, with bit-identical codes and outputs on the shapes of all three models, and PyTorch tensors are not allocated on transparent huge pages, which leaves outputs unchanged (Appendix C.5).
\item
\textbf{Statistics.} The score change is the mean over tasks of the per-task mean differences from dense attention. Its 95\% confidence interval comes from a paired bootstrap stratified by task (10,000 resamples). We also report paired sign and Wilcoxon tests and the fraction of samples whose output is identical to dense attention's.
\end{itemize}

\subsection{Perplexity}\label{app:C.7}

\begin{itemize}
\item
\textbf{Data.} We use the PG-19 test set (100 books; mirror of emozilla/pg19-test). For each context length $n$ we take, in dataset order, the first books with at least $n$ tokens and keep their first $n$ tokens: 16 books at 32K, 12 at 64K and 8 at 128K.
\item
\textbf{Scoring.} We prefill the first $n-513$ tokens densely and teacher-force the last 512 tokens one decoding step at a time, with each method attending through its own path in one batch. We record the negative log-likelihood of each true token, the KL divergence from dense attention's next-token distribution, and top-1 agreement with dense attention.
\item
\textbf{Top-$k$ control.} At each layer, step and query head, oracle top-$k$ reads 128 sink tokens, 128 local tokens and the keys with the highest true scores, as many in total as our method read there (at least 256). A second variant keeps layers 0 and 1 dense.
\item
\textbf{Runs.} Our per-step method at $\varepsilon=0.1$ and $0.2$ and both top-$k$ controls, matched to the keys of the per-step method, run in one batch per model and context length. At 32K, the fixed variant runs in the same batch. At 64K and 128K, it comes from an earlier run on the same books, and its perplexity change and KL divergence are taken against dense attention in that run (Appendix D.12).
\item
\textbf{Matched-bytes baselines.} At 32K, the sparser configurations of MagicPIG, RetroInfer and vAttention (Llama only) run in the same batch as our method, and those matched to our end-to-end bytes in a second run with dense attention on the same books (Appendix C.6). Perplexity changes are taken against dense attention in the first run, and KL divergences against dense attention in the method's own run.
\item
\textbf{Free generation (Table D.12c).} On Llama at 32K, the prompt of a book is its first 31,744 tokens, prefilled densely, and every method generates 1,024 tokens by Gumbel-max sampling at temperature 0.7 without truncation, with the same noise for every method at every step (seeded with crc32 of ``book/step''). Every method other than dense attention has a shadow row in the batch, dense attention fed the method's own tokens, which gives dense attention's next-token distribution on the method's prefix at every step. We record the KL divergence from dense attention's next-token distribution on that prefix, the negative log-likelihood under dense attention of each sampled token, and the share of the generated 4-grams that occurred earlier in the same continuation. The baselines run at the configurations matched to our end-to-end bytes (Appendix C.6). Seven books of the 32K set ran (dataset indices 0--3, 8, 10 and 12).
\item
\textbf{Statistics.} Perplexity is $\exp$ of the mean negative log-likelihood over all scored tokens. Confidence intervals for the relative change come from a paired bootstrap over books (2,000 resamples), seeded with crc32 of the model and context length. The same resampled books give an interval for the ratio of a method's mean KL divergence to ours.
\end{itemize}

\section{Additional Results}\label{app:D}

Unless stated otherwise, our method uses the value-scale measure with $\varepsilon=0.1$, $\delta=0.05$, $r=4$ and a window of 64 tokens. Errors are $\|\hat o-o\|/M$ over all (layer, query head, step) of a setting.

\begin{table}[tbp]
\centering\small
\caption*{\textbf{Table D.0.} Summary of the ablations at $\varepsilon=0.1$ (ranges over the ten settings unless noted).}
\begin{tabularx}{\linewidth}{>{\raggedright\arraybackslash}X>{\raggedright\arraybackslash}X>{\raggedright\arraybackslash}X}
\toprule
Component & Variant & Effect \\
\midrule
Certified adaptive stopping & Fixed threshold without certificate, closest number of keys read & p90 error 2.4--9.6$\times$ higher, max error 3.8--16$\times$ higher; 14\% of errors exceed 0.1 at 128K \\
GQA-shared scan & Each query head scans separately & Bytes read 2.7--4.0$\times$ higher (3.9--4.0$\times$ for $G=7$) \\
Certificate rule & Uncertified CLT rule with $\ge$ 30 tail samples (CLT30) & Ours reads 1.8--3.5$\times$ more keys (Llama 16K / 32K / 128K: 2.9 / 2.9 / 3.5$\times$; Qwen2.5: 1.8$\times$); CLT30's worst layer reaches a 4.9\% violation rate (denominator) \\
Certificate rule & Oracle stopping level (IDEAL, unrealizable) & Ours reads 4.1--4.9$\times$ more keys (2.7--3.0$\times$ at $\varepsilon=0.2$) \\
8-bit refinement & Removed & Bytes $-$1\% to +4\%; speedup $-$3\% to +6\% \\
\bottomrule
\end{tabularx}
\end{table}

\subsection{Error versus keys read on all settings}\label{app:D.1}

\begin{figure}[tbp]
\centering
\includegraphics[width=\linewidth]{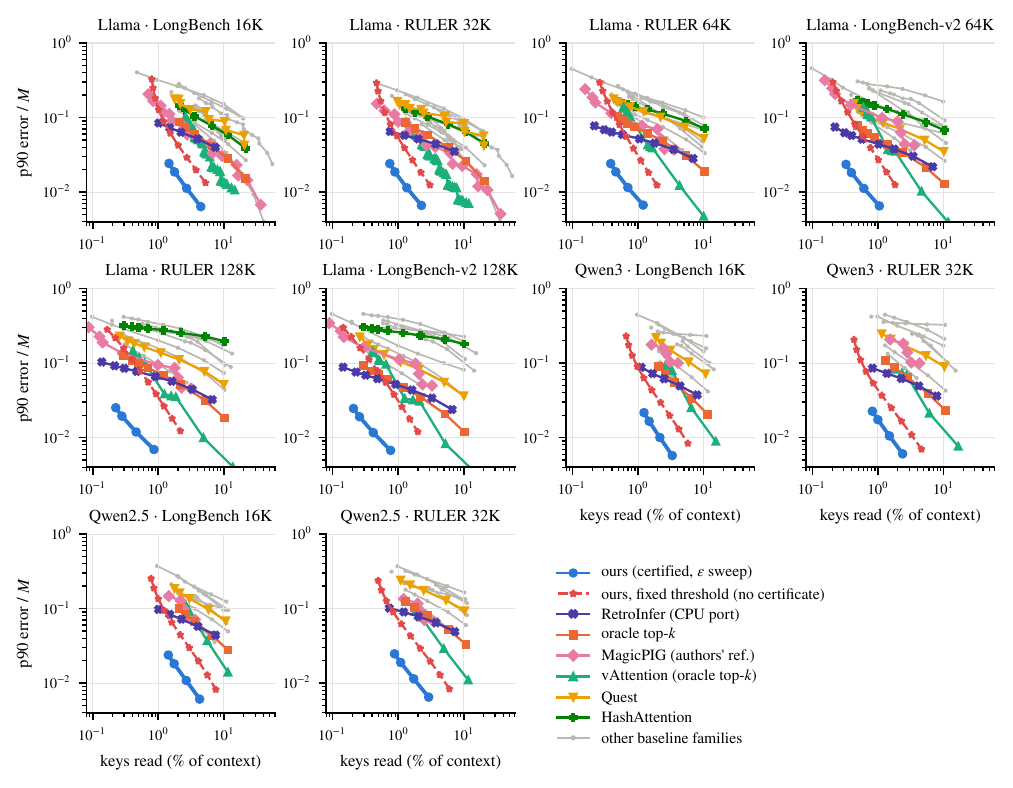}
\caption*{\textbf{Figure D.1.} 90th-percentile error/$M$ versus keys read on all ten settings. Each baseline family is drawn as the Pareto front of its tuning grid. Our method is shown for $\varepsilon\in\{0.05,0.1,0.2,0.3\}$ (solid), and the same estimator at fixed thresholds without certificates is shown dashed.}
\end{figure}

\subsection{Matched-budget comparison in detail}\label{app:D.2}

\begin{table}[tbp]
\centering\small
\caption*{\textbf{Table D.2a.} The three best baseline families at matched keys read (at most 1.1 times ours), with keys read, 90th-percentile error and maximum error.}
\begin{tabularx}{\linewidth}{>{\raggedright\arraybackslash}X>{\raggedright\arraybackslash}X>{\raggedright\arraybackslash}X}
\toprule
Model $\cdot$ data & Ours: keys / p90 / max & Best three baselines \\
\midrule
Llama $\cdot$ LongBench 16K & 2.70\% / 0.011 / 0.021 & MagicPIG (ref.) 2.75\%, 0.059 / 0.83; RetroInfer 2.22\%, 0.063 / 0.72; oracle top-$k$ 2.58\%, 0.074 / 0.43 \\
Llama $\cdot$ RULER 32K & 1.35\% / 0.011 / 0.025 & RetroInfer 1.24\%, 0.058 / 0.49; MagicPIG (ref.) 1.48\%, 0.066 / 1.21; oracle top-$k$ 1.30\%, 0.085 / 0.77 \\
Llama $\cdot$ RULER 64K & 0.70\% / 0.012 / 0.028 & RetroInfer 0.58\%, 0.059 / 0.68; oracle top-$k$ 0.69\%, 0.084 / 0.89; vAttention (oracle) 0.76\%, 0.101 / 0.60 \\
Llama $\cdot$ LongBench-v2 64K & 0.61\% / 0.011 / 0.021 & RetroInfer 0.57\%, 0.051 / 0.39; oracle top-$k$ 0.59\%, 0.068 / 0.43; oracle top-$p$ 0.61\%, 0.101 / 0.56 \\
Llama $\cdot$ RULER 128K & 0.46\% / 0.012 / 0.029 & RetroInfer 0.47\%, 0.077 / 0.65; oracle top-$k$ 0.50\%, 0.099 / 0.88; vAttention (oracle) 0.50\%, 0.117 / 0.58 \\
Llama $\cdot$ LongBench-v2 128K & 0.42\% / 0.012 / 0.022 & RetroInfer 0.32\%, 0.069 / 0.65; oracle top-$k$ 0.40\%, 0.078 / 0.44; vAttention (oracle) 0.43\%, 0.137 / 0.85 \\
Qwen3 $\cdot$ LongBench 16K & 2.15\% / 0.010 / 0.021 & RetroInfer 1.67\%, 0.072 / 1.46; oracle top-$k$ 2.07\%, 0.090 / 0.61; oracle top-$p$ 2.17\%, 0.102 / 0.58 \\
Qwen3 $\cdot$ RULER 32K & 1.50\% / 0.011 / 0.025 & RetroInfer 1.37\%, 0.073 / 1.62; oracle top-$k$ 1.29\%, 0.109 / 0.63; oracle top-$p$ 1.41\%, 0.132 / 0.61 \\
Qwen2.5 $\cdot$ LongBench 16K & 2.67\% / 0.011 / 0.022 & RetroInfer 2.26\%, 0.072 / 0.95; MagicPIG (ref.) 2.68\%, 0.077 / 1.07; oracle top-$k$ 2.57\%, 0.084 / 0.59 \\
Qwen2.5 $\cdot$ RULER 32K & 1.73\% / 0.011 / 0.027 & RetroInfer 1.22\%, 0.090 / 0.98; vAttention (oracle) 1.81\%, 0.091 / 0.47; oracle top-$k$ 1.79\%, 0.103 / 0.67 \\
\bottomrule
\end{tabularx}
\end{table}

\begin{table}[tbp]
\centering\small
\caption*{\textbf{Table D.2b.} A relaxed budget at 64K--128K (Llama): baselines may read up to 4 times our keys. All 15 families qualify in every setting.}
\begin{adjustbox}{max width=\linewidth}\begin{tabular}{lll}
\toprule
Data & Ours: keys / p90 & Best three baselines (keys, p90) \\
\midrule
RULER 64K & 0.70\% / 0.012 & MagicPIG (ref.) 2.53\%, 0.038; vAttention (oracle) 1.65\%, 0.039; RetroInfer 1.79\%, 0.045 \\
LongBench-v2 64K & 0.61\% / 0.011 & oracle top-$k$ 2.39\%, 0.034; vAttention (oracle) 1.83\%, 0.035; RetroInfer 1.77\%, 0.038 \\
RULER 128K & 0.46\% / 0.012 & vAttention (oracle) 1.57\%, 0.036; RetroInfer 1.61\%, 0.057; oracle top-$k$ 1.20\%, 0.069 \\
LongBench-v2 128K & 0.42\% / 0.012 & vAttention (oracle) 1.25\%, 0.033; RetroInfer 1.62\%, 0.043; oracle top-$k$ 1.19\%, 0.047 \\
\bottomrule
\end{tabular}\end{adjustbox}
\end{table}

With the sparse extension of the grids (Appendix C.2), the number of families with a configuration within 1.1 times our keys at 64K--128K rose from 1--3 to 12--14, and RetroInfer remained the strongest. Baseline maxima of 0.4--1.6 show that individual heads can deviate strongly, whereas ours stay below 0.03.

\begin{table}[tbp]
\centering\small
\caption*{\textbf{Table D.2c.} Matched bytes. Our bytes are measured, plus the 4 bytes per token written by the per-step perturbations; each baseline is charged a lower bound on its bytes (Appendix C.2). Best realizable baseline within 1.1 times our bytes; oracle methods and SANTA, which score every key exactly, read at least 50\% and are excluded.}
\begin{adjustbox}{max width=\linewidth}\begin{tabular}{llll}
\toprule
Model $\cdot$ data & Ours: bytes / keys / p90 / max & Best baseline: bytes, keys, p90 / max & Ratio p90 / max \\
\midrule
Llama $\cdot$ LongBench 16K & 28.6\% / 2.70\% / 0.011 / 0.021 & MagicPIG (ref.) 23.4\%, 22.8\%, 0.014 / 0.48 & 1.3$\times$ / 22$\times$ \\
Llama $\cdot$ RULER 32K & 22.6\% / 1.35\% / 0.011 / 0.025 & MagicPIG (ref.) 23.5\%, 22.9\%, 0.011 / 0.75 & 0.9$\times$ / 30$\times$ \\
Llama $\cdot$ RULER 64K & 20.1\% / 0.70\% / 0.012 / 0.028 & MagicPIG (ref.) 16.9\%, 16.4\%, 0.011 / 0.66 & 0.9$\times$ / 23$\times$ \\
Llama $\cdot$ LongBench-v2 64K & 19.9\% / 0.61\% / 0.011 / 0.021 & MagicPIG (ref.) 17.1\%, 16.7\%, 0.010 / 0.48 & 0.9$\times$ / 23$\times$ \\
Llama $\cdot$ RULER 128K & 19.5\% / 0.46\% / 0.012 / 0.029 & MagicPIG (ref.) 16.3\%, 15.9\%, 0.015 / 0.58 & 1.2$\times$ / 20$\times$ \\
Llama $\cdot$ LongBench-v2 128K & 19.3\% / 0.42\% / 0.012 / 0.022 & MagicPIG (ref.) 16.7\%, 16.2\%, 0.013 / 0.49 & 1.1$\times$ / 22$\times$ \\
Qwen3 $\cdot$ LongBench 16K & 29.3\% / 2.15\% / 0.010 / 0.021 & MagicPIG (ref.) 28.4\%, 27.8\%, 0.012 / 1.76 & 1.2$\times$ / 83$\times$ \\
Qwen3 $\cdot$ RULER 32K & 26.3\% / 1.50\% / 0.011 / 0.025 & RetroInfer 27.0\%, 20.7\%, 0.017 / 0.72 & 1.6$\times$ / 29$\times$ \\
Qwen2.5 $\cdot$ LongBench 16K & 36.9\% / 2.67\% / 0.011 / 0.022 & MagicPIG (ref.) 37.8\%, 36.9\%, 0.006 / 0.30 & 0.6$\times$ / 13$\times$ \\
Qwen2.5 $\cdot$ RULER 32K & 33.2\% / 1.73\% / 0.011 / 0.027 & MagicPIG (ref.) 27.6\%, 27.0\%, 0.011 / 0.94 & 0.9$\times$ / 34$\times$ \\
\bottomrule
\end{tabular}\end{adjustbox}
\end{table}

On Qwen2.5 LongBench 16K, the MagicPIG port with $K=9$ and $L=600$ reads 37\% of the keys and reaches 0.6 times our 90th-percentile error, with 13 times our maximum. The maximum ratios of 13--83 compare with the best baseline by 90th percentile. Over all realizable baseline configurations within 1.1 times our bytes, the smallest maximum error is 4.9--29 times ours. It is closest to ours for the denser RetroInfer configurations: 4.9--9.2 times on Llama at 64K--128K, where they read 13.6--13.8\% of the keys, and 7.7 times on Qwen2.5 RULER 32K, where one reads 29.7\%.

\begin{figure}[tbp]
\centering
\includegraphics[width=\linewidth]{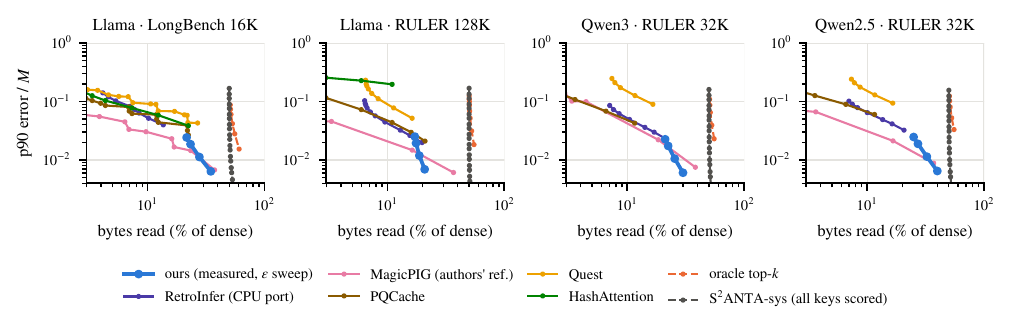}
\caption*{\textbf{Figure D.2.} 90th-percentile error/$M$ against bytes read, relative to dense attention: ours as measured ($\varepsilon\in\{0.05,0.1,0.2,0.3\}$) and, for each baseline family, the Pareto front of its configurations under the lower bounds of Appendix C.2, on the settings of Figure 2.}
\end{figure}

\begin{table}[tbp]
\centering\small
\caption*{\textbf{Table D.2d.} SANTA (Lee et al., 2026a) at matched values read (at most 1.1 times our keys read; best sample size per variant). SANTA also scores every key exactly.}
\begin{adjustbox}{max width=\linewidth}\begin{tabular}{lllll}
\toprule
Model $\cdot$ data & Ours: keys / p90 / max & SANTA & S$^2$ANTA-sys & S$^2$ANTA-strat \\
\midrule
Llama $\cdot$ LongBench 16K & 2.70\% / 0.011 / 0.021 & 2.67\% / 0.018 / 0.058 & 2.48\% / 0.011 / 0.025 & 2.41\% / 0.012 / 0.024 \\
Llama $\cdot$ RULER 32K & 1.35\% / 0.011 / 0.025 & 1.29\% / 0.019 / 0.063 & 1.45\% / 0.009 / 0.026 & 1.42\% / 0.010 / 0.026 \\
Llama $\cdot$ RULER 64K & 0.70\% / 0.012 / 0.028 & 0.65\% / 0.019 / 0.067 & 0.72\% / 0.010 / 0.031 & 0.71\% / 0.010 / 0.024 \\
Llama $\cdot$ LongBench-v2 64K & 0.61\% / 0.011 / 0.021 & 0.62\% / 0.018 / 0.045 & 0.56\% / 0.012 / 0.027 & 0.55\% / 0.012 / 0.037 \\
Llama $\cdot$ RULER 128K & 0.46\% / 0.012 / 0.029 & 0.45\% / 0.018 / 0.065 & 0.50\% / 0.010 / 0.024 & 0.49\% / 0.011 / 0.037 \\
Llama $\cdot$ LongBench-v2 128K & 0.42\% / 0.012 / 0.022 & 0.42\% / 0.018 / 0.045 & 0.38\% / 0.012 / 0.029 & 0.37\% / 0.012 / 0.027 \\
Qwen3 $\cdot$ LongBench 16K & 2.15\% / 0.010 / 0.021 & 2.18\% / 0.018 / 0.052 & 2.06\% / 0.010 / 0.027 & 2.01\% / 0.010 / 0.029 \\
Qwen3 $\cdot$ RULER 32K & 1.50\% / 0.011 / 0.025 & 1.48\% / 0.018 / 0.052 & 1.38\% / 0.010 / 0.032 & 1.35\% / 0.011 / 0.030 \\
Qwen2.5 $\cdot$ LongBench 16K & 2.67\% / 0.011 / 0.022 & 2.86\% / 0.018 / 0.043 & 2.71\% / 0.009 / 0.022 & 2.64\% / 0.010 / 0.027 \\
Qwen2.5 $\cdot$ RULER 32K & 1.73\% / 0.011 / 0.027 & 1.85\% / 0.018 / 0.042 & 1.71\% / 0.010 / 0.026 & 1.66\% / 0.011 / 0.036 \\
\bottomrule
\end{tabular}\end{adjustbox}
\end{table}

At equal values read, i.i.d. SANTA has 1.52--1.81 times our 90th-percentile error and 1.56--2.73 times our maximum. Its systematic and stratified variants reach 0.82--1.07 times our 90th percentile and 0.85--1.75 times our maximum. All three compute every score exactly, which reads all keys (at least 50\% of the bytes of dense attention, against our 19.3--36.9\%), and they have no stopping rule that adapts the sample to each head.

\textbf{Simpler alternatives.} Two simpler alternatives either read every key in low precision or sample from the 4-bit scores alone. Dense attention over a KIVI-style quantized cache \citep{liu2024kivi} at 4 bits reads 29--32\% of the bytes of dense attention, about as many as our method at 16K--32K (23--37\%) and more at 64K--128K (19--20\%). Its 90th-percentile error is 2.6--5.3 times ours, and its maximum error 7--28 times (Table D.2e). At 8 bits, its 90th-percentile error is 3--5 times lower than ours and its maximum error similar, but it reads 1.5--2.8 times our bytes. Quantization also shrinks the cache, which our index enlarges, and our estimator could run over a quantized cache, with certificates relative to that cache. The second alternative draws a Poisson PPS sample from the softmax of the 4-bit scores and weights the exact scores of the sampled keys by their inclusion probabilities. When it reads exactly as many keys as our method does for the same query, it matches our 90th-percentile error, but its maximum error is up to 5.9 times ours. Sampling with replacement from the same proposal is 4.5--6.3 times worse at the 90th percentile (Table D.2f). Neither alternative has a certificate, and without one the budget of each query must be fixed in advance. With the same mean keys read but one budget for all queries of a setting, Poisson PPS has 1.5--2.1 times our 90th-percentile error and 3.2--6.2 times our maximum error. The certificate thus matters for the adaptive budget per query, which the estimator alone does not provide.

\begin{table}[tbp]
\centering\small
\caption*{\textbf{Table D.2e.} Dense attention over a KIVI-style quantized KV cache (keys per channel, values per token in groups of 32 channels, fp16 scales, the last 128 tokens in 16 bits): bytes read / 90th-percentile / maximum error/$M$, against our method at $\varepsilon=0.1$ (with the 4 bytes per token written by the per-step perturbations). The key groups span 128 tokens, or 32 as in KIVI.}
\begin{tabularx}{\linewidth}{>{\raggedright\arraybackslash}X>{\raggedright\arraybackslash}X>{\raggedright\arraybackslash}X>{\raggedright\arraybackslash}X>{\raggedright\arraybackslash}X>{\raggedright\arraybackslash}X>{\raggedright\arraybackslash}X}
\toprule
Model $\cdot$ data & Ours & 2-bit, 128-token key groups & 4-bit, 128-token key groups & 8-bit, 128-token key groups & 2-bit, 32-token key groups & 4-bit, 32-token key groups \\
\midrule
Llama $\cdot$ LongBench 16K & 28.6\% / 0.011 / 0.021 & 17.5\% / 0.2394 / 1.855 & 29.8\% / 0.0458 / 0.284 & 54.5\% / 0.0026 / 0.017 & 19.8\% / 0.1732 / 1.283 & 32.1\% / 0.0371 / 0.268 \\
Llama $\cdot$ RULER 32K & 22.6\% / 0.011 / 0.025 & 16.9\% / 0.2363 / 2.047 & 29.3\% / 0.0464 / 0.236 & 54.2\% / 0.0026 / 0.013 & 19.2\% / 0.1749 / 1.925 & 31.6\% / 0.0381 / 0.251 \\
Llama $\cdot$ RULER 64K & 20.1\% / 0.012 / 0.028 & 16.6\% / 0.2370 / 2.661 & 29.1\% / 0.0472 / 0.325 & 54.0\% / 0.0027 / 0.018 & 19.0\% / 0.1764 / 1.448 & 31.4\% / 0.0382 / 0.201 \\
Llama $\cdot$ LongBench-v2 64K & 19.9\% / 0.011 / 0.021 & 16.7\% / 0.2133 / 1.269 & 29.1\% / 0.0427 / 0.254 & 54.1\% / 0.0024 / 0.013 & 19.0\% / 0.1592 / 1.410 & 31.5\% / 0.0343 / 0.170 \\
Llama $\cdot$ RULER 128K & 19.5\% / 0.012 / 0.029 & 16.5\% / 0.2160 / 3.702 & 29.0\% / 0.0406 / 0.262 & 54.0\% / 0.0024 / 0.017 & 18.9\% / 0.1604 / 1.715 & 31.4\% / 0.0337 / 0.232 \\
Llama $\cdot$ LongBench-v2 128K & 19.3\% / 0.012 / 0.022 & 16.5\% / 0.2069 / 1.001 & 29.0\% / 0.0377 / 0.276 & 54.0\% / 0.0022 / 0.019 & 18.9\% / 0.1397 / 1.291 & 31.4\% / 0.0299 / 0.162 \\
Qwen3 $\cdot$ LongBench 16K & 29.3\% / 0.010 / 0.021 & 17.5\% / 0.2935 / 6.964 & 29.8\% / 0.0476 / 0.399 & 54.5\% / 0.0028 / 0.020 & 19.8\% / 0.2246 / 2.424 & 32.2\% / 0.0392 / 0.342 \\
Qwen3 $\cdot$ RULER 32K & 26.3\% / 0.011 / 0.025 & 16.9\% / 0.2716 / 4.516 & 29.3\% / 0.0430 / 0.362 & 54.2\% / 0.0025 / 0.015 & 19.2\% / 0.2168 / 4.434 & 31.7\% / 0.0370 / 0.414 \\
Qwen2.5 $\cdot$ LongBench 16K & 36.9\% / 0.011 / 0.022 & 17.6\% / 0.3411 / 4.972 & 29.9\% / 0.0574 / 0.629 & 54.6\% / 0.0033 / 0.024 & 19.9\% / 0.2414 / 3.472 & 32.2\% / 0.0465 / 0.382 \\
Qwen2.5 $\cdot$ RULER 32K & 33.2\% / 0.011 / 0.027 & 16.9\% / 0.2775 / 3.934 & 29.3\% / 0.0487 / 0.392 & 54.2\% / 0.0028 / 0.026 & 19.2\% / 0.2131 / 2.903 & 31.6\% / 0.0408 / 0.255 \\
\bottomrule
\end{tabularx}
\end{table}

Over both group sizes, the 2-bit cache has 12--31 times our 90th-percentile error and 45--330 times our maximum at 0.48--0.98 times our bytes. The 4-bit cache has 2.6--5.3 and 7--28 times at 0.81--1.6 times our bytes, and the 8-bit cache 0.18--0.31 and 0.5--1.1 times at 1.5--2.8 times our bytes.

\begin{table}[tbp]
\centering\small
\caption*{\textbf{Table D.2f.} Sampling with fresh randomness from the 4-bit scores (no certificate): keys read (distinct keys for sampling with replacement) / 90th-percentile / maximum error/$M$. The first two read the number of keys our method read for the same query ($\varepsilon=0.1$, value scale), which a method without a certificate does not know; the last reads the same number of keys for every query, our mean over the setting.}
\begin{tabularx}{\linewidth}{>{\raggedright\arraybackslash}X>{\raggedright\arraybackslash}X>{\raggedright\arraybackslash}X>{\raggedright\arraybackslash}X>{\raggedright\arraybackslash}X}
\toprule
Model $\cdot$ data & Ours, $\varepsilon=0.1$ & With replacement from the 4-bit proposal & Poisson PPS on 4-bit probabilities & Poisson PPS, one budget per setting \\
\midrule
Llama $\cdot$ LongBench 16K & 2.70\% / 0.0112 / 0.021 & 1.47\% / 0.0616 / 0.645 & 2.70\% / 0.0115 / 0.108 & 2.70\% / 0.0192 / 0.093 \\
Llama $\cdot$ RULER 32K & 1.35\% / 0.0114 / 0.025 & 0.82\% / 0.0630 / 0.612 & 1.35\% / 0.0117 / 0.080 & 1.35\% / 0.0187 / 0.100 \\
Llama $\cdot$ RULER 64K & 0.70\% / 0.0115 / 0.028 & 0.43\% / 0.0624 / 0.689 & 0.70\% / 0.0118 / 0.033 & 0.70\% / 0.0188 / 0.090 \\
Llama $\cdot$ LongBench-v2 64K & 0.61\% / 0.0112 / 0.021 & 0.35\% / 0.0626 / 0.492 & 0.61\% / 0.0116 / 0.022 & 0.61\% / 0.0171 / 0.094 \\
Llama $\cdot$ RULER 128K & 0.46\% / 0.0119 / 0.029 & 0.28\% / 0.0574 / 0.706 & 0.46\% / 0.0122 / 0.031 & 0.46\% / 0.0194 / 0.094 \\
Llama $\cdot$ LongBench-v2 128K & 0.42\% / 0.0117 / 0.022 & 0.24\% / 0.0553 / 0.314 & 0.42\% / 0.0120 / 0.024 & 0.42\% / 0.0177 / 0.071 \\
Qwen3 $\cdot$ LongBench 16K & 2.15\% / 0.0100 / 0.021 & 1.23\% / 0.0632 / 2.524 & 2.15\% / 0.0104 / 0.126 & 2.15\% / 0.0204 / 0.131 \\
Qwen3 $\cdot$ RULER 32K & 1.50\% / 0.0106 / 0.025 & 0.93\% / 0.0552 / 2.203 & 1.50\% / 0.0110 / 0.024 & 1.50\% / 0.0198 / 0.134 \\
Qwen2.5 $\cdot$ LongBench 16K & 2.67\% / 0.0108 / 0.022 & 1.69\% / 0.0617 / 1.217 & 2.67\% / 0.0111 / 0.057 & 2.67\% / 0.0223 / 0.088 \\
Qwen2.5 $\cdot$ RULER 32K & 1.73\% / 0.0114 / 0.027 & 1.14\% / 0.0513 / 1.980 & 1.73\% / 0.0117 / 0.035 & 1.73\% / 0.0217 / 0.120 \\
\bottomrule
\end{tabularx}
\end{table}

A PPS sample from the 4-bit scores comes close to our method only with our per-query budget, at 1.02--1.04 times our 90th-percentile error and 0.97--5.9 times our maximum. With one budget per setting, the ratios grow to 1.5--2.1 and 3.2--6.2, and sampling with replacement has 4.5--6.3 and 14--120 times.

\subsection{Three error measures}\label{app:D.3}

\begin{table}[tbp]
\centering\small
\caption*{\textbf{Table D.3.} Keys read / bytes read / 90th-percentile error/$M$ / 90th-percentile error/$\|o\|$ for the three measures at $\varepsilon=0.1$. No configuration has a violation or a fallback.}
\begin{adjustbox}{max width=\linewidth}\begin{tabular}{llll}
\toprule
Model $\cdot$ data & Denominator & Value scale $M$ & Output norm $\|o\|$ \\
\midrule
Llama $\cdot$ LongBench 16K & 1.44\% / 20.4\% / 0.043 / 0.334 & 2.70\% / 27.8\% / 0.011 / 0.087 & 9.38\% / 50.7\% / 0.005 / 0.013 \\
Llama $\cdot$ RULER 32K & 0.75\% / 18.1\% / 0.040 / 0.278 & 1.35\% / 21.8\% / 0.011 / 0.075 & 4.70\% / 35.5\% / 0.005 / 0.014 \\
Llama $\cdot$ RULER 64K & 0.39\% / 17.2\% / 0.040 / 0.281 & 0.70\% / 19.3\% / 0.012 / 0.076 & 2.63\% / 28.2\% / 0.005 / 0.014 \\
Llama $\cdot$ LongBench-v2 64K & 0.32\% / 17.0\% / 0.040 / 0.319 & 0.61\% / 19.1\% / 0.011 / 0.085 & 2.64\% / 28.6\% / 0.005 / 0.014 \\
Llama $\cdot$ RULER 128K & 0.24\% / 17.1\% / 0.040 / 0.252 & 0.46\% / 18.8\% / 0.012 / 0.073 & 1.91\% / 25.5\% / 0.005 / 0.015 \\
Llama $\cdot$ LongBench-v2 128K & 0.21\% / 16.8\% / 0.039 / 0.263 & 0.42\% / 18.5\% / 0.012 / 0.081 & 1.93\% / 25.5\% / 0.005 / 0.014 \\
Qwen3 $\cdot$ LongBench 16K & 1.19\% / 21.6\% / 0.039 / 0.189 & 2.15\% / 28.5\% / 0.010 / 0.049 & 5.11\% / 39.8\% / 0.005 / 0.012 \\
Qwen3 $\cdot$ RULER 32K & 0.83\% / 20.1\% / 0.036 / 0.161 & 1.50\% / 25.5\% / 0.011 / 0.046 & 3.49\% / 34.4\% / 0.005 / 0.012 \\
Qwen2.5 $\cdot$ LongBench 16K & 1.57\% / 26.3\% / 0.029 / 0.152 & 2.67\% / 36.1\% / 0.011 / 0.057 & 7.26\% / 54.5\% / 0.005 / 0.012 \\
Qwen2.5 $\cdot$ RULER 32K & 1.01\% / 24.2\% / 0.029 / 0.136 & 1.73\% / 32.4\% / 0.011 / 0.054 & 4.82\% / 48.4\% / 0.006 / 0.013 \\
\bottomrule
\end{tabular}\end{adjustbox}
\end{table}

The output-norm measure reads 2.3--4.6 times the keys of the value scale. The ratio is largest on Llama at long contexts, where cancellation is strongest. The denominator measure does not control the output: its 90th-percentile error relative to $\|o\|$ is 0.14--0.33.

\begin{table}[tbp]
\centering\small
\caption*{\textbf{Table D.3b.} Per-call reseeding and a per-call failure probability for whole generations ($\varepsilon=0.1$): keys read / bytes read / 90th-percentile and maximum error, for perturbations fixed per KV head, drawn afresh for every call, and drawn afresh with $\delta=0.05/(HT)$ ($H$ query heads over all layers, $T=512$ steps). Errors are relative to $M$ in the upper table (value scale) and to $\|o\|$ in the lower one (output norm). No configuration has a violation or a fallback.}
\begin{adjustbox}{max width=\linewidth}\begin{tabular}{lllll}
\toprule
Value scale $\cdot$ model $\cdot$ data & Fixed & Reseeded & Reseeded, $\delta/(HT)$ & Keys ratio \\
\midrule
Llama $\cdot$ LongBench 16K & 2.70\% / 27.8\% / 0.0112 / 0.021 & 2.70\% / 27.8\% / 0.0112 / 0.037 & 4.18\% / 34.0\% / 0.0069 / 0.014 & 1.55$\times$ \\
Llama $\cdot$ RULER 32K & 1.35\% / 21.8\% / 0.0114 / 0.025 & 1.35\% / 21.8\% / 0.0114 / 0.023 & 2.11\% / 25.3\% / 0.0071 / 0.016 & 1.56$\times$ \\
Llama $\cdot$ RULER 64K & 0.70\% / 19.3\% / 0.0115 / 0.028 & 0.70\% / 19.3\% / 0.0115 / 0.030 & 1.11\% / 21.4\% / 0.0072 / 0.019 & 1.59$\times$ \\
Llama $\cdot$ LongBench-v2 64K & 0.61\% / 19.1\% / 0.0112 / 0.021 & 0.61\% / 19.1\% / 0.0113 / 0.025 & 0.98\% / 21.1\% / 0.0070 / 0.013 & 1.60$\times$ \\
Llama $\cdot$ RULER 128K & 0.46\% / 18.8\% / 0.0119 / 0.029 & 0.46\% / 18.8\% / 0.0120 / 0.027 & 0.77\% / 20.5\% / 0.0075 / 0.016 & 1.69$\times$ \\
Llama $\cdot$ LongBench-v2 128K & 0.42\% / 18.5\% / 0.0117 / 0.022 & 0.42\% / 18.5\% / 0.0117 / 0.025 & 0.70\% / 20.1\% / 0.0073 / 0.017 & 1.67$\times$ \\
Qwen3 $\cdot$ LongBench 16K & 2.15\% / 28.5\% / 0.0100 / 0.021 & 2.15\% / 28.6\% / 0.0100 / 0.021 & 3.22\% / 33.8\% / 0.0061 / 0.014 & 1.50$\times$ \\
Qwen3 $\cdot$ RULER 32K & 1.50\% / 25.5\% / 0.0106 / 0.025 & 1.50\% / 25.5\% / 0.0105 / 0.026 & 2.28\% / 29.8\% / 0.0064 / 0.016 & 1.53$\times$ \\
Qwen2.5 $\cdot$ LongBench 16K & 2.67\% / 36.1\% / 0.0108 / 0.022 & 2.66\% / 36.1\% / 0.0108 / 0.031 & 4.00\% / 43.2\% / 0.0067 / 0.015 & 1.50$\times$ \\
Qwen2.5 $\cdot$ RULER 32K & 1.73\% / 32.4\% / 0.0114 / 0.027 & 1.73\% / 32.4\% / 0.0114 / 0.023 & 2.68\% / 38.9\% / 0.0071 / 0.017 & 1.55$\times$ \\
\bottomrule
\end{tabular}\end{adjustbox}

\vspace{0.6em}
\begin{adjustbox}{max width=\linewidth}\begin{tabular}{lllll}
\toprule
Output norm $\cdot$ model $\cdot$ data & Fixed & Reseeded & Reseeded, $\delta/(HT)$ & Keys ratio \\
\midrule
Llama $\cdot$ LongBench 16K & 9.38\% / 50.7\% / 0.0132 / -- & 9.38\% / 50.7\% / 0.0131 / -- & 13.11\% / 60.2\% / 0.0079 / -- & 1.40$\times$ \\
Llama $\cdot$ RULER 32K & 4.70\% / 35.5\% / 0.0139 / -- & 4.70\% / 35.4\% / 0.0139 / -- & 6.90\% / 42.5\% / 0.0084 / -- & 1.47$\times$ \\
Llama $\cdot$ RULER 64K & 2.63\% / 28.2\% / 0.0143 / -- & 2.63\% / 28.2\% / 0.0143 / -- & 3.97\% / 33.1\% / 0.0087 / -- & 1.51$\times$ \\
Llama $\cdot$ LongBench-v2 64K & 2.64\% / 28.6\% / 0.0140 / -- & 2.64\% / 28.6\% / 0.0140 / -- & 3.95\% / 33.5\% / 0.0086 / -- & 1.50$\times$ \\
Llama $\cdot$ RULER 128K & 1.91\% / 25.5\% / 0.0146 / -- & 1.91\% / 25.5\% / 0.0146 / -- & 2.91\% / 29.3\% / 0.0090 / -- & 1.53$\times$ \\
Llama $\cdot$ LongBench-v2 128K & 1.93\% / 25.5\% / 0.0142 / -- & 1.93\% / 25.5\% / 0.0143 / -- & 2.89\% / 29.2\% / 0.0087 / -- & 1.50$\times$ \\
Qwen3 $\cdot$ LongBench 16K & 5.11\% / 39.8\% / 0.0118 / -- & 5.11\% / 39.8\% / 0.0118 / -- & 7.30\% / 47.2\% / 0.0070 / -- & 1.43$\times$ \\
Qwen3 $\cdot$ RULER 32K & 3.49\% / 34.4\% / 0.0125 / -- & 3.49\% / 34.4\% / 0.0124 / -- & 5.12\% / 40.7\% / 0.0074 / -- & 1.47$\times$ \\
Qwen2.5 $\cdot$ LongBench 16K & 7.26\% / 54.5\% / 0.0123 / -- & 7.25\% / 54.5\% / 0.0123 / -- & 9.77\% / 62.6\% / 0.0074 / -- & 1.35$\times$ \\
Qwen2.5 $\cdot$ RULER 32K & 4.82\% / 48.4\% / 0.0132 / -- & 4.82\% / 48.4\% / 0.0131 / -- & 6.83\% / 56.7\% / 0.0081 / -- & 1.42$\times$ \\
\bottomrule
\end{tabular}\end{adjustbox}
\end{table}

With $\delta/(HT)$, keys read grow by 1.50--1.69 times for the value scale and 1.35--1.53 times for the output norm, and bytes read by 1.6--7.1 and 3.7--9.5 points.

\subsection{Keys read versus context length}\label{app:D.4}

\begin{table}[tbp]
\centering\small
\caption*{\textbf{Table D.4.} Keys read (window plus sample) on Llama, value scale.}
\begin{adjustbox}{max width=\linewidth}\begin{tabular}{lllll}
\toprule
$\varepsilon$ & 16K & 32K & 64K (RULER / LongBench-v2) & 128K (RULER / LongBench-v2) \\
\midrule
0.05 & 728 & 742 & 784 / 688 & 1125 / 1003 \\
0.1 & 442 & 443 & 458 / 400 & 601 / 546 \\
0.2 & 290 & 292 & 300 / 258 & 365 / 337 \\
\bottomrule
\end{tabular}\end{adjustbox}
\end{table}

Most of the growth occurs between 64K and 128K. Theorem 2 bounds the sample size independently of $n$ in the worst case, in which all of the mass lies in the tail. In real heads, the window, the sinks and the heavy keys carry part of $D$ exactly, so a context whose tail holds a larger share of the mass needs more samples while staying below the bound.

Within a setting, the number of keys read varies by about two orders of magnitude across layers, heads and steps (Figure D.3). On Llama RULER 128K, the first layer reads a median of 2.9\% of the keys, about ten times the median of the other layers.

\begin{figure}[tbp]
\centering
\includegraphics[width=\linewidth]{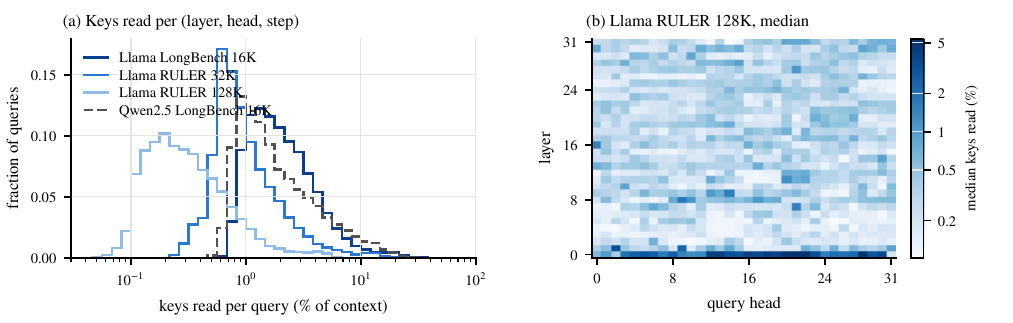}
\caption*{\textbf{Figure D.3.} Keys read per query at $\varepsilon=0.1$ (value scale). \emph{(a)} Distribution over all (layer, query head, step) of four settings. \emph{(b)} Median over samples and steps for each layer and query head of Llama RULER 128K, on a logarithmic color scale.}
\end{figure}

\begin{figure}[tbp]
\centering
\includegraphics[width=\linewidth]{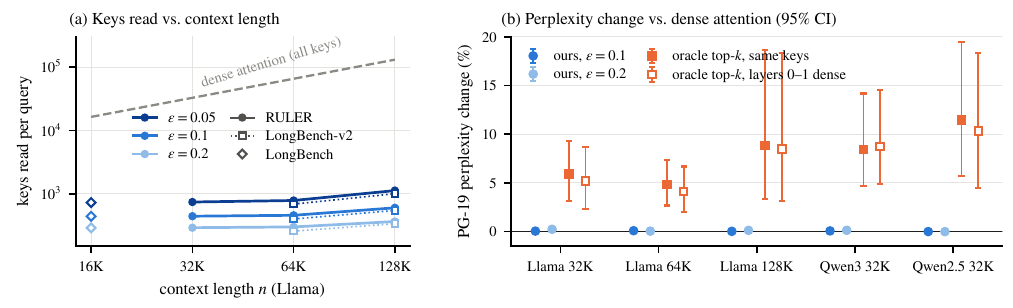}
\caption*{\textbf{Figure D.6.} \emph{(a)} Keys read per query (window plus sample) on Llama against the context length, for three values of $\varepsilon$ (Table D.4). Circles connected by lines are RULER at 32K--128K, squares LongBench-v2 at 64K and 128K, and diamonds LongBench at 16K; the dashed line is dense attention. \emph{(b)} Change in PG-19 perplexity relative to dense attention, with 95\% bootstrap intervals over books (Table D.12); our method draws fresh perturbations at every step. Oracle top-$k$ reads, at every layer, step and query head, as many keys as our method at $\varepsilon=0.1$.}
\end{figure}

\subsection{Certified stopping versus fixed thresholds}\label{app:D.5}

\begin{table}[tbp]
\centering\small
\caption*{\textbf{Table D.5.} Our certified rule against the same estimator at the fixed level $j$ whose keys read are closest (no certificate).}
\begin{adjustbox}{max width=\linewidth}\begin{tabular}{lllll}
\toprule
Model $\cdot$ data & Certified: keys / p90 / max & Fixed: level, keys / p90 / max & Fixed: errors > 0.1 & Ratio p90 / max \\
\midrule
Llama $\cdot$ LongBench 16K & 2.70\% / 0.011 / 0.021 & $j=36$: 2.72\% / 0.029 / 0.081 & 0 & 2.6$\times$ / 3.8$\times$ \\
Llama $\cdot$ RULER 32K & 1.35\% / 0.011 / 0.025 & $j=32$: 1.25\% / 0.039 / 0.128 & 0.01\% & 3.4$\times$ / 5.0$\times$ \\
Llama $\cdot$ RULER 64K & 0.70\% / 0.012 / 0.028 & $j=28$: 0.66\% / 0.056 / 0.189 & 0.30\% & 4.8$\times$ / 6.7$\times$ \\
Llama $\cdot$ LongBench-v2 64K & 0.61\% / 0.011 / 0.021 & $j=28$: 0.58\% / 0.056 / 0.178 & 0.25\% & 5.0$\times$ / 8.5$\times$ \\
Llama $\cdot$ RULER 128K & 0.46\% / 0.012 / 0.029 & $j=20$: 0.40\% / 0.112 / 0.404 & 13.9\% & 9.4$\times$ / 14$\times$ \\
Llama $\cdot$ LongBench-v2 128K & 0.42\% / 0.012 / 0.022 & $j=20$: 0.36\% / 0.112 / 0.353 & 14.3\% & 9.6$\times$ / 16$\times$ \\
Qwen3 $\cdot$ LongBench 16K & 2.15\% / 0.010 / 0.021 & $j=32$: 2.20\% / 0.030 / 0.136 & 0.01\% & 2.9$\times$ / 6.4$\times$ \\
Qwen3 $\cdot$ RULER 32K & 1.50\% / 0.011 / 0.025 & $j=32$: 1.61\% / 0.025 / 0.119 & 0.004\% & 2.4$\times$ / 4.8$\times$ \\
Qwen2.5 $\cdot$ LongBench 16K & 2.67\% / 0.011 / 0.022 & $j=32$: 2.98\% / 0.030 / 0.099 & 0 & 2.7$\times$ / 4.4$\times$ \\
Qwen2.5 $\cdot$ RULER 32K & 1.73\% / 0.011 / 0.027 & $j=28$: 1.54\% / 0.043 / 0.136 & 0.05\% & 3.8$\times$ / 5.0$\times$ \\
\bottomrule
\end{tabular}\end{adjustbox}
\end{table}

Fixed levels are spaced a quarter octave apart, so the matched level reads slightly more or fewer keys. At 128K, the next denser level ($j=24$, 0.49--0.53\% of the keys) still has a 90th-percentile error of 0.079, 6.6--6.8 times ours. Even without certificates, the PPS estimator beats most baselines at equal keys read (dashed lines in Figure D.1). Figure D.4 shows the full error distributions of the two rules on two settings.

\begin{figure}[tbp]
\centering
\includegraphics[width=\linewidth]{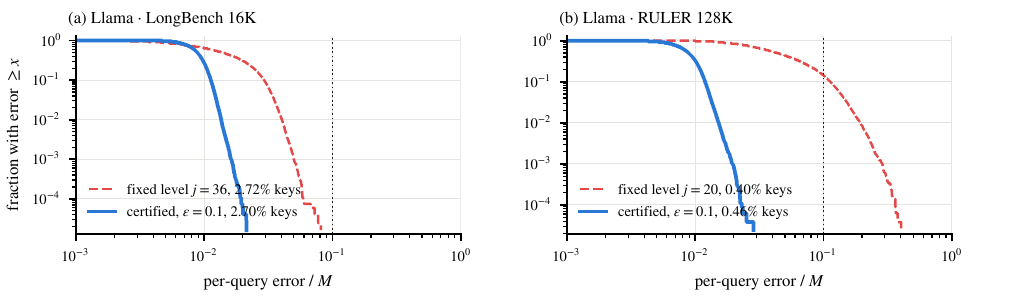}
\caption*{\textbf{Figure D.4.} Fraction of queries whose error$/M$ is at least $x$, for certified stopping at $\varepsilon=0.1$ and for the same estimator at the fixed level whose keys read are closest (Table D.5). The dotted line marks 0.1.}
\end{figure}

\subsection{GQA sharing and quantization}\label{app:D.6}

\begin{table}[tbp]
\centering\small
\caption*{\textbf{Table D.6.} Bytes read relative to dense attention when the query heads of a group share the candidate scan (default) and when each query head scans separately. Keys read are unchanged.}
\begin{adjustbox}{max width=\linewidth}\begin{tabular}{llll}
\toprule
Model $\cdot$ data & Shared & Separate & Ratio \\
\midrule
Llama $\cdot$ LongBench 16K & 27.8\% & 80.2\% & 2.9$\times$ \\
Llama $\cdot$ RULER 32K & 21.8\% & 71.0\% & 3.3$\times$ \\
Llama $\cdot$ RULER 64K & 19.3\% & 67.0\% & 3.5$\times$ \\
Llama $\cdot$ LongBench-v2 64K & 19.1\% & 66.6\% & 3.5$\times$ \\
Llama $\cdot$ RULER 128K & 18.8\% & 65.8\% & 3.5$\times$ \\
Llama $\cdot$ LongBench-v2 128K & 18.5\% & 65.6\% & 3.5$\times$ \\
Qwen3 $\cdot$ LongBench 16K & 28.5\% & 78.3\% & 2.7$\times$ \\
Qwen3 $\cdot$ RULER 32K & 25.5\% & 73.4\% & 2.9$\times$ \\
Qwen2.5 $\cdot$ LongBench 16K ($G=7$) & 36.1\% & 140.2\% & 3.9$\times$ \\
Qwen2.5 $\cdot$ RULER 32K ($G=7$) & 32.4\% & 130.0\% & 4.0$\times$ \\
\bottomrule
\end{tabular}\end{adjustbox}
\end{table}

We also test two quantization variants over the ten settings. One int8 level for the query weights instead of two changes bytes read by +3\% to +8\% (+6\% to +8\% on Qwen), and removing the 8-bit refinement changes them by $-$1\% to +4\%. Errors are unchanged in both cases. In time, removing the 8-bit level changes the speedup by $-$3\% to +3\% (Table D.11a). Since the 8-bit codes add 25 percentage points of storage, they can be dropped.

\subsection{Certificate rules}\label{app:D.7}

All rules below run on the same sample paths, with the same perturbations, and differ only in when they stop. ``Keys read'' here counts the sample only: the window and the candidate scan are excluded, so the numbers are comparable across rules but not with the other tables. The rules are:
\begin{itemize}
\item
\textbf{S4}: an uncentered certificate (a Bernstein bound for the denominator, a dyadic grid, $\delta_j$ split three ways);
\item
\textbf{KL1 / KL4}: the Poisson-KL denominator interval with the centered Pinelis bound, on a grid with $r=1$ or $r=4$ levels per octave. \textbf{KL4 is the rule in our kernel.}
\item
\textbf{KB4}: KL4 with Bousquet's inequality \citep{bousquet2002} in place of Pinelis's. Its deviation term uses an upper confidence bound on the operator norm of the value covariance, obtained from matrix Bernstein \citep{tropp2011};
\item
\textbf{CLT30}: a normal approximation used once at least 30 tail samples exist (no guarantee);
\item
\textbf{IDEAL}: the first level whose actual error is at most $\varepsilon$ (unrealizable; a lower bound).
\end{itemize}

\begin{table}[tbp]
\centering\small
\caption*{\textbf{Table D.7a.} Centering and the Chernoff interval (Llama LongBench 16K, 5 samples $\times$ 32 layers $\times$ 8 steps $\times$ 32 heads = 40,960 queries; mean keys read).}
\begin{adjustbox}{max width=\linewidth}\begin{tabular}{llllllll}
\toprule
Measure & $\varepsilon$ & S4 & KL1 & KL4 & KB4 & CLT30 & IDEAL \\
\midrule
Denominator & 0.05 & 6.57\% & 2.73\% & 2.38\% & 2.38\% & 1.18\% & 0.65\% \\
Denominator & 0.1 & 2.97\% & 1.55\% & 1.41\% & 1.41\% & 0.91\% & 0.63\% \\
Denominator & 0.2 & 1.51\% & 1.00\% & 0.95\% & 0.95\% & 0.83\% & 0.62\% \\
Output norm & 0.05 & 32.7\% & 17.0\% & 14.7\% & 12.5\% & 5.72\% & 3.21\% \\
Output norm & 0.1 & 20.7\% & 11.4\% & 9.76\% & 8.31\% & 3.22\% & 1.76\% \\
Output norm & 0.2 & 12.3\% & 7.47\% & 6.34\% & 5.46\% & 1.81\% & 1.03\% \\
Value scale & 0.05 & -- & 5.53\% & 4.66\% & 4.17\% & 1.33\% & 0.82\% \\
Value scale & 0.1 & -- & 3.22\% & 2.76\% & 2.61\% & 0.94\% & 0.67\% \\
Value scale & 0.2 & -- & 1.96\% & 1.74\% & 1.72\% & 0.83\% & 0.62\% \\
\bottomrule
\end{tabular}\end{adjustbox}
\end{table}

At $\varepsilon=0.1$ and the same dyadic grid, the Chernoff interval and the centering reduce keys read by 1.8--1.9 times (S4 $\rightarrow$ KL1). Four levels per octave save another 9--15\% (KL1 $\rightarrow$ KL4), so that KL4 reads 2.1 times fewer keys than S4. At $\varepsilon=0.1$, Bousquet's inequality saves a further 6\% (value scale) to 15\% (output norm). It requires a $d\times d$ Gram matrix and its top eigenvalue at every level, so we did not move it into the kernel. S4 was implemented only for the first two measures. Figure D.5 plots Table D.7a.

\begin{figure}[tbp]
\centering
\includegraphics[width=\linewidth]{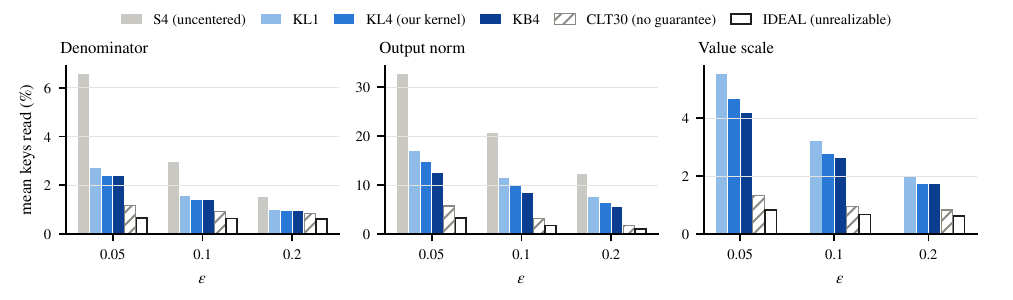}
\caption*{\textbf{Figure D.5.} Mean keys read of the certificate rules of Table D.7a (Llama LongBench 16K). Blue bars are certified rules, and KL4 is the rule of our kernel. CLT30 has no guarantee, and IDEAL is unrealizable.}
\end{figure}

\begin{table}[tbp]
\centering\small
\caption*{\textbf{Table D.7b.} The cost of the guarantee across settings (value scale; KL4 is our kernel's rule). CLT30's worst single-layer violation rate is given for the value scale and for the denominator measure.}
\begin{adjustbox}{max width=\linewidth}\begin{tabular}{lllllllll}
\toprule
Setting & $\varepsilon$ & KL4 & KB4 & CLT30 & IDEAL & KL4 / CLT30 & KL4 / IDEAL & CLT30 worst layer: $M$ / denominator \\
\midrule
Llama LongBench 16K & 0.1 & 2.76\% & 2.61\% & 0.94\% & 0.67\% & 2.95$\times$ & 4.15$\times$ & 0.08\% / 4.8\% \\
Llama RULER 32K & 0.1 & 1.18\% & 1.14\% & 0.40\% & 0.28\% & 2.94$\times$ & 4.26$\times$ & 0.23\% / 2.5\% \\
Llama RULER 128K & 0.1 & 0.43\% & 0.40\% & 0.12\% & 0.09\% & 3.46$\times$ & 4.88$\times$ & 0.62\% / 4.9\% \\
Qwen2.5 LongBench 16K ($G=7$) & 0.1 & 2.52\% & 2.38\% & 1.38\% & 0.57\% & 1.83$\times$ & 4.39$\times$ & 0\% / 2.5\% \\
Llama LongBench 16K & 0.2 & 1.74\% & 1.72\% & 0.83\% & 0.62\% & 2.09$\times$ & 2.79$\times$ & 0\% / 3.9\% \\
Llama RULER 32K & 0.2 & 0.72\% & 0.72\% & 0.37\% & 0.26\% & 1.93$\times$ & 2.70$\times$ & 0.08\% / 3.7\% \\
Llama RULER 128K & 0.2 & 0.24\% & 0.24\% & 0.11\% & 0.08\% & 2.24$\times$ & 3.01$\times$ & 0.16\% / 3.2\% \\
Qwen2.5 LongBench 16K ($G=7$) & 0.2 & 1.59\% & 1.57\% & 1.27\% & 0.53\% & 1.25$\times$ & 2.98$\times$ & 0.18\% / 2.1\% \\
\bottomrule
\end{tabular}\end{adjustbox}
\end{table}

No certified rule has a violation in any setting. CLT30 stays at or below the nominal $\delta=0.05$ in every layer on these data, but its worst layers come close (4.9\% for the denominator at 128K), and it carries no guarantee. A normal-approximation rule without a minimum number of tail samples violated in 11--57\% of queries in a preliminary experiment.

\begin{table}[tbp]
\centering\small
\caption*{\textbf{Table D.7c.} Violation rates of the uncertified CLT30 rule per (layer, query head) across samples and steps (KL4, our rule, has none in any head): overall rate, largest per-head rate (count), and the number of heads whose count exceeds $\delta=0.05$ significantly (one-sided binomial test, Bonferroni over all heads).}
\begin{adjustbox}{max width=\linewidth}\begin{tabular}{llllll}
\toprule
Setting & Measure & $\varepsilon$ & Overall & Largest per head & Significant heads \\
\midrule
Llama $\cdot$ LongBench 16K (40,960 queries) & denominator & 0.05 & 0.0162 & 0.300 (12/40) & 1 \\
Llama $\cdot$ LongBench 16K (40,960 queries) & denominator & 0.1 & 0.0112 & 0.250 (10/40) & 1 \\
Llama $\cdot$ LongBench 16K (40,960 queries) & denominator & 0.2 & 0.0058 & 0.225 (9/40) & 0 \\
Llama $\cdot$ LongBench 16K (40,960 queries) & output norm & 0.05 & 0.0002 & 0.050 (2/40) & 0 \\
Llama $\cdot$ LongBench 16K (40,960 queries) & output norm & 0.1 & 0.0000 & 0.025 (1/40) & 0 \\
Llama $\cdot$ LongBench 16K (40,960 queries) & output norm & 0.2 & 0.0001 & 0.100 (4/40) & 0 \\
Llama $\cdot$ RULER 128K & denominator & 0.05 & 0.0175 & 0.150 (6/40) & 0 \\
Llama $\cdot$ RULER 128K & denominator & 0.1 & 0.0121 & 0.175 (7/40) & 0 \\
Llama $\cdot$ RULER 128K & denominator & 0.2 & 0.0068 & 0.275 (11/40) & 1 \\
Llama $\cdot$ RULER 128K & output norm & 0.05 & 0.0009 & 0.075 (3/40) & 0 \\
Llama $\cdot$ RULER 128K & output norm & 0.1 & 0.0003 & 0.025 (1/40) & 0 \\
Llama $\cdot$ RULER 128K & output norm & 0.2 & 0.0004 & 0.050 (2/40) & 0 \\
Llama $\cdot$ RULER 128K & value scale & 0.05 & 0.0006 & 0.100 (4/40) & 0 \\
Llama $\cdot$ RULER 128K & value scale & 0.1 & 0.0003 & 0.100 (4/40) & 0 \\
Llama $\cdot$ RULER 128K & value scale & 0.2 & 0.0000 & 0.050 (2/40) & 0 \\
Llama $\cdot$ RULER 32K & denominator & 0.05 & 0.0119 & 0.175 (7/40) & 0 \\
Llama $\cdot$ RULER 32K & denominator & 0.1 & 0.0062 & 0.125 (5/40) & 0 \\
Llama $\cdot$ RULER 32K & denominator & 0.2 & 0.0029 & 0.125 (5/40) & 0 \\
Llama $\cdot$ RULER 32K & output norm & 0.05 & 0.0007 & 0.125 (5/40) & 0 \\
Llama $\cdot$ RULER 32K & output norm & 0.1 & 0.0007 & 0.100 (4/40) & 0 \\
Llama $\cdot$ RULER 32K & output norm & 0.2 & 0.0005 & 0.050 (2/40) & 0 \\
Llama $\cdot$ RULER 32K & value scale & 0.05 & 0.0003 & 0.100 (4/40) & 0 \\
Llama $\cdot$ RULER 32K & value scale & 0.1 & 0.0002 & 0.025 (1/40) & 0 \\
Llama $\cdot$ RULER 32K & value scale & 0.2 & 0.0000 & 0.025 (1/40) & 0 \\
Qwen2.5 $\cdot$ LongBench 16K & denominator & 0.05 & 0.0150 & 0.183 (11/60) & 0 \\
Qwen2.5 $\cdot$ LongBench 16K & denominator & 0.1 & 0.0104 & 0.100 (6/60) & 0 \\
Qwen2.5 $\cdot$ LongBench 16K & denominator & 0.2 & 0.0055 & 0.117 (7/60) & 0 \\
Qwen2.5 $\cdot$ LongBench 16K & output norm & 0.05 & 0.0002 & 0.033 (2/60) & 0 \\
Qwen2.5 $\cdot$ LongBench 16K & output norm & 0.1 & 0.0002 & 0.033 (2/60) & 0 \\
Qwen2.5 $\cdot$ LongBench 16K & output norm & 0.2 & 0.0001 & 0.033 (2/60) & 0 \\
Qwen2.5 $\cdot$ LongBench 16K & value scale & 0.05 & 0.0001 & 0.017 (1/60) & 0 \\
Qwen2.5 $\cdot$ LongBench 16K & value scale & 0.1 & 0.0000 & 0.000 (0/60) & 0 \\
Qwen2.5 $\cdot$ LongBench 16K & value scale & 0.2 & 0.0001 & 0.017 (1/60) & 0 \\
\bottomrule
\end{tabular}\end{adjustbox}
\end{table}

\begin{table}[tbp]
\centering\small
\caption*{\textbf{Table D.7d.} Our certified rule (KL4) against the uncertified CLT30 rule at a matched number of keys read: KL4 runs at the target $\varepsilon$ whose mean keys read is closest to CLT30's (within 25\%), on the same sample paths. Keys read (sample only, as in Table D.7a) / 90th-percentile / maximum error in the measure's own units, and the share of queries above CLT30's target.}
\begin{adjustbox}{max width=\linewidth}\begin{tabular}{llllll}
\toprule
Setting & Measure & CLT30 at $\varepsilon$ & CLT30 & KL4 at $\varepsilon$ & KL4 \\
\midrule
Llama $\cdot$ RULER 32K & output norm & 0.05 & 2.48\% / 0.0283 / 0.108 / 0.07\% & 0.2 & 2.74\% / 0.0245 / 0.053 / 0.00\% \\
Llama $\cdot$ RULER 32K & output norm & 0.1 & 1.28\% / 0.0551 / 0.170 / 0.07\% & 0.5 & 1.49\% / 0.0455 / 0.106 / 0.01\% \\
Llama $\cdot$ RULER 32K & value scale & 0.05 & 0.55\% / 0.0266 / 0.095 / 0.03\% & 0.3 & 0.57\% / 0.0242 / 0.057 / 0.00\% \\
Llama $\cdot$ RULER 32K & value scale & 0.1 & 0.40\% / 0.0491 / 0.156 / 0.02\% & 0.5 & 0.45\% / 0.0326 / 0.085 / 0.00\% \\
Llama $\cdot$ RULER 32K & value scale & 0.2 & 0.37\% / 0.0724 / 0.227 / 0.00\% & 0.5 & 0.45\% / 0.0326 / 0.085 / 0.00\% \\
Qwen2.5 $\cdot$ LongBench 16K & output norm & 0.05 & 4.84\% / 0.0267 / 0.056 / 0.02\% & 0.2 & 4.88\% / 0.0217 / 0.052 / 0.00\% \\
Qwen2.5 $\cdot$ LongBench 16K & output norm & 0.1 & 3.12\% / 0.0525 / 0.117 / 0.02\% & 0.5 & 3.05\% / 0.0407 / 0.100 / 0.00\% \\
Qwen2.5 $\cdot$ LongBench 16K & value scale & 0.05 & 1.76\% / 0.0261 / 0.056 / 0.01\% & 0.2 & 1.59\% / 0.0180 / 0.042 / 0.00\% \\
Qwen2.5 $\cdot$ LongBench 16K & value scale & 0.1 & 1.38\% / 0.0503 / 0.095 / 0.00\% & 0.3 & 1.26\% / 0.0237 / 0.048 / 0.00\% \\
Qwen2.5 $\cdot$ LongBench 16K & value scale & 0.2 & 1.27\% / 0.0881 / 0.249 / 0.01\% & 0.3 & 1.26\% / 0.0237 / 0.048 / 0.00\% \\
\bottomrule
\end{tabular}\end{adjustbox}
\end{table}

In every row, KL4 has the lower 90th-percentile error (CLT30's is 1.10--3.72 times KL4's) and the lower maximum (1.08--5.24 times), while reading 0.90--1.22 times CLT30's keys.

\subsection{Sensitivity and perturbation seeds}\label{app:D.8}

\begin{table}[tbp]
\centering\small
\caption*{\textbf{Table D.8.} Changes relative to the default, as ranges over the ten settings (value scale, $\varepsilon=0.1$ unless varied). No setting has a violation.}
\begin{adjustbox}{max width=\linewidth}\begin{tabular}{llll}
\toprule
Parameter & Values & Keys read & p90 error/$M$ \\
\midrule
$\varepsilon$ & 0.05 / 0.1 / 0.2 / 0.3 & 1.55--1.87$\times$ / 1 / 0.61--0.69$\times$ / 0.49--0.57$\times$ & 0.006--0.007 / 0.010--0.012 / 0.017--0.019 / 0.022--0.025 \\
$\delta$ & 0.01 / 0.05 / 0.2 & +7--9\% / 0 / $-$7--8\% & 0.009--0.011 / 0.010--0.012 / 0.011--0.013 \\
Grid density $r$ & 1 / 2 / 4 / 8 & +16--20\% / +5--6\% / 0 / $-$2--3\% & 0.009--0.011 / 0.010--0.011 / 0.010--0.012 / 0.010--0.012 \\
Window & 0 / 64 / 256 & $-$1\% to $-$18\% / 0 / +27--63\% & unchanged \\
\bottomrule
\end{tabular}\end{adjustbox}
\end{table}

\begin{itemize}
\item
Keys read scale roughly as $\varepsilon^{-0.6}$ to $\varepsilon^{-0.9}$. The exponent grows in magnitude with the context: $\varepsilon=0.05$ reads 1.84--1.87 times the default at 128K and 1.55--1.65 times at 16K.
\item
Tightening $\delta$ fivefold, from 0.05 to 0.01, costs about 8\% more keys, as the logarithmic dependence predicts.
\item
\textbf{Seeds.} We reran the main configurations with four further perturbation seeds on Llama RULER 32K and 128K, Qwen3 LongBench 16K and Qwen2.5 LongBench 16K: the value-scale and output-norm measures, $\varepsilon\in\{0.05,0.1,0.2,0.3\}$, 32 configurations per seed. Across the five seeds:
\begin{itemize}
\item
keys read vary by at most 0.25\% (relative range);
\item
the 90th-percentile error varies by at most 0.0002 (1.2\% relative);
\item
the maximum error varies more, being a tail statistic (for example 0.053--0.074 on Llama 128K at $\varepsilon=0.3$), but stays far below $\varepsilon$;
\item
no seed has a violation or a fallback.
\end{itemize}
\end{itemize}

\subsection{End-to-end results}\label{app:D.9}

\begin{table}[tbp]
\centering\small
\caption*{\textbf{Table D.9a.} All end-to-end runs on the full task sets against dense attention (baselines on subsets are in Table D.9b, the aggregation task in Tables D.9c and D.9d). ``Per step'' draws fresh perturbations at every step (our default), and ``fixed'' keeps them per KV head (ablation). The per-step runs of Qwen3 and Qwen2.5, and those of Llama at $\varepsilon=0.2$, are separate runs on the same samples, whose dense outputs are identical to those of the fixed runs. W/T/L counts samples on which our score is higher, equal or lower.}
\begin{adjustbox}{max width=\linewidth}\begin{tabular}{lllllll}
\toprule
Model $\cdot$ data (samples) & Method & Keys read & $\Delta$ score [95\% CI] & W / T / L & Same output & Audit violations \\
\midrule
Llama $\cdot$ LongBench, 6 tasks (300) & per step, $\varepsilon=0.1$ & 5.26\% & $-$0.29 [$-$0.63, $-$0.03] & 18 / 255 / 27 & 0.80 & 0 \\
 & per step, $\varepsilon=0.2$ & 3.78\% & $-$0.39 [$-$0.93, +0.02] & 28 / 244 / 28 & 0.72 & 0 \\
 & fixed, $\varepsilon=0.1$ & 5.26\% & $-$0.07 [$-$0.30, +0.12] & 22 / 255 / 23 & 0.78 & 0 \\
 & fixed, $\varepsilon=0.2$ & 3.78\% & +0.00 [$-$0.24, +0.23] & 23 / 245 / 32 & 0.72 & 0 \\
Llama $\cdot$ RULER 32K, 4 tasks (200) & per step, $\varepsilon=0.1$ & 1.50\% & $-$0.65 [$-$2.05, +0.35] & 2 / 196 / 2 & 0.93 & 0 \\
 & per step, $\varepsilon=0.2$ & 0.95\% & $-$0.65 [$-$2.05, +0.35] & 3 / 194 / 3 & 0.91 & 0 \\
 & fixed, $\varepsilon=0.1$ & 1.50\% & +0.25 [$-$0.30, +0.95] & 4 / 192 / 4 & 0.89 & 0 \\
 & fixed, $\varepsilon=0.2$ & 0.95\% & +0.05 [$-$1.70, +1.85] & 6 / 186 / 8 & 0.88 & 0 \\
Llama $\cdot$ RULER 64K, 4 tasks (100) & per step, $\varepsilon=0.1$ & 0.78\% & 0.00 & 0 / 100 / 0 & 0.98 & 0 \\
 & fixed, $\varepsilon=0.1$ & 0.77\% & $-$0.60 [$-$1.80, 0.00] & 0 / 99 / 1 & 0.98 & 0 \\
 & fixed, $\varepsilon=0.2$ & 0.48\% & $-$0.60 [$-$2.00, +0.40] & 1 / 97 / 2 & 0.95 & 0 \\
Llama $\cdot$ RULER 128K, 4 tasks (48) & per step, $\varepsilon=0.1$ & 0.49\% & 0.00 & 0 / 48 / 0 & 0.90 & 0 \\
 & fixed, $\varepsilon=0.1$ & 0.47\% & 0.00 & 0 / 48 / 0 & 0.96 & 0 \\
 & fixed, $\varepsilon=0.2$ & 0.28\% & 0.00 & 0 / 48 / 0 & 0.94 & 0 \\
Qwen3 $\cdot$ LongBench (150) & per step, $\varepsilon=0.1$ & 5.10\% & +0.17 [+0.06, +0.28] & 16 / 129 / 5 & 0.80 & 0 \\
 & per step, $\varepsilon=0.2$ & 3.77\% & $-$0.01 [$-$0.23, +0.19] & 14 / 126 / 10 & 0.77 & 0 \\
 & fixed, $\varepsilon=0.1$ & 5.11\% & +0.10 [$-$0.05, +0.25] & 17 / 128 / 5 & 0.81 & 0 \\
 & fixed, $\varepsilon=0.2$ & 3.78\% & +0.05 [$-$0.10, +0.20] & 17 / 126 / 7 & 0.78 & 0 \\
Qwen3 $\cdot$ RULER 32K (100) & per step, $\varepsilon=0.1$ & 1.53\% & +1.00 [$-$0.20, +3.10] & 2 / 97 / 1 & 0.85 & 0 \\
 & per step, $\varepsilon=0.2$ & 1.01\% & $-$1.40 [$-$3.90, +0.20] & 2 / 93 / 5 & 0.82 & 0 \\
 & fixed, $\varepsilon=0.1$ & 1.53\% & +1.00 [0.00, +3.00] & 1 / 99 / 0 & 0.89 & 0 \\
 & fixed, $\varepsilon=0.2$ & 1.02\% & +1.10 [$-$0.30, +3.30] & 4 / 95 / 1 & 0.87 & 0 \\
Qwen2.5 $\cdot$ LongBench (150) & per step, $\varepsilon=0.1$ & 5.38\% & +0.07 [$-$0.09, +0.25] & 12 / 125 / 13 & 0.79 & 0 \\
 & per step, $\varepsilon=0.2$ & 3.85\% & $-$0.14 [$-$0.62, +0.23] & 15 / 123 / 12 & 0.76 & 0 \\
 & fixed, $\varepsilon=0.1$ & 5.38\% & +0.28 [$-$0.22, +1.02] & 16 / 124 / 10 & 0.77 & 0 \\
 & fixed, $\varepsilon=0.2$ & 3.86\% & +0.22 [$-$0.41, +1.02] & 15 / 122 / 13 & 0.76 & 0 \\
Qwen2.5 $\cdot$ RULER 32K (100) & per step, $\varepsilon=0.1$ & 1.73\% & $-$0.10 [$-$0.30, 0.00] & 0 / 99 / 1 & 0.83 & 0 \\
 & per step, $\varepsilon=0.2$ & 1.08\% & $-$0.90 [$-$3.30, +0.60] & 4 / 92 / 4 & 0.73 & 0 \\
 & fixed, $\varepsilon=0.1$ & 1.73\% & +0.20 [$-$0.50, +1.20] & 2 / 95 / 3 & 0.81 & 0 \\
 & fixed, $\varepsilon=0.2$ & 1.08\% & +1.10 [$-$0.70, +3.60] & 3 / 94 / 3 & 0.76 & 0 \\
\bottomrule
\end{tabular}\end{adjustbox}
\end{table}

\begin{itemize}
\item
On Llama LongBench with fresh perturbations, the interval excludes zero, but sign and Wilcoxon tests do not ($p=0.23$, $0.16$), and two further seed families give $-$0.24 [$-$0.75, +0.13] and $-$0.11 [$-$0.56, +0.17] (Table D.9e).
\item
On Qwen3 LongBench, our method scores above dense attention with both kinds of perturbations: 16 wins against 5 losses with fresh perturbations (sign test $p=0.027$) and 17 against 5 with fixed ones ($p=0.017$). The mean change is at most +0.17, and among this many comparisons it is likely a chance finding.
\item
An interval end of exactly zero means that a single sample changes score: on Qwen2.5 RULER 32K with fresh perturbations, one sample out of 100 scores lower than with dense attention, and on Qwen3 RULER 32K with fixed perturbations, one scores higher (ours is correct, dense is not).
\item
At $\varepsilon=0.2$, every interval contains zero with either kind of perturbation.
\item
On Llama RULER 128K, dense attention scores 91.7 on niah\_multikey\_2, 75.0 on vt, 50.0 on qa\_1 and 0 on cwe, and all three of our variants match it task by task. On Llama RULER 64K, dense attention scores only 3.6 on cwe, which cannot separate the methods.
\item
Keys read are higher end to end than at the operator level because many LongBench prompts are much shorter than 32K, and the average runs over all decoding steps and layers.
\end{itemize}

\begin{table}[tbp]
\centering\small
\caption*{\textbf{Table D.9b.} Comparison with baselines on short-output subsets (Llama; configurations in Appendix C.6). Our method draws fresh perturbations at every step, in a separate run on the same samples whose dense outputs are identical; the fixed-perturbation ablation ran in the baselines' batch.}
\begin{adjustbox}{max width=\linewidth}\begin{tabular}{lllll}
\toprule
Data & Method & Keys read & $\Delta$ score [95\% CI] & Same output \\
\midrule
LongBench, 5 tasks $\times$ 20 & ours, $\varepsilon=0.1$ & 5.44\% & $-$0.48 [$-$1.34, +0.06] & 0.91 \\
 & ours, fixed perturbations & 5.46\% & $-$0.34 [$-$0.97, +0.07] & 0.88 \\
 & oracle top-$k$ & 5.92\% & +0.36 [$-$0.43, +1.24] & 0.61 \\
 & vAttention (HashAttention) & 7.46\% & $-$0.32 [$-$2.93, +1.80] & 0.52 \\
 & HashAttention & 5.92\% & $-$1.20 [$-$3.13, +0.36] & 0.57 \\
 & Quest & 5.17\% & $-$1.92 [$-$4.86, +0.40] & 0.49 \\
RULER 32K, 3 tasks $\times$ 20 & ours, $\varepsilon=0.1$ & 1.51\% & 0.00 & 1.00 \\
 & ours, fixed perturbations & 1.52\% & +0.67 [0.00, +2.00] & 0.97 \\
 & oracle top-$k$ & 2.80\% & $-$0.33 [$-$1.00, 0.00] & 0.97 \\
 & Quest & 2.49\% & 0.00 [$-$5.00, +5.00] & 0.75 \\
 & HashAttention & 2.80\% & $-$4.00 [$-$9.33, 0.00] & 0.82 \\
 & vAttention (HashAttention) & 4.08\% & $-$7.67 [$-$14.0, $-$2.3] & 0.62 \\
\bottomrule
\end{tabular}\end{adjustbox}
\end{table}

In Table D.9b, with 20 samples per task, the intervals of Quest, HashAttention and vAttention span 3.5--5.3 points on LongBench and 9.3--11.7 points on RULER, so only large losses are detectable.

\begin{table}[tbp]
\centering\small
\caption*{\textbf{Table D.9c.} RULER common-words extraction at 32K: score change against dense attention and the paired difference between our method and oracle top-$k$ reading, at every layer, step and query head, as many keys as ours (95\% bootstrap CIs; sign-test p). Each top-$k$ run is matched to our method of its own run: with fresh perturbations at every step (our default) or with perturbations fixed per KV head (an earlier run, which also has top-$k$ with dense layers 0--1).}
\begin{adjustbox}{max width=\linewidth}\begin{tabular}{lllllll}
\toprule
Model (samples) & Our perturbations & Ours $-$ dense & Top-$k$ $-$ dense & Top-$k$ (layers 0--1 dense) $-$ dense & Ours $-$ top-$k$ & p (sign) \\
\midrule
Llama (50) & per step & $-$2.6 [$-$8.2, +1.4] & $-$15.8 [$-$26.6, $-$5.0] & -- & +13.2 [+3.0, +23.8] & 0.17 \\
 & fixed & +0.2 [$-$1.6, +2.4] & $-$16.0 [$-$27.2, $-$5.0] & $-$15.4 [$-$26.4, $-$4.8] & +16.2 [+5.4, +27.0] & 0.015 \\
Qwen3 (25) & per step & 0.0 [$-$1.2, +1.2] & $-$41.6 [$-$52.0, $-$30.4] & -- & +41.6 [+30.8, +52.0] & 3.6e-05 \\
 & fixed & +0.0 [+0.0, +0.0] & $-$45.2 [$-$54.8, $-$35.2] & $-$46.0 [$-$54.8, $-$36.8] & +45.2 [+35.6, +54.4] & 3e-06 \\
Qwen2.5 (25) & per step & $-$0.4 [$-$1.2, 0.0] & $-$40.4 [$-$50.4, $-$30.8] & -- & +40.0 [+30.4, +50.4] & 2.4e-07 \\
 & fixed & +0.8 [$-$2.0, +4.8] & $-$45.2 [$-$54.8, $-$35.6] & $-$45.6 [$-$55.2, $-$36.0] & +46.0 [+36.0, +55.6] & 2.4e-07 \\
\bottomrule
\end{tabular}\end{adjustbox}
\end{table}

In Table D.9c, dense attention scores 27.8 (Llama), 86.0 (Qwen3) and 74.8 (Qwen2.5). With fresh perturbations, our method scores 25.2, 86.0 and 74.4, and oracle top-$k$ 12.0, 44.4 and 34.4. On Llama, our advantage over top-$k$ has a bootstrap interval above zero, but 9 of 50 samples favor top-$k$ and the sign test does not reject (17 wins, 9 losses). With fixed perturbations, our method scored 28.0, 86.0 and 75.6, and top-$k$ 11.8, 40.8 and 29.6 (12.4, 40.0 and 29.2 with layers 0--1 dense). Our method reads 1.5\% of the keys and top-$k$ 1.6\% (7.0--8.5\% with layers 0--1 dense, which it reads in full), so the loss of top-$k$ comes from the attention mass it discards. The closed-loop audit of our method records no violation.

\begin{table}[tbp]
\centering\small
\caption*{\textbf{Table D.9d.} RULER common-words extraction at 32K against the strongest baselines: score, keys read and bytes (ours measured, with the 4 bytes per token written by per-step perturbations; baselines a lower bound), score change against dense attention and our per-step method's paired advantage, with 95\% CIs. The baselines run at the configurations matched to our end-to-end bytes (Appendix C.6), which here read 1.0--1.3 times our bytes, and on Llama also at the sparser configurations of RetroInfer and vAttention. The matched Llama RetroInfer and vAttention ran in a second run, whose dense and per-step outputs are identical to the first, and are paired with our method in that run. The fixed-perturbation row is the third run described in Appendix F.4.}
\begin{adjustbox}{max width=\linewidth}\begin{tabular}{llllllll}
\toprule
Model (samples) & Method & Score & Keys read & Bytes read & $\Delta$ vs dense [95\% CI] & Ours $-$ method [95\% CI] & p (sign) \\
\midrule
Llama (50) & dense & 27.8 & 100\% & 100\% & -- & -- & -- \\
 & ours, per-step perturbations & 25.2 & 1.5\% & 23.9\% & $-$2.6 [$-$8.2, +1.4] & -- & -- \\
 & ours, fixed perturbations & 30.0 & 1.5\% & 23.2\% & +2.2 [$-$0.8, +6.2] & $-$4.8 [$-$10.4, $-$0.4] & 0.73 \\
 & oracle top-$k$ (our keys per head) & 12.0 & 1.6\% & -- & $-$15.8 [$-$26.6, $-$5.0] & +13.2 [+3.0, +23.8] & 0.17 \\
 & MagicPIG (reference), $K=8$, $L=200$ & 24.8 & 22.8\% & $\ge$ 23.4\% & $-$3.0 [$-$7.2, $-$0.2] & +0.4 [$-$5.4, +5.8] & 0.23 \\
 & RetroInfer (CPU port), retrieval 14.4\% & 28.8 & 13.8\% & $\ge$ 20.0\% & +1.0 [$-$5.6, +7.6] & $-$3.6 [$-$10.6, +2.8] & 0.58 \\
 & RetroInfer (CPU port), retrieval 20\% & 29.0 & 19.2\% & $\ge$ 25.4\% & +1.2 [$-$4.4, +6.8] & $-$3.8 [$-$10.0, +1.2] & 1 \\
 & vAttention (HashAttention), top-$k$ 5\% & 27.2 & 9.2\% & $\ge$ 10.0\% & $-$0.6 [$-$9.0, +7.8] & $-$2.0 [$-$10.4, +5.8] & 1 \\
 & vAttention (HashAttention), top-$k$ 20\% & 27.6 & 25.7\% & $\ge$ 26.5\% & $-$0.2 [$-$7.2, +6.8] & $-$2.4 [$-$11.0, +6.0] & 1 \\
Qwen3 (25) & dense & 86.0 & 100\% & 100\% & -- & -- & -- \\
 & ours, per-step perturbations & 86.0 & 1.4\% & 25.7\% & 0.0 [$-$1.2, +1.2] & -- & -- \\
 & oracle top-$k$ (our keys per head) & 44.4 & 1.6\% & -- & $-$41.6 [$-$52.0, $-$30.4] & +41.6 [+30.8, +52.0] & 3.6e-05 \\
 & MagicPIG (reference), $K=9$, $L=450$ & 83.2 & 28.7\% & $\ge$ 29.4\% & $-$2.8 [$-$5.2, $-$0.8] & +2.8 [+0.4, +6.0] & 0.22 \\
 & RetroInfer (CPU port), retrieval 25\% & 86.0 & 25.6\% & $\ge$ 31.8\% & 0.0 [$-$2.8, +3.6] & 0.0 [$-$3.2, +2.8] & 1 \\
Qwen2.5 (25) & dense & 74.8 & 100\% & 100\% & -- & -- & -- \\
 & ours, per-step perturbations & 74.4 & 1.5\% & 31.1\% & $-$0.4 [$-$1.2, 0.0] & -- & -- \\
 & oracle top-$k$ (our keys per head) & 34.4 & 1.6\% & -- & $-$40.4 [$-$50.4, $-$30.8] & +40.0 [+30.4, +50.4] & 2.4e-07 \\
 & MagicPIG (reference), $K=9$, $L=600$ & 74.8 & 36.9\% & $\ge$ 37.8\% & 0.0 & $-$0.4 [$-$1.2, 0.0] & 1 \\
 & RetroInfer (CPU port), retrieval 35\% & 73.2 & 34.3\% & $\ge$ 40.6\% & $-$1.6 [$-$3.6, 0.0] & +1.2 [$-$0.4, +3.2] & 0.62 \\
\bottomrule
\end{tabular}\end{adjustbox}
\end{table}

In Table D.9d, MagicPIG scores below dense attention on Llama and Qwen3 with bootstrap intervals that exclude zero ($-$3.0 and $-$2.8), and on Qwen3 our paired advantage over it has one too (+2.8 [+0.4, +6.0]); no paired difference between our method and MagicPIG, RetroInfer or vAttention is significant by the sign test. On Llama, the matched RetroInfer and vAttention score 29.0 and 27.6 against 25.2 for our method (ours minus them $-$3.8 [$-$10.0, +1.2] and $-$2.4 [$-$11.0, +6.0]). On this task, our method is thus far ahead of oracle top-$k$ at equal keys and not significantly different from the strongest baselines at equal or larger bytes.

\begin{table}[tbp]
\centering\small
\caption*{\textbf{Table D.9e.} Perturbation seed families on Llama LongBench (6 tasks, 300 samples, $\varepsilon=0.1$): score change against dense attention with task-stratified 95\% CIs, W/T/L and two-sided sign and Wilcoxon p-values. Family 1 is the default seeding (Table D.9a), and families 2 and 3 append a tag to the seed string (Appendix C.6). ``Mean of 3'' averages each sample's score over the three families before differencing.}
\begin{adjustbox}{max width=\linewidth}\begin{tabular}{llllll}
\toprule
Perturbations & Seeds & $\Delta$ score [95\% CI] & W / T / L & p (sign / Wilcoxon) & Same output \\
\midrule
Per step & family 1 & $-$0.29 [$-$0.63, $-$0.03] & 18 / 255 / 27 & 0.23 / 0.16 & 0.80 \\
 & family 2 & $-$0.24 [$-$0.75, +0.13] & 22 / 253 / 25 & 0.77 / 0.68 & 0.79 \\
 & family 3 & $-$0.11 [$-$0.56, +0.17] & 25 / 257 / 18 & 0.36 / 0.29 & 0.78 \\
 & mean of 3 & $-$0.21 [$-$0.56, +0.05] & 25 / 238 / 37 & 0.16 / 0.15 & -- \\
Fixed & family 1 & $-$0.07 [$-$0.30, +0.12] & 22 / 255 / 23 & 1.0 / 0.88 & 0.78 \\
 & family 2 & $-$0.25 [$-$0.72, +0.08] & 23 / 251 / 26 & 0.78 / 0.67 & 0.78 \\
 & family 3 & $-$0.03 [$-$0.24, +0.14] & 25 / 252 / 23 & 0.89 / 0.73 & 0.78 \\
 & mean of 3 & $-$0.12 [$-$0.32, +0.05] & 31 / 236 / 33 & 0.90 / 0.83 & -- \\
Per step $-$ fixed & means of 3 & $-$0.10 [$-$0.35, +0.10] & 26 / 241 / 33 & 0.43 / 0.38 & -- \\
\bottomrule
\end{tabular}\end{adjustbox}
\end{table}

In Table D.9e, only family 1 of the per-step method has an interval that excludes zero, and no sign or Wilcoxon test is significant. Between seed families of the same variant, the change varies as much as between the variants ($-$0.03 to $-$0.25 with fixed perturbations). With fresh perturbations at $\varepsilon=0.2$, the change is $-$0.39 [$-$0.93, +0.02] (28 / 244 / 28).

\begin{table}[tbp]
\centering\small
\caption*{\textbf{Table D.9f.} RULER common-words extraction at 32K with temperature sampling ($T=0.7$, no truncation, four draws per sample; draw $j$ of every method uses the same Gumbel noise at every step). Baselines at the matched configurations of Table D.9d. Score: mean over samples and draws; differences are per sample (mean over the draws), with 95\% bootstrap CIs over samples and the two-sided sign test (wins / losses of our method). Same as dense: share of draws whose whole sequence equals dense attention's sequence of the same draw; first divergence: median step at which it first differs (sequence length if never); KL: mean KL divergence from dense attention's next-token distribution over the steps before the first divergence.}
\begin{adjustbox}{max width=\linewidth}\begin{tabular}{lllllll}
\toprule
Model $\cdot$ method (11 samples $\times$ 4 draws) & Score & Minus dense & Ours minus method (W/L, $p$) & Same as dense & First divergence & KL before divergence \\
\midrule
Llama $\cdot$ dense attention & 19.8 & --- & --- & --- & --- & --- \\
Llama $\cdot$ ours, $\varepsilon=0.1$ & 19.8 & +0.0 [$-$1.4, +1.4] & --- & 77\% & 41 & 4.1e-04 \\
Llama $\cdot$ MagicPIG (reference) & 20.7 & +0.9 [$-$2.3, +4.1] & $-$0.9 [$-$4.3, +2.5] (4/4, 1.00) & 66\% & 41 & 1.1e-03 \\
Llama $\cdot$ RetroInfer (CPU port) & 21.1 & +1.4 [$-$2.0, +6.1] & $-$1.4 [$-$6.4, +2.5] (3/4, 1.00) & 50\% & 41 & 3.3e-03 \\
Llama $\cdot$ vAttention (HashAttention) & 22.7 & +3.0 [$-$0.7, +6.6] & $-$3.0 [$-$6.6, +0.9] (1/5, 0.22) & 61\% & 41 & 2.6e-03 \\
Qwen2.5 $\cdot$ dense attention & 74.1 & --- & --- & --- & --- & --- \\
Qwen2.5 $\cdot$ ours, $\varepsilon=0.1$ & 73.0 & $-$1.1 [$-$3.0, +0.7] & --- & 57\% & 68 & 6.4e-04 \\
Qwen2.5 $\cdot$ MagicPIG (reference) & 73.6 & $-$0.5 [$-$1.8, +0.7] & $-$0.7 [$-$2.5, +0.9] (1/3, 0.62) & 68\% & 71 & 2.8e-04 \\
Qwen2.5 $\cdot$ RetroInfer (CPU port) & 73.4 & $-$0.7 [$-$2.3, +0.9] & $-$0.5 [$-$3.2, +1.8] (3/3, 1.00) & 43\% & 49 & 8.3e-04 \\
\bottomrule
\end{tabular}\end{adjustbox}
\end{table}

In Table D.9f, no score difference against dense attention or between our method and a baseline has a bootstrap interval that excludes zero or a significant sign test; with 11 samples, the intervals are 3--10 points wide. The share of draws equal to dense attention's, ours minus the baseline (95\% bootstrap CI over samples, in points), is +11 [$-$9, +30] against MagicPIG, +27 [+14, +41] against RetroInfer and +16 [0, +32] against vAttention on Llama, and $-$11 [$-$34, +9] against MagicPIG and +14 [$-$9, +36] against RetroInfer on Qwen2.5. The KL divergence before the first divergence averages each method over its own steps, so a method that diverges early contributes fewer, earlier steps. The closed-loop audit of our method records no violation.

\subsection{Index construction and append cost}\label{app:D.10}

\begin{table}[tbp]
\centering\small
\caption*{\textbf{Table D.10.} Indexing the prompt (all KV heads of all layers) and appending one token per KV head per step, against the time of our attention query in the same step (value scale, $\varepsilon=0.1$). Measured on one NUMA node of an idle machine. On the 8358, 24 threads use hyper-threads.}
\begin{tabularx}{\linewidth}{ll>{\raggedright\arraybackslash}X>{\raggedright\arraybackslash}X>{\raggedright\arraybackslash}X>{\raggedright\arraybackslash}X}
\toprule
CPU & Model $\cdot$ data & Indexing: per KV head / all heads / keys per s (1 thread) & Append per step (all heads) & Query: 1 / 8 / 24 threads & Append / query (24 threads) \\
\midrule
8555C & Llama $\cdot$ 16K & 0.030 s / 7.8 s / 536K & 0.70 ms & 297 / 37 / 13 ms & 5.2\% \\
8555C & Llama $\cdot$ 32K & 0.060 s / 15.5 s / 535K & 0.69 ms & 348 / 45 / 17 ms & 4.1\% \\
8555C & Llama $\cdot$ 64K & 0.122 s / 31.4 s / 528K & 0.73 ms & 516 / 66 / 25 ms & 2.9\% \\
8555C & Llama $\cdot$ 128K & 0.270 s / 69.2 s / 483K & 0.71 ms & 1039 / 133 / 60 ms & 1.2\% \\
8555C & Qwen2.5 $\cdot$ 16K & 0.030 s / 3.4 s / 539K & 0.31 ms & 194 / 26 / 9 ms & 3.3\% \\
8358 & Llama $\cdot$ 16K & 0.028 s / 7.1 s / 586K & 0.96 ms & 313 / 40 / 22 ms & 4.3\% \\
8358 & Llama $\cdot$ 128K & 0.242 s / 61.9 s / 540K & 0.90 ms & 1089 / 190 / 136 ms & 0.7\% \\
\bottomrule
\end{tabularx}
\end{table}

Appending costs 2.7--3.7 $\mu$s per KV head and step, independent of the context length. Indexing is linear but cheap: single-threaded, about half a million keys per second. With the parallel construction of Appendix B.4, the first decoding step, including indexing, takes 0.9--3.1 s (Table D.11b), compared with 226--1981 s for the fp32 prefill at 32K--128K on the 8358.

\subsection{Attention and decoding-step time}\label{app:D.11}

\begin{table}[tbp]
\centering\small
\caption*{\textbf{Table D.11a.} Attention time of one decoding step (all layers and KV heads of one sample) on 26 cores of one NUMA node (8555C) or the 32 cores of one socket (8358), as speedups over our dense kernel at 1 / 24 threads. Perturbations are drawn afresh at every call, our default, except in the column ``fixed'' (drawn once per KV head). Our dense kernel and all configurations are timed in turn within each of three repetitions, and each uses its best repetition. ``Denom.'': the denominator measure; ``one int8'': one int8 level for the query weights instead of two.}
\begin{adjustbox}{max width=\linewidth}\begin{tabular}{lllllllll}
\toprule
CPU $\cdot$ model $\cdot$ data & Dense (ms) & $M$, 0.1 & $M$, 0.1, fixed & $M$, 0.2 & $\|o\|$, 0.1 & Denom., 0.1 & $M$, 0.1, no 8-bit & $M$, 0.1, one int8 \\
\midrule
8555C $\cdot$ Llama $\cdot$ LongBench 16K & 347 / 22.2 & 1.19 / 1.61 & 1.22 / 1.60 & 1.56 / 2.14 & 0.56 / 0.75 & 2.01 / 2.73 & 1.18 / 1.61 & 1.18 / 1.61 \\
8555C $\cdot$ Llama $\cdot$ RULER 32K & 710 / 46.1 & 1.49 / 2.06 & 1.56 / 2.08 & 1.93 / 2.67 & 0.75 / 0.95 & 2.27 / 3.07 & 1.51 / 2.06 & 1.48 / 2.02 \\
8555C $\cdot$ Llama $\cdot$ RULER 64K & 1432 / 94.1 & 2.07 / 2.67 & 2.21 / 2.77 & 2.50 / 3.29 & 1.16 / 1.34 & 2.79 / 3.71 & 2.05 / 2.75 & 2.10 / 2.73 \\
8555C $\cdot$ Llama $\cdot$ RULER 128K & 2923 / 202 & 2.30 / 2.64 & 2.47 / 2.76 & 2.73 / 3.24 & 1.43 / 1.51 & 2.99 / 3.50 & 2.30 / 2.72 & 2.32 / 2.54 \\
8555C $\cdot$ Llama $\cdot$ LongBench-v2 128K & 2928 / 202 & 2.89 / 3.50 & 3.11 / 3.70 & 3.24 / 4.04 & 1.74 / 1.84 & 3.50 / 4.26 & 2.89 / 3.61 & 2.98 / 3.52 \\
8555C $\cdot$ Qwen3 $\cdot$ LongBench 16K & 391 / 24.4 & 1.25 / 1.51 & 1.28 / 1.50 & 1.53 / 1.88 & 0.80 / 0.99 & 1.93 / 2.37 & 1.23 / 1.52 & 1.22 / 1.48 \\
8555C $\cdot$ Qwen2.5 $\cdot$ LongBench 16K & 225 / 14.0 & 0.87 / 1.09 & 0.88 / 1.05 & 1.13 / 1.43 & 0.49 / 0.57 & 1.37 / 1.69 & 0.86 / 1.06 & 0.85 / 1.06 \\
8358 $\cdot$ Llama $\cdot$ LongBench 16K & 333 / 22.5 & 1.02 / 1.41 & 1.07 / 1.44 & 1.35 / 1.88 & 0.48 / 0.61 & 1.76 / 2.40 & 1.04 / 1.43 & 1.01 / 1.38 \\
8358 $\cdot$ Llama $\cdot$ RULER 32K & 660 / 44.5 & 1.29 / 1.60 & 1.35 / 1.66 & 1.66 / 2.09 & 0.62 / 0.71 & 1.96 / 2.47 & 1.28 / 1.59 & 1.29 / 1.57 \\
8358 $\cdot$ Llama $\cdot$ RULER 64K & 1323 / 94.4 & 1.75 / 2.01 & 1.88 / 2.11 & 2.16 / 2.45 & 0.95 / 1.03 & 2.44 / 2.73 & 1.77 / 1.96 & 1.80 / 1.98 \\
8358 $\cdot$ Llama $\cdot$ RULER 128K & 2809 / 209 & 1.95 / 2.13 & 2.13 / 2.22 & 2.33 / 2.53 & 1.18 / 1.28 & 2.52 / 2.72 & 1.95 / 2.11 & 1.96 / 2.09 \\
8358 $\cdot$ Llama $\cdot$ LongBench-v2 128K & 2833 / 210 & 2.46 / 2.64 & 2.74 / 2.81 & 2.74 / 2.96 & 1.45 / 1.53 & 3.01 / 3.15 & 2.45 / 2.67 & 2.57 / 2.66 \\
8358 $\cdot$ Qwen3 $\cdot$ LongBench 16K & 377 / 25.0 & 1.08 / 1.31 & 1.13 / 1.34 & 1.33 / 1.63 & 0.69 / 0.83 & 1.71 / 2.06 & 1.09 / 1.35 & 1.07 / 1.29 \\
8358 $\cdot$ Qwen2.5 $\cdot$ LongBench 16K & 258 / 16.7 & 0.87 / 1.04 & 0.89 / 1.06 & 1.14 / 1.37 & 0.49 / 0.53 & 1.40 / 1.68 & 0.88 / 1.04 & 0.85 / 1.00 \\
\bottomrule
\end{tabular}\end{adjustbox}
\end{table}

\begin{itemize}
\item
The speedup grows with the context length, as bytes read fall from 29\% to 18\%.
\item
The 8555C gains more than the 8358 in every row except Qwen2.5 on one thread. Both have AVX-512 VNNI; we attribute the difference to memory bandwidth and caches but did not measure it separately.
\item
From 1 to 24 threads the speedup grows (by 8--38\%), so our kernel scales at least as well as the dense kernel.
\item
Fresh perturbations take 0.96--1.14 times the time of fixed ones (median 1.04; medians over repetitions of the ratio within a repetition). The cost is largest on one thread at 128K (1.07--1.14) and at most 1.06 on 24 threads.
\item
The output-norm measure is slower than dense attention at 32K and below, and at 64K on one thread of the 8358.
\item
No configuration falls back to dense attention.
\end{itemize}

\begin{table}[tbp]
\centering\small
\caption*{\textbf{Table D.11b.} Whole fp32 decoding step of Llama (one RULER sequence per length; $\varepsilon=0.1$, value scale, perturbations drawn afresh at every step). After one dense prefill, three methods decode 32 tokens each from the prompt's cache, one after another, in five rounds: the model's attention over its fp32 cache (SDPA), our dense kernel over the same 16-bit cache as our method (dense16), and our method. ``Steady'' is the median of steps 2--32 in each method's fastest round. ``Attention'' is the time of all attention calls of a steady step. ``Rest'' is the remainder of the dense16 step (linear layers, sampling, cache update). ``Break-even'' is the number of steady steps after which the extra cost of our first step, which indexes the prompt, is recovered.}
\begin{tabularx}{\linewidth}{>{\raggedright\arraybackslash}X>{\raggedright\arraybackslash}X>{\raggedright\arraybackslash}Xl>{\raggedright\arraybackslash}X>{\raggedright\arraybackslash}X>{\raggedright\arraybackslash}Xl}
\toprule
CPU (threads) $\cdot$ context & Steady step: SDPA / dense16 / ours & Attention: SDPA / dense16 / ours & Rest & Step speedup over dense16 / SDPA & Attention speedup over dense16 & Ours: first step & Break-even \\
\midrule
8555C (26) $\cdot$ 32K & 421 / 306 / 273 ms & 219 / 105 / 66 ms & 202 ms & 1.12$\times$ / 1.54$\times$ & 1.58$\times$ & 0.91 s & 18 steps \\
8555C (26) $\cdot$ 64K & 715 / 416 / 314 ms & 508 / 210 / 104 ms & 206 ms & 1.32$\times$ / 2.27$\times$ & 2.02$\times$ & 1.56 s & 11 steps \\
8555C (26) $\cdot$ 128K & 1192 / 629 / 439 ms & 976 / 412 / 218 ms & 217 ms & 1.43$\times$ / 2.72$\times$ & 1.89$\times$ & 3.15 s & 13 steps \\
8358 (32) $\cdot$ 32K & 343 / 330 / 298 ms & 128 / 115 / 81 ms & 215 ms & 1.11$\times$ / 1.15$\times$ & 1.42$\times$ & 0.91 s & 18 steps \\
8358 (32) $\cdot$ 64K & 470 / 434 / 347 ms & 250 / 219 / 126 ms & 215 ms & 1.25$\times$ / 1.35$\times$ & 1.74$\times$ & 1.57 s & 13 steps \\
8358 (32) $\cdot$ 128K & 724 / 649 / 479 ms & 503 / 432 / 257 ms & 217 ms & 1.35$\times$ / 1.51$\times$ & 1.68$\times$ & 2.92 s & 13 steps \\
\bottomrule
\end{tabularx}
\end{table}

\begin{figure}[tbp]
\centering
\includegraphics[width=\linewidth]{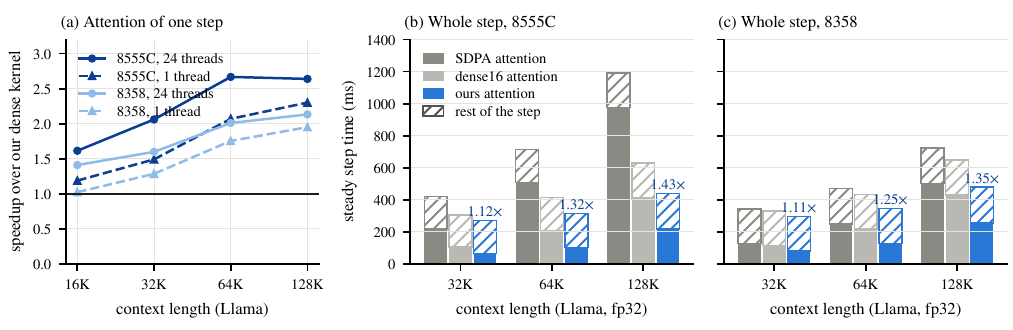}
\caption*{\textbf{Figure D.7.} CPU speed on Llama (value scale, $\varepsilon=0.1$, fresh perturbations at every step). \emph{(a)} Attention for all heads of one decoding step, as speedup over our dense kernel on the same 16-bit keys and values (LongBench at 16K, RULER at 32K--128K; Table D.11a). \emph{(b, c)} Steady time of a whole fp32 decoding step for the model's attention over its fp32 cache (SDPA), our dense kernel (dense16) and our method, split into attention and the rest of the step (fastest of five rounds; Table D.11b). The numbers give our step speedup over dense16.}
\end{figure}

\begin{itemize}
\item
The model's attention over its fp32 cache is 1.1--1.2 (8358) and 2.1--2.4 (8555C) times slower than dense16 on the same exact computation, so speedups over SDPA (1.2--2.7 times per step) mix precision and implementation with sparsity. We therefore report the speedup over dense16 in the main text.
\item
The linear layers take 200--220 ms of every step regardless of the attention method, which caps the whole-step gain at batch size 1.
\item
Inside a step, layers are called one at a time, because each layer's query depends on the previous layer, so each attention call parallelizes over the 8 KV heads of one layer and leaves most threads idle for both kernels. Measured alone over all heads of a step (Table D.11a), attention is faster for both kernels and our speedup is larger (2.0--3.5$\times$ at 128K, against 1.7--1.9$\times$ inside the step). Splitting a head's scan across threads would lift this limit, but we have not implemented it.
\item
Over the five rounds, each method's steady step varied by at most 5\%, except ours at 128K (7\% on the 8555C, 10\% on the 8358), and the median over rounds of the speedup within a round differs from the table's by at most 0.03.
\item
Our generated tokens match those of dense16 in every round of all six runs, and those of SDPA at 64K and 128K. At 32K, our method and dense16 depart from SDPA after the answer, through 16-bit rounding. All three methods answer correctly at every length.
\end{itemize}

\begin{table}[tbp]
\centering\small
\caption*{\textbf{Table D.11c.} Attention of one Llama-3.1-8B decoding step (32 layers $\times$ 8 KV heads, 16-bit K/V) on the 8358 (16 cores of one socket): our dense kernel against llama.cpp's CPU flash-attention operator.}
\begin{adjustbox}{max width=\linewidth}\begin{tabular}{lllllll}
\toprule
Context & Ours, 1 thread (ms) & llama.cpp, 1 thread (ms) & Ratio & Ours, 16 threads (ms) & llama.cpp, 16 threads (ms) & Ratio \\
\midrule
16K & 356 & 687 & 1.93$\times$ & 28.8 & 47.9 & 1.66$\times$ \\
32K & 717 & 1405 & 1.96$\times$ & 58.8 & 98.7 & 1.68$\times$ \\
64K & 1446 & 2877 & 1.99$\times$ & 121.4 & 210.2 & 1.73$\times$ \\
128K & 2865 & 5961 & 2.08$\times$ & 252.5 & 476.7 & 1.89$\times$ \\
\bottomrule
\end{tabular}\end{adjustbox}
\end{table}

llama.cpp was built for the host CPU, and its operator was timed with \texttt{test-backend-ops} in perf mode, with two local changes (the thread count and the Llama-3.1-8B decode shapes). Both kernels ran back to back on cores 16--31 of one socket of an 8358. Our single-threaded times are 2--9\% above those of Table D.11a, whose contexts are up to 1\% shorter and which ran on another 8358 machine.

\begin{table}[tbp]
\centering\small
\caption*{\textbf{Table D.11d.} Whole fp32 decoding step of Llama with perturbations fixed per KV head, drawn afresh at every step (our default) and drawn afresh with $\delta=0.05/(HT)$ ($9.5\times10^{-8}$ per call), against dense16. All four methods ran back to back in one session on 16 threads of one NUMA node of an 8358 machine, one RULER sequence per length, while its other NUMA nodes were busy. Columns as in Table D.11b.}
\begin{adjustbox}{max width=\linewidth}\begin{tabular}{lllllll}
\toprule
Context $\cdot$ method & Steady step & Attention & Speedup over dense16: step / attention & Ratio to fixed: step / attention & First step & Keys read \\
\midrule
32K $\cdot$ dense16 & 791 ms & 171 ms & -- & -- & 2.49 s & 100\% \\
32K $\cdot$ fixed & 590 ms & 104 ms & 1.34$\times$ / 1.64$\times$ & -- & 1.78 s & 1.47\% \\
32K $\cdot$ per step & 582 ms & 102 ms & 1.36$\times$ / 1.67$\times$ & 0.99 / 0.98 & 1.78 s & 1.43\% \\
32K $\cdot$ per step, $\delta/(HT)$ & 715 ms & 143 ms & 1.11$\times$ / 1.20$\times$ & 1.21 / 1.37 & 2.09 s & 2.13\% \\
128K $\cdot$ dense16 & 917 ms & 510 ms & -- & -- & 5.45 s & 100\% \\
128K $\cdot$ fixed & 782 ms & 316 ms & 1.17$\times$ / 1.61$\times$ & -- & 5.36 s & 0.75\% \\
128K $\cdot$ per step & 766 ms & 328 ms & 1.20$\times$ / 1.55$\times$ & 0.98 / 1.04 & 5.62 s & 0.75\% \\
128K $\cdot$ per step, $\delta/(HT)$ & 827 ms & 400 ms & 1.11$\times$ / 1.27$\times$ & 1.06 / 1.27 & 5.96 s & 1.07\% \\
\bottomrule
\end{tabular}\end{adjustbox}
\end{table}

Fresh perturbations cost no measurable time: the step and attention times differ from those with fixed perturbations by $-$2\% to +4\%, within the variation between methods that do the same work. Measured alone with interleaved repetitions, fresh perturbations add a median 4\% to the attention time (Table D.11e). The per-step method and $\delta/(HT)$ generate the same tokens as dense16 at both lengths, and fixed perturbations depart from them at 32K without changing the answer. All methods answer correctly. Speedups over dense16 are lower than in Table D.11b at 128K (1.17--1.20$\times$ against 1.35--1.43$\times$), with half the threads and other jobs on the machine.

\begin{table}[tbp]
\centering\small
\caption*{\textbf{Table D.11e.} Attention time of one decoding step (all layers and KV heads of one sample; value scale, $\varepsilon=0.1$) with perturbations fixed per KV head, drawn afresh at every step (our default) and drawn afresh with $\delta=0.05/(HT)$, at 1 / 8 / 16 threads on one NUMA node of an 8358 machine whose other nodes were busy (samples of Table D.11a). Our dense kernel and the three configurations are timed in turn within each of 7 repetitions. Speedups over dense use the best repetition of each, and the two time ratios are medians over repetitions of the ratio within a repetition, which cancels load that changes between repetitions. Keys read are the fraction of the context.}
\begin{tabularx}{\linewidth}{>{\raggedright\arraybackslash}X>{\raggedright\arraybackslash}X>{\raggedright\arraybackslash}X>{\raggedright\arraybackslash}X>{\raggedright\arraybackslash}X>{\raggedright\arraybackslash}X>{\raggedright\arraybackslash}X}
\toprule
Model $\cdot$ data & Dense (ms) & Speedup, fixed & Speedup, per step & Per step ÷ fixed & $\delta/(HT)$ ÷ per step & Keys: per step / $\delta/(HT)$ \\
\midrule
Llama $\cdot$ LongBench 16K & 356 / 56.8 / 38.6 & 1.09 / 1.34 / 1.54 & 1.06 / 1.27 / 1.51 & 1.04 / 1.05 / 1.04 & 1.32 / 1.32 / 1.34 & 2.75\% / 4.41\% \\
Llama $\cdot$ RULER 32K & 702 / 112.8 / 81.2 & 1.37 / 1.53 / 1.77 & 1.32 / 1.47 / 1.70 & 1.06 / 1.03 / 1.04 & 1.31 / 1.31 / 1.34 & 1.68\% / 2.92\% \\
Llama $\cdot$ RULER 64K & 1385 / 235 / 175.1 & 1.82 / 1.97 / 2.15 & 1.69 / 1.82 / 2.05 & 1.08 / 1.06 / 1.04 & 1.25 / 1.27 / 1.25 & 0.89\% / 1.49\% \\
Llama $\cdot$ RULER 128K & 2932 / 514 / 385 & 2.01 / 1.99 / 2.23 & 1.79 / 1.88 / 2.16 & 1.10 / 1.05 / 1.03 & 1.22 / 1.22 / 1.24 & 0.63\% / 1.04\% \\
Llama $\cdot$ LongBench-v2 128K & 2927 / 521 / 386 & 2.62 / 2.58 / 2.86 & 2.32 / 2.41 / 2.70 & 1.13 / 1.06 / 1.04 & 1.17 / 1.17 / 1.17 & 0.39\% / 0.64\% \\
Qwen3 $\cdot$ LongBench 16K & 398 / 63.1 / 46.0 & 1.14 / 1.25 / 1.58 & 1.08 / 1.22 / 1.47 & 1.05 / 1.03 / 1.03 & 1.26 / 1.25 / 1.27 & 1.99\% / 3.03\% \\
Qwen2.5 $\cdot$ LongBench 16K & 270 / 41.5 / 30.5 & 0.89 / 0.98 / 1.18 & 0.88 / 0.96 / 1.22 & 1.02 / 1.02 / 0.99 & 1.27 / 1.31 / 1.29 & 3.13\% / 4.88\% \\
\bottomrule
\end{tabularx}
\end{table}

\begin{itemize}
\item
Fresh perturbations add 10--13\% on one thread at 128K and $-$1\% to +6\% on 8 and 16 threads. No configuration falls back to dense attention.
\item
With $\delta/(HT)$, attention reads 1.5--1.7 times the keys and is slower than our dense kernel at 16K on one and eight threads (0.69--0.97$\times$), and on Qwen2.5 also on 16 threads (0.90$\times$).
\item
On one thread, speedups are within 0.12 (fixed perturbations) and 0.16 (per step) of those on the 8358 in Table D.11a, with the largest gaps at 128K, where this table's runs shared the machine with other jobs.
\end{itemize}

\subsection{Perplexity}\label{app:D.12}

\begin{table}[tbp]
\centering\small
\caption*{\textbf{Table D.12.} PG-19 perplexity change relative to dense attention, with 95\% CIs (bootstrap over books), and the mean KL divergence from dense attention's next-token distribution. Our method draws fresh perturbations at every step; the ablation with perturbations fixed per KV head is given below the table. The top-$k$ control reads, per head, as many keys as our method at $\varepsilon=0.1$.}
\begin{tabularx}{\linewidth}{>{\raggedright\arraybackslash}Xl>{\raggedright\arraybackslash}X>{\raggedright\arraybackslash}X>{\raggedright\arraybackslash}X>{\raggedright\arraybackslash}X}
\toprule
Model $\cdot$ context (books) & Dense PPL & Ours, $\varepsilon=0.1$: keys / $\Delta$PPL / KL & Ours, $\varepsilon=0.2$: keys / $\Delta$PPL / KL & Oracle top-$k$: $\Delta$PPL / KL & Top-$k$, layers 0--1 dense: $\Delta$PPL \\
\midrule
Llama $\cdot$ 32K (16) & 11.076 & 1.79\% / +0.03\% [$-$0.06, +0.13] / 0.0005 & 1.10\% / +0.22\% [+0.07, +0.36] / 0.0014 & +5.9\% [+3.1, +9.3] / 0.095 & +5.2\% [+2.3, +8.7] \\
Llama $\cdot$ 64K (12) & 11.487 & 0.98\% / +0.07\% [$-$0.05, +0.19] / 0.0006 & 0.59\% / +0.03\% [$-$0.10, +0.18] / 0.0014 & +4.8\% [+2.7, +7.3] / 0.085 & +4.1\% [+2.0, +6.7] \\
Llama $\cdot$ 128K (8) & 8.757 & 0.58\% / +0.01\% [$-$0.03, +0.05] / 0.0005 & 0.34\% / +0.11\% [$-$0.08, +0.31] / 0.0013 & +8.8\% [+3.4, +18.7] / 0.106 & +8.5\% [+3.1, +18.3] \\
Qwen3 $\cdot$ 32K (16) & 12.536 & 1.95\% / +0.05\% [$-$0.03, +0.13] / 0.0003 & 1.26\% / +0.12\% [+0.01, +0.22] / 0.0007 & +8.4\% [+4.6, +14.2] / 0.114 & +8.7\% [+4.9, +14.6] \\
Qwen2.5 $\cdot$ 32K (16) & 10.071 & 2.25\% / $-$0.02\% [$-$0.06, +0.02] / 0.0002 & 1.35\% / $-$0.02\% [$-$0.11, +0.06] / 0.0006 & +11.5\% [+5.7, +19.4] / 0.126 & +10.3\% [+4.5, +18.4] \\
\bottomrule
\end{tabularx}
\end{table}

At $\varepsilon=0.1$, top-1 agreement with dense attention is 98.9--99.2\%, against 87--88\% for oracle top-$k$, and the closed-loop audit records no violation. At $\varepsilon=0.2$, the interval excludes zero only on Llama 32K (+0.22\% [+0.07, +0.36]) and Qwen3 32K (+0.12\% [+0.01, +0.22]). With perturbations fixed per KV head, run in the same batch at 32K, perplexity at $\varepsilon=0.1$ changes by +0.03\% [$-$0.08, +0.12] on Llama, +0.04\% [$-$0.09, +0.17] on Qwen3 and +0.01\% [$-$0.06, +0.07] on Qwen2.5, and the KL divergence is 1.78 [1.25, 2.40], 1.57 [1.30, 1.87] and 1.61 [1.44, 1.79] times that of fresh perturbations, higher on 15, 15 and 16 of 16 books (Table D.12b). On Llama at 64K and 128K, the fixed variant comes from an earlier run on the same books, whose dense negative log-likelihoods differ from this run's by at most $2\cdot10^{-5}$ and $1.8\cdot10^{-4}$. It changes perplexity by $-$0.01\% [$-$0.14, +0.11] and +0.05\% [$-$0.08, +0.18] against dense attention and by $-$0.08\% [$-$0.25, +0.08] and +0.05\% [$-$0.09, +0.18] against fresh perturbations, and its KL divergence is 1.60 [1.36, 1.93] and 1.41 [1.27, 1.54] times theirs, higher on all 12 and all 8 books. In the earlier runs of all five settings, fixed perturbations at $\varepsilon=0.2$ changed perplexity by +0.06\% to +0.16\%, with an interval that excludes zero only on Qwen2.5 (+0.13\% [+0.02, +0.24]). Top-$k$ reads slightly more keys than our method because every head reads at least the 256 sink and local tokens. Its variant with dense layers 0--1 reads 6.7--9.1\% of the keys, since those two layers are read in full.

\begin{table}[tbp]
\centering\small
\caption*{\textbf{Table D.12b.} PG-19 at 32K (and Llama at 64K and 128K) with the per-step method as the reference: keys and bytes read (ours with the 4 bytes per token written by per-step perturbations; baselines: lower bound), perplexity change against dense attention and against our method in percent, mean KL divergence from dense attention's next-token distribution, and the ratio of mean KL divergences to ours, with 95\% CIs from a paired bootstrap over books. Baselines whose bytes are within 0.9--1.1 times ours run at the configurations matched to our end-to-end bytes; the others run at the sparser configurations of Appendix C.6.}
\begin{tabularx}{\linewidth}{>{\raggedright\arraybackslash}X>{\raggedright\arraybackslash}Xll>{\raggedright\arraybackslash}X>{\raggedright\arraybackslash}Xl>{\raggedright\arraybackslash}X}
\toprule
Model $\cdot$ context (books) & Method & Keys read & Bytes read & Perplexity vs dense [95\% CI] & Perplexity vs ours [95\% CI] & KL mean & KL / ours [95\% CI] \\
\midrule
Llama $\cdot$ 32K (16) & ours, per-step perturbations & 1.79\% & 26.0\% & +0.03 [$-$0.06, +0.13] & -- & 0.0005 & 1 \\
Llama $\cdot$ 32K (16) & ours, fixed perturbations & 1.79\% & 25.2\% & +0.03 [$-$0.08, +0.12] & 0.00 [$-$0.18, +0.16] & 0.0009 & 1.78$\times$ [1.25, 2.40] \\
Llama $\cdot$ 32K (16) & ours, $\varepsilon=0.2$, per step & 1.10\% & 22.3\% & +0.22 [+0.07, +0.36] & +0.19 [+0.06, +0.32] & 0.0014 & 2.83$\times$ [2.58, 3.21] \\
Llama $\cdot$ 32K (16) & oracle top-$k$ (our keys per head) & 1.84\% & oracle & +5.89 [+3.09, +9.26] & +5.85 [+3.09, +9.19] & 0.0951 & 188$\times$ [145, 236] \\
Llama $\cdot$ 32K (16) & oracle top-$k$, layers 0--1 dense & 7.80\% & oracle & +5.20 [+2.32, +8.71] & +5.17 [+2.31, +8.67] & 0.0911 & 180$\times$ [136, 227] \\
Llama $\cdot$ 32K (16) & MagicPIG (reference), $K=8$, $L=200$ & 23.41\% & 24.0\% ($\ge$) & +0.09 [$-$0.02, +0.19] & +0.06 [$-$0.11, +0.21] & 0.0009 & 1.83$\times$ [1.67, 2.01] \\
Llama $\cdot$ 32K (16) & RetroInfer (CPU port), retrieval 14.4\% & 14.81\% & 21.1\% ($\ge$) & +0.26 [$-$0.01, +0.52] & +0.23 [$-$0.09, +0.52] & 0.0039 & 7.67$\times$ [6.24, 9.12] \\
Llama $\cdot$ 32K (16) & RetroInfer (CPU port), retrieval 20\% & 20.26\% & 26.5\% ($\ge$) & +0.11 [$-$0.12, +0.31] & +0.07 [$-$0.20, +0.32] & 0.0024 & 4.69$\times$ [3.65, 5.72] \\
Llama $\cdot$ 32K (16) & vAttention (HashAttention), top-$k$ 5\% & 9.69\% & 10.5\% ($\ge$) & +1.24 [+0.79, +1.70] & +1.21 [+0.77, +1.64] & 0.0122 & 24.1$\times$ [20.3, 28.0] \\
Llama $\cdot$ 32K (16) & vAttention (HashAttention), top-$k$ 20\% & 25.89\% & 26.7\% ($\ge$) & +0.17 [+0.03, +0.34] & +0.14 [$-$0.03, +0.34] & 0.0019 & 3.84$\times$ [3.38, 4.21] \\
\bottomrule
\end{tabularx}
\end{table}

\begin{table}[tbp]
\centering\small
\caption*{\textbf{Table D.12b} (continued)}
\begin{tabularx}{\linewidth}{>{\raggedright\arraybackslash}X>{\raggedright\arraybackslash}Xll>{\raggedright\arraybackslash}X>{\raggedright\arraybackslash}Xl>{\raggedright\arraybackslash}X}
\toprule
Model $\cdot$ context (books) & Method & Keys read & Bytes read & Perplexity vs dense [95\% CI] & Perplexity vs ours [95\% CI] & KL mean & KL / ours [95\% CI] \\
\midrule
Llama $\cdot$ 64K (12) & ours, per-step perturbations & 0.98\% & 22.5\% & +0.07 [$-$0.05, +0.19] & -- & 0.0006 & 1 \\
Llama $\cdot$ 64K (12) & ours, $\varepsilon=0.2$, per step & 0.59\% & 20.2\% & +0.03 [$-$0.10, +0.18] & $-$0.04 [$-$0.14, +0.06] & 0.0014 & 2.46$\times$ [2.05, 2.92] \\
Llama $\cdot$ 64K (12) & oracle top-$k$ (our keys per head) & 1.00\% & oracle & +4.80 [+2.67, +7.32] & +4.73 [+2.64, +7.19] & 0.0847 & 148$\times$ [101, 214] \\
Llama $\cdot$ 64K (12) & oracle top-$k$, layers 0--1 dense & 7.09\% & oracle & +4.12 [+1.95, +6.67] & +4.04 [+1.93, +6.58] & 0.0794 & 138$\times$ [92.9, 203] \\
Llama $\cdot$ 128K (8) & ours, per-step perturbations & 0.58\% & 20.6\% & +0.01 [$-$0.03, +0.05] & -- & 0.0005 & 1 \\
Llama $\cdot$ 128K (8) & ours, $\varepsilon=0.2$, per step & 0.34\% & 18.9\% & +0.11 [$-$0.08, +0.31] & +0.10 [$-$0.08, +0.27] & 0.0013 & 2.61$\times$ [2.35, 2.90] \\
Llama $\cdot$ 128K (8) & oracle top-$k$ (our keys per head) & 0.59\% & oracle & +8.85 [+3.36, +18.68] & +8.84 [+3.36, +18.69] & 0.1060 & 211$\times$ [118, 382] \\
Llama $\cdot$ 128K (8) & oracle top-$k$, layers 0--1 dense & 6.73\% & oracle & +8.47 [+3.11, +18.32] & +8.46 [+3.12, +18.32] & 0.1038 & 207$\times$ [114, 379] \\
\bottomrule
\end{tabularx}
\end{table}

\begin{table}[tbp]
\centering\small
\caption*{\textbf{Table D.12b} (continued)}
\begin{tabularx}{\linewidth}{>{\raggedright\arraybackslash}X>{\raggedright\arraybackslash}Xll>{\raggedright\arraybackslash}X>{\raggedright\arraybackslash}Xl>{\raggedright\arraybackslash}X}
\toprule
Model $\cdot$ context (books) & Method & Keys read & Bytes read & Perplexity vs dense [95\% CI] & Perplexity vs ours [95\% CI] & KL mean & KL / ours [95\% CI] \\
\midrule
Qwen3 $\cdot$ 32K (16) & ours, per-step perturbations & 1.95\% & 30.8\% & +0.05 [$-$0.03, +0.13] & -- & 0.0003 & 1 \\
Qwen3 $\cdot$ 32K (16) & ours, fixed perturbations & 1.95\% & 30.0\% & +0.04 [$-$0.09, +0.17] & $-$0.01 [$-$0.15, +0.14] & 0.0005 & 1.57$\times$ [1.30, 1.87] \\
Qwen3 $\cdot$ 32K (16) & ours, $\varepsilon=0.2$, per step & 1.26\% & 26.7\% & +0.12 [+0.01, +0.22] & +0.07 [$-$0.01, +0.14] & 0.0007 & 2.59$\times$ [2.40, 2.77] \\
Qwen3 $\cdot$ 32K (16) & oracle top-$k$ (our keys per head) & 2.00\% & oracle & +8.44 [+4.65, +14.18] & +8.38 [+4.61, +14.05] & 0.1143 & 396$\times$ [270, 598] \\
Qwen3 $\cdot$ 32K (16) & oracle top-$k$, layers 0--1 dense & 7.39\% & oracle & +8.72 [+4.90, +14.55] & +8.66 [+4.88, +14.43] & 0.1155 & 400$\times$ [273, 603] \\
Qwen3 $\cdot$ 32K (16) & MagicPIG (reference), $K=9$, $L=300$ & 18.54\% & 19.0\% ($\ge$) & +0.18 [+0.01, +0.33] & +0.13 [$-$0.08, +0.33] & 0.0016 & 5.71$\times$ [4.82, 6.49] \\
Qwen3 $\cdot$ 32K (16) & MagicPIG (reference), $K=9$, $L=450$ & 29.38\% & 30.1\% ($\ge$) & +0.03 [$-$0.05, +0.11] & $-$0.02 [$-$0.12, +0.09] & 0.0005 & 1.57$\times$ [1.30, 1.88] \\
Qwen3 $\cdot$ 32K (16) & RetroInfer (CPU port), retrieval 14.4\% & 15.40\% & 21.7\% ($\ge$) & +0.02 [$-$0.24, +0.29] & $-$0.03 [$-$0.29, +0.23] & 0.0023 & 7.90$\times$ [6.94, 8.82] \\
Qwen3 $\cdot$ 32K (16) & RetroInfer (CPU port), retrieval 25\% & 25.77\% & 32.0\% ($\ge$) & $-$0.07 [$-$0.24, +0.10] & $-$0.12 [$-$0.30, +0.05] & 0.0009 & 3.25$\times$ [2.83, 3.66] \\
\bottomrule
\end{tabularx}
\end{table}

\begin{table}[tbp]
\centering\small
\caption*{\textbf{Table D.12b} (continued)}
\begin{tabularx}{\linewidth}{>{\raggedright\arraybackslash}X>{\raggedright\arraybackslash}Xll>{\raggedright\arraybackslash}X>{\raggedright\arraybackslash}Xl>{\raggedright\arraybackslash}X}
\toprule
Model $\cdot$ context (books) & Method & Keys read & Bytes read & Perplexity vs dense [95\% CI] & Perplexity vs ours [95\% CI] & KL mean & KL / ours [95\% CI] \\
\midrule
Qwen2.5 $\cdot$ 32K (16) & ours, per-step perturbations & 2.25\% & 41.2\% & $-$0.02 [$-$0.06, +0.02] & -- & 0.0002 & 1 \\
Qwen2.5 $\cdot$ 32K (16) & ours, fixed perturbations & 2.24\% & 40.4\% & +0.01 [$-$0.06, +0.07] & +0.02 [$-$0.07, +0.12] & 0.0004 & 1.61$\times$ [1.44, 1.79] \\
Qwen2.5 $\cdot$ 32K (16) & ours, $\varepsilon=0.2$, per step & 1.35\% & 34.3\% & $-$0.02 [$-$0.11, +0.06] & 0.00 [$-$0.08, +0.06] & 0.0006 & 2.62$\times$ [2.46, 2.78] \\
Qwen2.5 $\cdot$ 32K (16) & oracle top-$k$ (our keys per head) & 2.30\% & oracle & +11.49 [+5.66, +19.43] & +11.51 [+5.67, +19.44] & 0.1257 & 555$\times$ [318, 893] \\
Qwen2.5 $\cdot$ 32K (16) & oracle top-$k$, layers 0--1 dense & 9.11\% & oracle & +10.34 [+4.50, +18.38] & +10.36 [+4.51, +18.42] & 0.1190 & 525$\times$ [286, 866] \\
Qwen2.5 $\cdot$ 32K (16) & MagicPIG (reference), $K=9$, $L=300$ & 16.54\% & 17.0\% ($\ge$) & +0.13 [+0.01, +0.26] & +0.15 [+0.02, +0.28] & 0.0020 & 8.79$\times$ [5.90, 13.6] \\
Qwen2.5 $\cdot$ 32K (16) & MagicPIG (reference), $K=9$, $L=600$ & 37.25\% & 38.2\% ($\ge$) & 0.00 [$-$0.02, +0.03] & +0.02 [$-$0.03, +0.07] & 0.0001 & 0.45$\times$ [0.40, 0.50] \\
Qwen2.5 $\cdot$ 32K (16) & RetroInfer (CPU port), retrieval 14.4\% & 14.61\% & 20.9\% ($\ge$) & +0.28 [+0.06, +0.50] & +0.29 [+0.09, +0.50] & 0.0039 & 17.4$\times$ [14.8, 20.5] \\
Qwen2.5 $\cdot$ 32K (16) & RetroInfer (CPU port), retrieval 35\% & 34.57\% & 40.8\% ($\ge$) & $-$0.02 [$-$0.15, +0.10] & 0.00 [$-$0.14, +0.11] & 0.0007 & 3.24$\times$ [2.86, 3.62] \\
\bottomrule
\end{tabularx}
\end{table}

\begin{table}[tbp]
\centering\small
\caption*{\textbf{Table D.12c.} Free generation on PG-19 at 32K (Llama): 1,024 tokens per book sampled at $T=0.7$ with the same Gumbel noise for every method, after a dense prefill of the preceding 31,744 tokens. The KL divergence from dense attention's next-token distribution is measured at every step on the method's own prefix (a dense shadow sequence fed the method's tokens): mean and maximum per book, averaged over books, and the mean over the first and the last quarter of the steps. Ratios to ours with 95\% bootstrap CIs over books. NLL: negative log-likelihood under dense attention of the tokens the method sampled, minus that of dense attention's own samples; repeated 4-grams: share of the generated 4-grams that occurred earlier in the same continuation, minus dense attention's. Baselines at the matched configurations of Table D.12b.}
\begin{tabularx}{\linewidth}{>{\raggedright\arraybackslash}Xll>{\raggedright\arraybackslash}X>{\raggedright\arraybackslash}X>{\raggedright\arraybackslash}X>{\raggedright\arraybackslash}X>{\raggedright\arraybackslash}X}
\toprule
Method (7 books) & Mean KL & Max KL & First / last quarter & Mean KL / ours & Max KL / ours & NLL minus dense & Repeated 4-grams minus dense (points) \\
\midrule
dense attention & --- & --- & --- & --- & --- & NLL 1.157 & 9.5\% \\
ours, $\varepsilon=0.1$ & 4.5e-04 & 5.8e-02 & 4.0e-04 / 4.1e-04 & 1 & 1 & $-$0.011 [$-$0.125, +0.083] & $-$2.3 [$-$5.1, +0.6] \\
MagicPIG (reference) & 7.3e-04 & 4.8e-02 & 1.0e-03 / 5.3e-04 & 1.63 [1.32, 2.01] & 0.84 [0.68, 1.71] & $-$0.024 [$-$0.093, +0.049] & +0.0 [$-$2.1, +2.6] \\
RetroInfer (CPU port) & 1.9e-03 & 1.2e-01 & 2.1e-03 / 1.7e-03 & 4.25 [3.40, 5.43] & 1.99 [1.02, 7.16] & $-$0.212 [$-$0.424, $-$0.026] & +5.3 [$-$2.5, +14.6] \\
vAttention (HashAttention) & 1.7e-03 & 2.1e-01 & 2.5e-03 / 1.3e-03 & 3.89 [2.93, 5.14] & 3.71 [0.42, 11.82] & $-$0.085 [$-$0.254, +0.080] & $-$1.8 [$-$6.2, +2.0] \\
\bottomrule
\end{tabularx}
\end{table}

\begin{itemize}
\item
Our KL divergence does not grow from the first to the last quarter of the 1,024 steps ($4.0\cdot10^{-4}$ and $4.1\cdot10^{-4}$); the baselines' decreases.
\item
Our largest KL divergence per book is not lower than MagicPIG's (0.84 [0.68, 1.71] times ours).
\item
RetroInfer's samples have a lower negative log-likelihood under dense attention than dense attention's own ($-$0.212 [$-$0.424, $-$0.026] nats per token), together with more repeated 4-grams (+5.3 [$-$2.5, +14.6] points), which is consistent with more repetitive text. No other difference in negative log-likelihood or repetition excludes zero.
\end{itemize}

\section{Reproducibility}\label{app:E}

\subsection{Code}\label{app:E.1}

The release will contain the following (paths relative to the repository root):

\begin{table}[tbp]
\centering\small
\begin{tabularx}{\linewidth}{>{\raggedright\arraybackslash}X>{\raggedright\arraybackslash}X}
\toprule
Component & Contents \\
\midrule
Kernel and binding & \texttt{csrc/certattn/}: the C++ kernel with AVX-512 intrinsics, its pybind11 binding and a build script \\
Reference implementation and tests & the NumPy reference with exact scores (\texttt{attn/pps\_ref.py}) and the kernel and index tests (\texttt{attn/test\_certattn.py}, \texttt{attn/test\_append\_many.py}) \\
Certificate rules & \texttt{attn/certify.py} and the rule comparison of Appendix D.7 (\texttt{attn/cert\_ablation.py}) \\
Operator-level evaluation & KV-cache capture (\texttt{probe/capture.py}), baseline configurations and evaluation (\texttt{attn/methods.py}, \texttt{attn/eval\_maskers.py}) with the chunked routines of Appendix C.3, the reference ports (\texttt{attn/official\_ref.py}, \texttt{attn/retroinfer\_port.py}) and SANTA (\texttt{attn/eval\_santa.py}) \\
End-to-end evaluation and perplexity & \texttt{attn/e2e.py} with one configuration file per method, \texttt{attn/e2e\_stats.py} and \texttt{attn/ppl.py} \\
Timing & attention and whole-step timing, the append cost (\texttt{attn/bench\_append.py}) and the comparison with llama.cpp (\texttt{attn/bench\_dense\_vs\_llamacpp.py}, with the llama.cpp patch) \\
Tables and figures & one summary script per table and figure, and \texttt{paper/tools/figs.py} \\
Job scripts & one script per experiment (capture, operator-level runs, end-to-end runs, perplexity, timing, summaries) and the RULER generation script \\
\bottomrule
\end{tabularx}
\end{table}

The release documents the command that produces each table and figure from the run directories. The kernel is built and tested as follows:

\begin{small}\begin{verbatim}
# build the kernel for the host CPU
bash csrc/certattn/build.sh
# correctness tests (Appendix B.5)
python attn/test_certattn.py
\end{verbatim}\end{small}

\subsection{Third-party code, data and weights}\label{app:E.2}

\begin{table}[tbp]
\centering\small
\begin{tabularx}{\linewidth}{>{\raggedright\arraybackslash}X>{\raggedright\arraybackslash}X}
\toprule
Item & Version \\
\midrule
sparse-attention-hub (baseline maskers) & xAlg-ai/sparse-attention-hub, commit 497da5b \\
MagicPIG reference & Infini-AI-Lab/MagicPIG, commit ac9aa36 \\
Quest reference & mit-han-lab/Quest, commit 01c1623 \\
RetroInfer (ported) & microsoft/RetrievalAttention, commit 03f912c \\
HashAttention weights (Llama-3.1-8B, 64K patch v1) & sha256 \texttt{8b496b02\ldots{}} \\
LongBench prompts and metrics & THUDM/LongBench, commit 2e00731f8d \\
LongBench-v2 data & \texttt{data.json}, sha256 \texttt{15d61c22\ldots{}} \\
RULER generator & NVIDIA/RULER, commit c3f5e3b4f87f97e048793bb510a3a6b19a46bf3a; \texttt{prepare.py}, seed 42, base template, maximum sequence length $n-128$ \\
RULER generated files & sha256 checksums of all 14 files, included in the release; the first 8 records of every file match the files used for KV capture \\
PG-19 test set & mirror of emozilla/pg19-test, 100 books, sha256 \texttt{9aae5ddf\ldots{}} \\
llama.cpp (dense-kernel check) & ggml-org/llama.cpp, commit 99b9548, with a patch for the thread count and the decode shapes (included in the release) \\
\bottomrule
\end{tabularx}
\end{table}

Each run directory records the host, the CPU affinity, the load average, the library versions, the code commit and the arguments. All reported numbers are produced by the summary scripts from these run directories.

\subsection{Compute}\label{app:E.3}

All experiments ran on four shared dual-socket Xeon servers (Appendix C.4) without GPUs. We did not meter total CPU hours. For scale: a 128K fp32 prefill of Llama-3.1-8B takes 38--40 minutes in the whole-step timing setup (Appendix C.5), a 128K RULER sample with all end-to-end methods takes about 80 minutes, and the baselines' research code needs about 36 s per decoding step at 32K.

\subsection{Data availability}\label{app:E.4}

\textbf{Data availability.} All models (Llama-3.1-8B-Instruct, Qwen3-8B, Qwen2.5-7B-Instruct) and datasets (LongBench, LongBench-v2, PG-19) are publicly available. RULER data were generated with the official scripts (seed 42), and the checksums of all generated files are part of the release (Appendix E.2). Code, captured-data generation scripts, result tables and the scripts that produce every number in this paper will be released.

\section{Extended Discussion}\label{app:F}

\subsection{Related work in detail}\label{app:F.1}

\textbf{Sparse attention for decoding.} Some training-free methods keep attention sinks and recent tokens \citep{xiao2024sinks}, accumulated heavy hitters \citep{zhang2023h2o} or a compressed prompt cache \citep{li2024snapkv}. Others estimate which keys matter for the current query from page summaries \citep{tang2024quest,xiao2024infllm}, low-rank or partial projections \citep{ribar2024sparq,yang2024double,sun2025shadowkv}, product quantization \citep{zhang2025pqcache}, clustering \citep{liu2025clusterkv}, learned hash codes \citep{desai2025hash} or adaptive top-$p$ budgets \citep{lin2025twilight}. Other work separates retrieval heads from streaming heads \citep{xiao2025duo}, trains sparsity into the model \citep{yuan2025nsa,lu2025moba}, or finds heavy keys with approximate nearest-neighbor indices \citep{liu2025retrieval,chen2026retroinfer,mao2024iceformer}. These methods reduce the number of keys read, but they drop the mass of the unselected keys and give no per-query bound on the resulting error.

\textbf{Sampling and estimation.} Several methods estimate attention from a sample of keys. MagicPIG \citep{chen2025magicpig} uses locality-sensitive hashing as an importance-sampling proposal. vAttention \citep{desai2026vattention} adds uniform samples of the remaining keys to a predicted top-$k$ set. It chooses the sample size so that the numerator, the denominator and their ratio are $(\varepsilon,\delta)$-accurate under the central limit theorem, with variances estimated from a base sample. Its released implementation, which we evaluate, budgets for the denominator alone. SANTA \citep{lee2026stochastic} computes all scores exactly and draws a fixed number of values i.i.d. from the resulting softmax, which gives an unbiased output estimate with a variance and a vector Bernstein bound. A variant of SANTA also samples the score computation. The concurrent SANTA++ \citep{lee2026santa} samples clusters through representatives with inverse-probability weights. SOCKET \citep{joshi2026socket} uses soft collision kernels. LSH attention \citep{kitaev2020reformer}, random features \citep{choromanski2021performer} and kernel-density or low-rank estimators \citep{zandieh2023kde,han2024hyper,chen2021scatterbrain,zandieh2024subgen} approximate whole attention matrices. Their guarantees, where given, bound the error of the whole matrix. For top-$k$ truncation, the discarded tail mass yields deterministic per-query bounds once all exact scores are known \citep{tzachristas2025topk}. Our estimator differs from these methods in three respects. First, it samples keys in proportion to their weights without scoring the unsampled keys exactly, and it keeps heavy keys exact. Second, its confidence bounds hold for finite samples without variance plug-ins, and they hold at a stopping level that each head chooses from the data. Third, Lemma 1 lets them bound the output error directly, without splitting the error budget between numerator and denominator. To our knowledge, no prior work uses perturbed scores to turn retrieval into certified attention estimation.

\textbf{Theory.} Fast attention requires bounded entries \citep{alman2023}, and compression and streaming lower bounds limit how far a KV cache can be summarized \citep{haris2025,chen2026streaming,kochetkova2025,addanki2023}. Our lower bound concerns a different resource, the number of values read by an algorithm that knows all keys. On the sampling side, threshold and priority sampling estimate subset sums with near-optimal variance \citep{duffield2005threshold,duffield2007priority,szegedy2006}. Exponential and Gumbel perturbations implement weighted sampling \citep{efraimidis2006,maddison2014astar,kool2019}, and perturbed or hashed near-neighbor search estimates partition functions and kernel densities \citep{mussmann2016,mussmann2017,spring2017,charikar2017,charikar2020,karppa2022}. The work closest to our use of perturbations is that of \citet{mussmann2017}, who append Gumbel perturbations to a database and use maximum inner product search to sample from log-linear models with provable guarantees. \citet{mussmann2026sum} assign each object to one of $O(\log n)$ levels with exponentially decaying probabilities and query a similarity-search index per level, which gives an unbiased estimate of a sum such as a softmax normalizing constant from $O(\log n)$ retrieved vectors, with a high-probability relative error bound. We use a threshold instead of the top perturbed scores or a top-$k$ per level, which gives nested samples with constant HT weights and a certificate that can be checked at every level, so the sample size adapts to each query. We also certify an attention output, which is a ratio of vector sums. Our certificates combine Chernoff bounds for Poisson-binomial counts, a Bernstein-type inequality in Hilbert space \citep{pinelis1994} and a union bound over a geometric grid. Sharper tools, such as Bousquet's inequality \citep{bousquet2002}, matrix and empirical Bernstein bounds \citep{tropp2011,maurer2009} and time-uniform confidence sequences \citep{howard2021,waudbysmith2023}, could tighten our certificates (Appendix D.7 and Section 7).

\subsection{Conservative certificates and linear candidate generation}\label{app:F.2}

\textbf{Conservative certificates.} Observed errors are about $\varepsilon/10$ at the 90th percentile (Section 6.6 gives the cost in keys). The slack comes from the union bound over levels and events (a logarithm of about 10, against 1.9 for a normal quantile), the range term driven by the largest value norm, and the upper confidence bound on the spread of the light values. Time-uniform confidence sequences and betting estimators \citep{howard2021,waudbysmith2023} address the first and the third.

\textbf{Linear candidate generation.} The 4-bit scan moves a fixed 16.4\% of the bytes of dense attention, counting the write of the per-step perturbations (19.3--19.5\% in total at 128K). This floor caps the speedup, and it grows with $n$ while the keys and values we read do not. At equal bytes it also lets the best baselines read many more keys and match or beat our 90th-percentile error on five of ten settings (down to 0.6 times), although not our maximum error (Section 6.2). A sublinear index that misses at most an $\eta$ fraction of the attention mass in expectation would bias the denominator by at most $\eta D$ (Proposition 2, Appendix A.9). Certified sublinear candidate generation for attention keys, however, remains open.

\subsection{Large groups, strict measures and the start of the grid}\label{app:F.3}

\textbf{Large groups and strict measures.} With seven query heads per KV head (Qwen2.5), the candidate union is large and dense attention is cheap, so we see no speedup at 16K. The output-norm measure is slower than dense attention at 32K and below (and at 64K on one thread of the 8358), and Theorem 3 shows that this cost is partly inherent. Because the grid starts at the largest weight, near-uniform attention makes the method read $O(D)$ keys, but real heads have $D/n$ below 1\% (Appendix B.5). At $\varepsilon=0.1$, 0.013\% of the head-steps (value scale) stop at the first level, all at 128K, where they read 3.9--8.1\% of the keys against a median of 0.26--0.27\%. Starting the grid above $\tau=1$, for example at an upper bound on $D$ from the 4-bit scan divided by a minimum sample size, would serve such heads and keep Theorem 4 valid, since the grid would then depend on the query and keys but not on the perturbations.

\textbf{Starting the grid from a bound on $D$.} We implemented this start as a kernel option. Each head sums $D_4=\sum_{i\in P}e^{u_i-c^\ast}$ over the pool from its 4-bit upper bounds, which bounds the pool's part of $D$, and starts the grid at $\tau_0=\max(1,D_4/s)$ for a target size $s$ of the first sample, by raising $c^\ast$ by $\log\tau_0$. The union bound over the $J+1$ levels is unchanged, and $\tau_0$ depends on the query and the keys but not on the perturbations, so Theorem 4 holds. With $s=256$ at $\varepsilon=0.1$ (value scale, all ten settings), the head-steps that stopped at the first level read 3.4--5.7\% of the keys (mean 4.4\%) instead of 3.9--8.1\% (mean 5.3\%). Of all head-steps, 1.7\% read a different number of keys, the mean keys read of a setting changes by at most 0.23\%, and the 90th-percentile errors change by at most 0.00001 and the maximum errors not at all except one lower maximum (Qwen2.5 LongBench), with no violation and no fallback; $s=1024$ gives the same result on these head-steps and changes 0.3\% of all head-steps. The first level thus overshot by about a fifth, and the rest is the sample that a value-scale certificate at $\varepsilon=0.1$ needs on a near-uniform head (Proposition 1). Given these small changes, all other experiments use the default $\tau_0=1$.

\subsection{Fixed perturbations and reproducibility}\label{app:F.4}

\textbf{Fixed perturbations.} Theorem 4 assumes that the query, keys and values of a call are independent of its perturbations. Fresh perturbations at every step, our default, keep this in closed-loop generation. Perturbations fixed per KV head do not, because later queries, and the keys and values of generated tokens, depend on earlier estimates. As an ablation, fixed perturbations make attention alone a median 4\% faster but whole decoding steps no faster measurably (Tables D.11a, D.11d and D.11e), and on Llama LongBench, averaged over three seed families, they score like fresh ones (per step minus fixed: $-$0.10 [$-$0.35, +0.10]; Table D.9e). On PG-19, their KL divergence from dense attention is 1.78 [1.25, 2.40], 1.60 [1.36, 1.93] and 1.41 [1.27, 1.54] times that of fresh perturbations on Llama at 32K, 64K and 128K, and 1.57 [1.30, 1.87] and 1.61 [1.44, 1.79] times on Qwen3 and Qwen2.5 at 32K (Appendix D.12). On Llama common-words extraction, fresh perturbations score 2.6 points below dense attention ([$-$8.2, +1.4]; 2 wins and 2 losses), with identical outputs in all four runs that include them. The fixed variant scored 28.0 in two runs and 30.0 in a third with the same seeds, which changed 3 of its 50 outputs, against 27.8 for dense attention; the runs differed in CPU model and thread count, which change the floating-point rounding (below). Fresh minus fixed perturbations is therefore $-$2.8 [$-$7.8, +1.2] in the first runs and $-$4.8 [$-$10.4, $-$0.4] in the third (3 wins and 5 losses, sign test $p=0.73$). Reuse would matter only where rehashing the pool is expensive, and bounding its cost, for example with adaptive data analysis \citep{dwork2015}, is open.

\textbf{Reproducibility of end-to-end outputs.} Table F.1 compares every pair of end-to-end runs that decoded the same sample with the same method and seeds. Runs on the same CPU model gave identical outputs except one pair of the fixed variant, whose runs used one NUMA node and the whole machine, and thus different numbers of PyTorch threads. Across the two CPU models, dense attention never changed its output, our default changed one of 270 (a LongBench code-completion output) and the fixed variant 12 of 261, among them the three common-words outputs above. Our kernel gives bitwise identical outputs on both CPU models and for 1 and 8 threads. The dense fp32 prefill that all methods share is rounded differently with 16 and 26 PyTorch threads (its KV cache differs from the second layer on). Reruns with code, batch and seeds unchanged reproduce the changed common-words outputs when both the thread count and the CPU model match the original run, so the CPU model matters as well, through rounding we did not locate. Dense attention's greedy decoding absorbs such rounding differences, whereas our method makes discrete decisions: a rounding-level change in a query can move a key across a threshold or change the stopping level, which changes the estimate by an amount of the order of its error, and occasionally a token. End-to-end outputs are thus bitwise reproducible for a given CPU model and thread count, and the certificates hold in either case.

\begin{table}[tbp]
\centering\small
\caption*{\textbf{Table F.1.} Pairs of end-to-end runs that decoded the same sample with the same method configuration and seeds: identical outputs / pairs, by whether the shards that produced them ran on the same CPU model (128 pairs whose host could not be determined are left out; all of them were identical).}
\begin{adjustbox}{max width=\linewidth}\begin{tabular}{lll}
\toprule
Method & Same CPU model & Different CPU models \\
\midrule
Dense attention (model SDPA) & 1195 / 1195 & 1245 / 1245 \\
Ours, fresh perturbations at every step & 637 / 637 & 269 / 270 \\
Ours, perturbations fixed per KV head & 193 / 194 & 249 / 261 \\
\bottomrule
\end{tabular}\end{adjustbox}
\end{table}

\subsection{Scope}\label{app:F.5}

\textbf{Scope.} We evaluate only on CPUs and have no GPU implementation. On GPUs, dense attention runs as an IO-aware fused kernel \citep{dao2022flash} over paged caches \citep{kwon2023paged}. At batch size 1, the fp32 linear layers dominate a CPU step and cap the whole-step gain (Section 6.5). We compare baselines by the keys and bytes they read and do not time them, since we run them as sparse-attention-hub maskers and ports in PyTorch rather than as optimized kernels. Dense attention over an 8-bit quantized cache has a 3--5 times lower 90th-percentile error than our method but reads 1.5--2.8 times our bytes (Section 6.2); where those bytes are affordable, it is the simpler choice. With 48--300 samples per setting, the end-to-end confidence intervals exclude large losses, and smaller differences remain within their width. The sampling runs are smaller (11 cwe samples on Llama and Qwen2.5, none on Qwen3, and seven PG-19 books) and detect only large differences. Uses that act on the certificate itself remain to be evaluated, such as sending a request to exact attention when its heads cannot be certified within a budget, or dividing a compute budget among requests by the keys their certificates need. Whether the guarantee also comes free in speed at equal bytes and quality depends on timing our kernel and the baselines' kernels on the same hardware. Baseline hyperparameters at 64K--128K extrapolate grids tuned at 16K.

\end{document}